%% file: main.tex
\documentclass[11pt]{article}

\usepackage{acl}

\usepackage{times}
\usepackage{latexsym}
\usepackage[T1]{fontenc}
\usepackage[utf8]{inputenc}
\usepackage{microtype}
\usepackage{inconsolata}
\usepackage{cuted}

\usepackage{graphicx}
\usepackage{amsmath}
\usepackage{amsfonts}
\usepackage{bm}
\usepackage{array}
\usepackage{booktabs}
\usepackage{multirow}
\usepackage{tabularx}
\usepackage{wrapfig}
\usepackage{subcaption,ragged2e}
\usepackage{caption}
\usepackage{adjustbox}
\usepackage{float}
\usepackage{enumitem}

\usepackage{xcolor}
\definecolor{mypurple}{RGB}{128, 0, 128}
\usepackage{pifont}

\usepackage{listings}

\usepackage{tikz}
\usetikzlibrary{positioning,calc}
\usepackage[most,skins,theorems,breakable]{tcolorbox}

\hypersetup{
    colorlinks=true,
    citecolor=mypurple,
    linkcolor=mypurple,
    urlcolor=mypurple
}
\usepackage{breakcites}
\usepackage[capitalize]{cleveref}

\usepackage{fontawesome5}

\usepackage[normalem]{ulem}

\usepackage{footnote}

\newcolumntype{Y}{>{\raggedright\arraybackslash}X}

\definecolor{InstructBlue}{HTML}{1F4E79}
\definecolor{ThinkingOrange}{HTML}{C65911}

\lstdefinestyle{promptroman}{
  basicstyle=\small\rmfamily,
  breaklines=true,
  breakatwhitespace=false,
  columns=fullflexible,
  keepspaces=true,
  showstringspaces=false,
  upquote=true,
  literate={─}{{-}}1,
  escapeinside={(@}{@)},                  
  moredelim=[is][\color{InstructBlue}]{@@INSTRUCT@@}{@@/INSTRUCT@@},
  moredelim=[is][\color{ThinkingOrange}]{@@THINKING@@}{@@/THINKING@@},
}

\usepackage{xcolor}
\newcommand{\tokent}[4]{\colorbox[RGB]{#1,#2,#3}{\strut #4}\hskip 0pt plus 0.5pt\relax}
\newcommand{\tokensub}[1]{\textsubscript{\textcolor[RGB]{26,79,216}{\tiny #1}}\allowbreak}

\newcommand{\prev}{WebGym}
\newcommand{\ours}{AsyncWebRL}

\newenvironment{smalldisplay}{%
  \par
  \begingroup
  \setlength{\abovedisplayskip}{10pt plus 3pt minus 7pt}%
  \setlength{\belowdisplayskip}{10pt plus 3pt minus 7pt}%
  \setlength{\abovedisplayshortskip}{0pt plus 3pt}%
  \setlength{\belowdisplayshortskip}{6.5pt plus 3.5pt minus 3pt}%
  \small
}{%
  \endgroup
}

\usepackage[most,skins,theorems]{tcolorbox}

\tcbset{
  aibox/.style={
    width=\linewidth,
    top=8pt,
    bottom=4pt,
    colback=blue!6!white,
    colframe=black,
    colbacktitle=black,
    enhanced,
    center,
    attach boxed title to top left={yshift=-0.1in,xshift=0.15in},
    boxed title style={boxrule=0pt,colframe=white,},
  }
}
\newtcolorbox{AIbox}[2][]{aibox,title=#2,#1}

\tcbset{panelboxbase/.style={enhanced jigsaw,sharp corners,boxrule=0.4pt,
  colback=white,colframe=black!55,left=2pt,right=2pt,top=2pt,bottom=2pt,
  fonttitle=\normalfont\bfseries\footnotesize,coltitle=black,
  colbacktitle=white,
  fontupper=\footnotesize,
  before upper={\raggedright\noindent}}}
\tcbset{panelboxfirst/.style={panelboxbase,bottomrule=0pt,bottom=0pt,
  before skip=2pt,after skip=0pt}}
\tcbset{panelboxpiece/.style={panelboxbase,toprule=0pt,bottomrule=0pt,
  top=0pt,bottom=0pt,
  before skip=-\parskip,after skip=0pt}}
\tcbset{panelboxlast/.style={panelboxbase,toprule=0pt,top=0pt,
  before skip=-\parskip,after skip=2pt}}
\tcbset{panelbox/.style={panelboxbase,before skip=2pt,after skip=2pt}}
\newenvironment{panelmd}[1]{%
  \par\nobreak\vspace{2pt}%
  \noindent\hrule height 0.4pt%
  \vspace{2pt}%
  {\noindent\normalfont\bfseries\footnotesize #1\par}%
  \nobreak\vspace{1pt}%
  \noindent\hrule height 0.4pt%
  \vspace{2pt}%
  \footnotesize\raggedright\noindent\ignorespaces%
}{%
  \par\nobreak\vspace{2pt}%
  \noindent\hrule height 0.4pt%
  \par\vspace{2pt}%
}

\title{\raisebox{-0.32\height}{\includegraphics[height=1.5em]{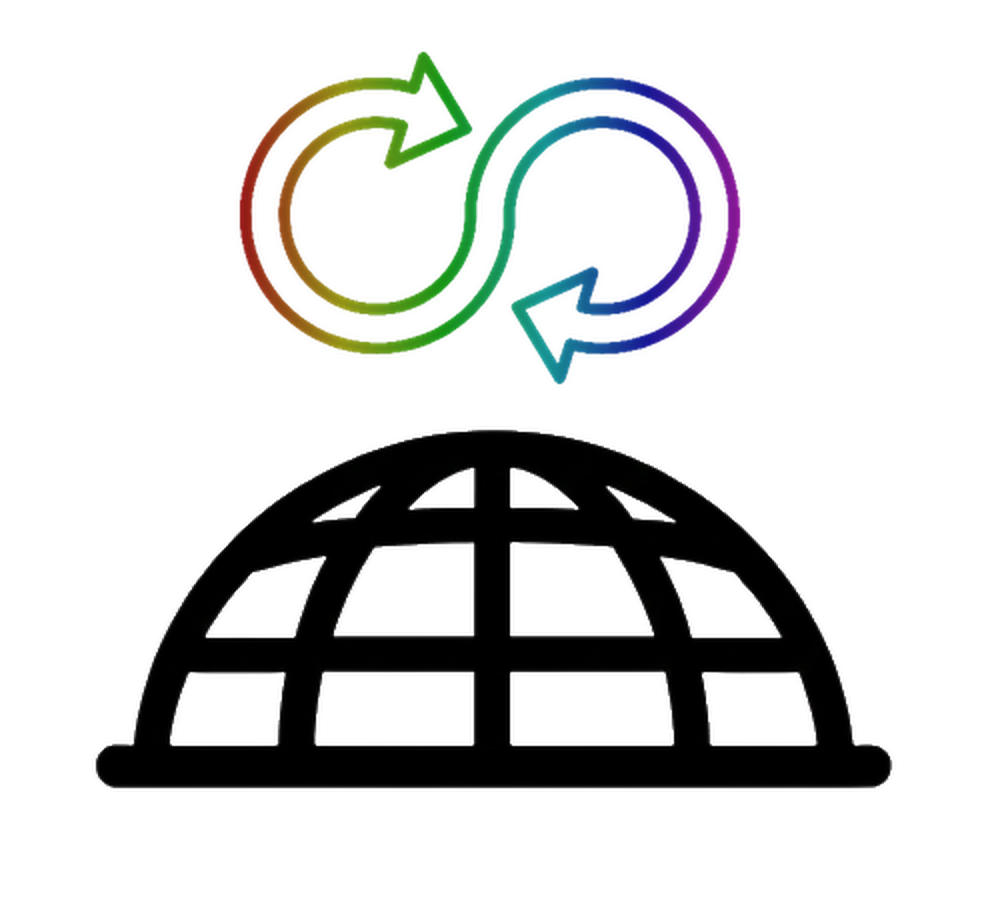}}\hspace{0.35em}AsyncWebRL: Efficient Asynchronous Reinforcement Learning for Multi-Step Visual Web Agents}

\author{\textbf{Hao Bai}$^{1,2}$ \quad \textbf{Rui Yang}$^{1,2}$ \quad \textbf{Chenlu Ye}$^{1}$ \\
  \textbf{Spencer Whitehead}$^{2}$ \quad  \textbf{Aviral Kumar}$^{3}$ \quad \textbf{Tong Zhang}$^{1}$ \\[0.5em]
  $^{1}$UIUC \quad $^{2}$Microsoft \quad $^{3}$CMU \\[0.7em]
  \normalsize
  \href{https://asyncwebrl-website.github.io/}{\faGlobe\,\textsc{Project Page}}\hspace{1.8em}
  \href{https://github.com/microsoft/webgym/tree/main}{\faGithub\,\textsc{Code}}\hspace{1.8em}
  \href{https://huggingface.co/datasets/microsoft/webgym_tasks}{\faDatabase\,\textsc{Task Set}}
}

\begin{document}

\maketitle

\begin{abstract}
Training vision-language web agents with multi-step RL is compute-intensive, with two dominant forms of inefficiency: idle GPUs in synchronous RL, and trajectories that use more steps and tokens than necessary. We present \textbf{AsyncWebRL}, which addresses both. On the system side, an asynchronous design overlaps rollout, gradient update, and policy refresh across iterations, paired with two web-agent-specific adaptations, namely an everlasting rollout pool and lightweight screenshot handling, that together deliver up to a $\bm{2.9\times}$ end-to-end training-throughput speedup over the previously fastest open synchronous pipeline (WebGym). On the algorithmic side, we identify the per-trajectory normalizer $1/|\tau_i|$ in multi-step GRPO as the root cause of trajectory-level and token-level inefficiency: because failures are systematically longer than successes, it down-weights the negative gradient on failed tokens, so the policy keeps producing verbose memory schemas. Replacing $1/|\tau_i|$ with a constant $1/k$ breaks this coupling, contracting trajectories while preserving aggregate success. Together, these contributions set a new open-source state of the art on the WebGym out-of-distribution test split ($\bm{+5.8\%}$ relative over the $42.9\%$ prior best), with the largest gains on the harder slices ($\bm{+42\%}$ relative on Medium, $\bm{+48\%}$ relative on Hard).
\end{abstract}

\section{Introduction}

Vision-language web agents trained with multi-step reinforcement learning have rapidly become a leading recipe for autonomous browsers that complete real-world tasks across the long tail of the web~\citep{webgym, ui-tars-2, webagent-r1, gui-libra}, part of a broader shift from single-turn preference tuning to long-horizon, environment-grounded agent training. The defining feature of this regime is compute: each training run consumes hundreds of GPU-hours against hundreds of concurrent browser sessions, and progress is largely a function of how many trajectories the trainer can profitably consume per wall-clock hour. Against this backdrop, any compute inefficiency translates directly into a lower-performing agent at fixed budget. This paper targets both forms of inefficiency within a single framework, combining (1) a fully asynchronous system design tailored to per-step visual rollouts with (2) a one-line algorithmic fix that contracts trajectory and token usage without harming ending success.

The natural systems solution for GPU idleness is \emph{async RL}, where rollout, gradient update, and policy refresh proceed concurrently across iteration boundaries~\citep{a3c, impala, decoupled-policy-optimization}. On the surface this appears to be a solved problem: single-step LLM-RL is already async, and \prev{}~\citep{webgym} already makes within-iteration rollouts async. Combining the three properties \emph{visual, multi-step, and fully async}, however, requires a coordinated design that none of the existing open-source systems carry together (\Cref{sec:related-work}). The core difficulty has two parts. First, the natural payload of a visual multi-step rollout, tens of high-resolution screenshots per trajectory shared across workers every step, swamps the standard inter-worker data store and pushes the framework into a slow disk-spill path that erases any async benefit. Second, an iteration-synchronous rollout pool of the kind \prev{} uses pays a warm-up cost on hundreds of browser sessions every round. 

We address both with two web-agent-specific designs: lightweight screenshot handling and an everlasting rollout pool. The former keeps image tensors out of the shared data store entirely, routing only lightweight references between rollout workers and the trainer. The latter keeps workers continuously alive across iteration boundaries so episodes hand off immediately and parameter updates of $\pi_\theta$ never stall collection (\Cref{sec:method-system}). Closing the resulting off-policy gap also requires an algorithmic correction. We adopt the decoupled-PPO factorization of \citet{decoupled-policy-optimization}, splitting the standard $\pi_\theta/\pi_{\mathrm{behave}}$ ratio into a rollout-staleness term and a current-update term and centering PPO-style clipping~\citep{ppo} on a proximal policy $\pi_{\mathrm{prox}}\approx\pi_\theta$. As we will show in \Cref{sec:method-algorithm} and \Cref{sec:exp-efficiency}, this roughly halves the clip-trigger rate and substantially speeds reward improvement. The resulting system is, to our knowledge, the fastest reported open multi-step RL framework for visual web agents, at a $\bm{2.4}$ to $\bm{2.9\times}$ end-to-end speedup over prior open rollout pools~\citep{webgym}.

Beyond this form of off-policy correction, we identify that per-trajectory step-number based normalization\footnote{Throughout the paper, ``step-number normalization'' refers to the multi-step GRPO factor $1/|\tau_i|$ that normalizes the loss by the number of \emph{steps} (not tokens) in a trajectory. This is distinct from Dr.~GRPO's removal of the single-turn factor $1/|y_i|$, which normalizes by the number of tokens in a response. We call the latter ``token-number normalization.''}, i.e., $\bm{1/|\tau_i|}$ in multi-step GRPO~\citep{deepseekmath} as the root cause of trajectory- and token-level inefficiency, and replace it with a constant $\bm{1/k}$ (\Cref{sec:memory-drift}). The intuition follows Dr.~GRPO~\citep{dr.grpo} but at a step-level granularity rather than token granularity: in our setting failures average $12.5$ steps against $5.1$ for successes, so $1/|\tau_i|$ attenuates the gradient on failed tokens by $\approx 2.4\times$, and the policy responds by producing verbose memory schemas. Combined with the off-policy correction, this sets a new state of the art on the \prev{} OOD test split (\Cref{sec:exp-main}).

\textbf{Our contributions.} \textbf{(1)~System.} The first open multi-step RL framework for visual web agents that is fully async end-to-end, via an everlasting rollout pool, lightweight screenshot handling, and a decoupled importance-sampling ratio (\Cref{sec:method-system}); this delivers a $2.4$ to $2.9\times$ end-to-end speedup over the previously fastest open sync pipeline and a new open-source state of the art on the \prev{} OOD test split ($+5.8\%$ relative over the $42.9\%$ prior best). \textbf{(2)~Algorithm.} A diagnosis of the per-trajectory step-number normalizer $1/|\tau_i|$ in multi-step GRPO as the root cause of trajectory- and token-level inefficiency, traced through length-coupled memory drift (\Cref{sec:memory-drift}), and a one-line fix replacing $1/|\tau_i|$ with a constant $1/k$ that drives the largest gains on the harder OOD slices ($+42\%$ relative on Medium, $+48\%$ relative on Hard over the prior \prev{} SOTA).

\vspace{-0.2cm}
\section{Related Work} \label{sec:related-work}
\vspace{-0.2cm}

\textbf{System: visual, multi-step, and async RL.}
Multi-step RL on VLM policies has been studied for web browsing~\citep{webgym, webagent-r1, gui-libra}, computer use~\citep{digirl, digi-q, ui-tars-2}, embodied learning~\citep{robot-r1, rlvla}, and sequential decision-making~\citep{rl4vlm}. Open-source frameworks split into two disjoint subsets that \ours{} bridges. \emph{Async LLM-RL on text} and \emph{async single-turn VLM-RL}~\citep{areal, roll-flash, laminar, llamarl, asyncflow, streamrl, april, async-rlhf, verl} ship at most one image per training example; pushing multi-turn web-agent rollouts (tens of high-resolution screenshots per trajectory, hundreds concurrent) through their shared data store exhausts its budget and triggers a disk-spill path that erases the async benefit. \emph{Sync multi-step VLM-RL}~\citep{webgym, webagent-r1, digirl, digi-q, openwebrl} carries the right workload but resynchronizes at every iteration boundary, leaving GPU bubbles. DART-GUI~\citep{dart-gui} also decouples rollout and training for multi-turn visual agents, but targets a substantially more I/O-bound OS-level regime, where slow environment interaction and adaptive data curation are the primary bottlenecks; in contrast, our high-cadence web setting is limited by sustained screenshot traffic and iteration-level synchronization. Switching synchronous multi-step VLM-RL to fully async execution opens an off-policy gap that sync frameworks never see: training batches contain tokens drawn from several policy versions at once, and the trust region must be re-centered on a proximal policy~\citep{decoupled-policy-optimization} so that clipping reflects current-update movement rather than rollout staleness. The closed-source UI-TARS-2~\citep{ui-tars-2} reportedly satisfies all three properties but releases only model weights. \ours{} is, to our knowledge, the first open framework to combine all three in the high-throughput visual web-agent setting.

\textbf{Algorithm: loss shape under length asymmetry.} On the algorithmic side, Dr.~GRPO~\citep{dr.grpo} showed that the single-turn GRPO normalizer $1/|y_i|$ reweights per-token gradients between long and short responses, and removing it (replacing $1/|y_i|$ with a constant) corrects an unintended length bias inside a single response. We extend the same diagnosis one level up. In multi-step GRPO~\citep{deepseekmath, mt-grpo}, the analogous factor is the per-trajectory step-number normalizer $1/|\tau_i|$, which couples per-token gradient scale to trajectory length. What is new at the step granularity that the single-turn analysis does not see is that the multi-step web-agent setting has two structural features which turn Dr.~GRPO's step-level prediction into a self-reinforcing failure mode: failed trajectories are dominated by horizon exhaustion rather than clearly wrong actions, and an additive memory prompt against a multi-step visual environment mechanically lengthens per-step responses as step count grows. We trace these features in \Cref{sec:memory-drift} and show that the multi-step counterpart of Dr.~GRPO's fix breaks the loop at its source.

\section{\ours{}} \label{sec:method}

\subsection{System} \label{sec:method-system}

\begin{figure}[t]
    \centering
    \includegraphics[width=\linewidth]{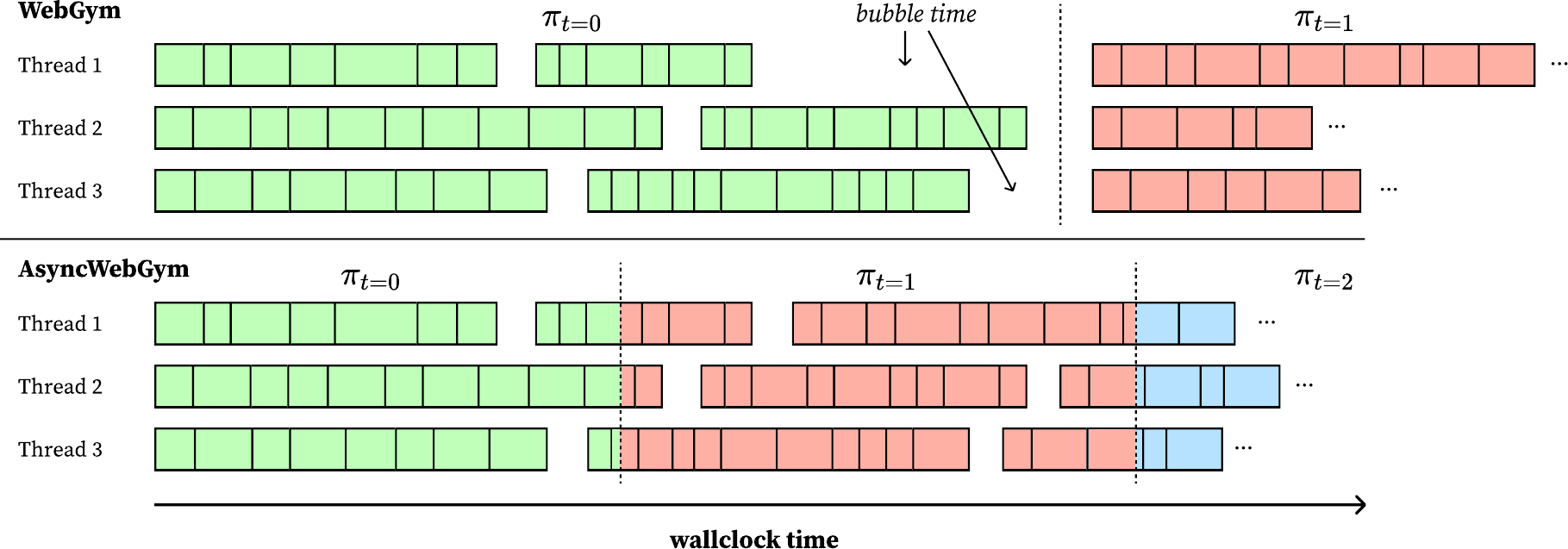}
    \caption{\small \textbf{Multi-step Asynchronous Management.} Compared to \prev{}, \ours{} eliminates the \textit{inter-iteration bubble time} caused by reconstructing the rollout pool at every iteration boundary and waiting for the policy refresh of $\pi_t$ to complete. Colored blocks denote concurrent rollout workers producing trajectories, gradient updates on $\pi_t$, and policy refreshes that broadcast new weights to the rollout workers. White gaps under synchronous RL (top) are bubble time. \ours{} (bottom) eliminates these gaps by maintaining an everlasting rollout pool so that rollout, gradient update, and policy refresh overlap continuously.}
    \label{fig:async-system}
\end{figure}

Our goals are to keep GPUs from waiting on each other, and to keep the shared inter-worker store from saturating under per-step image traffic. On top of the synchronous multi-step rollout pool of \prev{}~\citep{webgym}, we make two changes: \emph{fully asynchronous execution with an everlasting rollout pool}, and \emph{lightweight screenshot handling}. The first keeps rollout workers continuously alive across iteration boundaries and overlaps rollout, gradient update, and policy refresh end to end: when an episode ends, the next begins immediately on the same worker without waiting for the rest of the batch or for the next iteration. The parameter update of $\pi_\theta$ can happen at any time while rollout continues; new weights are broadcast in place to the inference workers and the next rollout segment is sampled under the updated policy (\Cref{fig:async-system}). The second (\Cref{app:impl-details}) keeps per-step image tensors out of the shared store and routes only lightweight references between rollout workers and the trainer, which prevents the disk-spill path that hundreds of concurrent high-resolution rollouts otherwise induce. Together these eliminate the GPU idle time spent waiting for the slowest trajectory in a batch and the warm-up cost paid on hundreds of browser sessions when the rollout pool is rebuilt each round.

\subsection{Algorithm} \label{sec:method-algorithm}

\begin{figure*}[t]
\centering
\begin{minipage}{\textwidth}
\begin{align}
\mathcal J(\theta) = \mathbb E_{\tau \sim \pi_{\mathrm{behave}}}\!\left[\frac{1}{G \cdot {\color{red}k}} \sum_{i=1}^{G} \sum_{j=1}^{|\tau_i|} \sum_{t=1}^{|\tau_{i,j}|} \min\!\left(\frac{\pi_\theta}{\pi_{\mathrm{behave}}}\hat A_i,\; {\color{red}\frac{\pi_{\mathrm{prox}}}{\pi_{\mathrm{behave}}}}\,\mathrm{clip}\!\left({\color{red}\frac{\pi_\theta}{\pi_{\mathrm{prox}}}},\, 1{-}\epsilon,\, 1{+}\epsilon\right)\hat A_i\right)\right]\!. \label{equ:multi-step-grpo}
\end{align}
\end{minipage}
\end{figure*}

\textbf{Decoupled off-policy correction.} Asynchronous execution shifts the sampling distribution: the policy that generated a given token ($\pi_{\mathrm{behave}}$) is several updates behind the policy $\pi_\theta$ the trainer is now updating, and a single trajectory in a training batch can be stitched from different policy snapshots. The natural correction is importance sampling: weight every token by $\pi_\theta / \pi_{\mathrm{behave}}$ for an unbiased off-policy gradient, and clip this ratio to $[1 - \epsilon, 1 + \epsilon]$ as in PPO. Under async RL the same ratio must capture two quantities at once: how much the policy has moved since the rollout was sampled (\emph{rollout staleness}), and how much the optimizer has moved the policy during the current gradient update. Clipping a single coupled ratio confounds the two, so rollout staleness alone triggers clip events at many token positions and slows training substantially. We adopt the decoupled-PPO factorization of \citet{decoupled-policy-optimization} (also used by AReaL~\citep{areal} for text RL) and split the ratio into a rollout-staleness factor $\pi_{\mathrm{prox}} / \pi_{\mathrm{behave}}$ and a current-update factor $\pi_\theta / \pi_{\mathrm{prox}}$, where $\pi_{\mathrm{prox}}$ is the policy snapshot at the start of the current update. The PPO clip is then centered around $\pi_{\mathrm{prox}}$ only, so that clip events reflect how much the optimizer has moved the policy during this update rather than staleness inherited from the rollout. Empirically this roughly halves the clip rate relative to the coupled formulation (\Cref{sec:exp-efficiency}), and enters the final loss as the inner $\min$ of \Cref{equ:multi-step-grpo} (a negative-advantage dual-clip extension is given in \Cref{equ:dual-clip}).

\textbf{Removing trajectory-length normalization.} Separately from the off-policy correction, we change the loss aggregation. The standard multi-step GRPO~\citep{deepseekmath} loss normalizes each rollout by its own step count $|\tau_i|$, which under our setting introduces a length-coupled gradient attenuation we trace in \Cref{sec:memory-drift}. The fix is one line: replace $1/|\tau_i|$ with a constant $1/k$, where $k$ is the Easy-difficulty horizon ($10$ throughout this paper). That is, instead of every rollout entering the loss with total weight $1$ regardless of length, each rollout enters with weight $|\tau_i|/k$, restoring full per-token gradient weight on the long failures the policy must learn to avoid. This change appears as the constant $1/k$ outer factor in \Cref{equ:multi-step-grpo}.

We treat each trajectory as a bandit and share a single trajectory-level advantage $\hat A_i = (r_i - \mathrm{mean}(\mathbf r))/\mathrm{std}(\mathbf r)$ across all tokens in trajectory $i$. With $G$ rollouts per task, $|\tau_i|$ steps per rollout, and $|\tau_{i,j}|$ tokens per step, the \ours{} loss combines the two changes above with the standard GRPO surrogate (\Cref{equ:multi-step-grpo}).

The red-colored terms in \Cref{equ:multi-step-grpo} mark these two changes: the constant $1/k$ in place of $1/|\tau_i|$, and the decoupled clip centered around $\pi_{\mathrm{prox}}$ with $\pi_{\mathrm{prox}}/\pi_{\mathrm{behave}}$ as an unclipped weight. Following DAPO~\citep{dapo}, we also drop the reference-KL term and apply dynamic sampling, skipping all-success or all-failure groups and gathering $128$ mixed trajectories ($16$ groups) before launching training of that step.

\textbf{RAFT++ as a baseline.} As a contrasting off-policy baseline we also consider RAFT++~\citep{raft++}, which can be viewed as vanilla multi-step GRPO with the same per-trajectory normalizer $1/|\tau_i|$, but with group normalization disabled and the group-relative advantage replaced by a success filter ($r>0$). Only successful trajectories contribute gradient, so RAFT++ effectively performs behavior cloning on a rolling buffer of positives and provides no contrastive signal on below-average trajectories. Decoupled importance sampling is still applied to keep the off-policy gradient unbiased.

\section{Experiments}

\subsection{Setup} \label{sec:exp-setup}

\textbf{Environment.} We train and test under the protocol from WebGym~\citep{webgym}. We picked \prev{} because it is the largest open multi-step visual web-agent training environment to date: roughly 290k training tasks across 128k real-world websites in three difficulty levels (Easy, Medium, Hard), evaluated on a 1,167 task OOD test split whose websites do not appear in training. The scale of the OOD test split (1{,}167 tasks across held-out websites) is what makes single-seed comparisons informative here. \prev{}'s own analysis reports that two runs differing in early-training trajectory dynamics nonetheless converge to overlapping final test performance, with error bars across runs small enough that single-seed comparisons are informative (\citealp{webgym}, Figure 8). We inherit this property and report single-seed results throughout, focusing reporting budget on algorithm-level ablations rather than seed sweeps. Rewards are binary, produced by WebGym's GPT-4o rubric evaluator. Following WebGym, we use the coordinate-based action space $\{\texttt{click}, \texttt{type}, \texttt{scroll}, \texttt{go\_back}, \texttt{navigate}, \texttt{ANSWER}\}$ on raw screenshots, with per-difficulty horizons of $(10, 20, 30)$ steps. The Easy-difficulty horizon $k{=}10$ is the WebGym recommended trade-off between truncation and rollout cost, and our constant $1/k$ fix uses $k{=}10$ throughout. All hyperparameter details are in \Cref{app:hparams}.

\textbf{Model Variants.} We use two Qwen3-VL-8B variants: Instruct (Qwen3-VL-8B-Instruct) and Thinking (Qwen3-VL-8B-Thinking), both with the additive memory prompt template shown in \Cref{app:prompts}. At each step the policy sees only the previous screenshot and its previous response. For the Thinking variant the previous step's thinking tokens are hidden from history, following standard practice~\citep{webgym}.

\textbf{Comparisons.} REINFORCE style on-policy algorithms present a type mismatch for an async framework that is not on-policy by design. We therefore compare to the closest off-policy baseline, RAFT++ (defined in \Cref{sec:method-algorithm}). We run three configurations:

\begin{itemize}[leftmargin=*,itemsep=2pt,topsep=2pt]
\item \textcolor{red!70!black}{\textbf{\prev{}}}: prior synchronous RL pipeline with the Filtered BC objective of \citet{webgym}, the published WebGym numbers.
\item \textcolor{blue!70!black}{\textbf{\ours{}-RAFT++}}: our async framework with RAFT++ as the loss, the async compatible substitute for WebGym's sync REINFORCE.
\item \textcolor{mypurple}{\textbf{\ours{} (full)}}: our async framework with the full method from \Cref{sec:method}, namely multi-step GRPO with decoupled importance sampling and the constant $1/k$ step-number normalizer.
\end{itemize}

\subsection{Main Results} \label{sec:exp-main}

\begin{table}[t]
\centering
\caption{\small Peak test set success rate (\%) on the \prev{}~\citep{webgym} OOD test split. Best per column within each variant in bold.}
\label{tab:main-results-summary}
\small
\setlength{\tabcolsep}{4pt}
\begin{tabular}{@{}l c c c c@{}}
\toprule
Algorithm & Easy & Med. & Hard & Avg \\
\midrule
\multicolumn{5}{@{}l}{\textit{\texttt{Model: Qwen3-VL-8B-Instruct}}} \\
\midrule
Base (no RL)                                             & 32.5          & 11.2          & 0.0          & 26.2          \\
\textcolor{red!70!black}{\prev{}} (sync REINFORCE)       & 50.9          & 24.1          & 4.8          & 42.9          \\
\textcolor{blue!70!black}{\ours{}-RAFT++}                & 46.6          & 27.8          & 5.5          & 39.3          \\
\textcolor{mypurple}{\textbf{\ours{} (full)}}            & \textbf{52.4} & \textbf{34.3} & \textbf{7.1} & \textbf{45.4} \\
\midrule
\multicolumn{5}{@{}l}{\textit{\texttt{Model: Qwen3-VL-8B-Thinking}}} \\
\midrule
Base (no RL)                                             & 37.4          & 24.3          & 1.2           & 32.0          \\
\textcolor{blue!70!black}{\ours{}-RAFT++}                & 47.3          & 30.0          & 5.2           & 40.5          \\
\textcolor{mypurple}{\textbf{\ours{} (full)}}            & \textbf{51.8} & \textbf{35.1} & \textbf{11.3} & \textbf{44.4} \\
\bottomrule
\end{tabular}
\end{table}

\begin{figure*}[t]
    \centering
    \includegraphics[width=\linewidth]{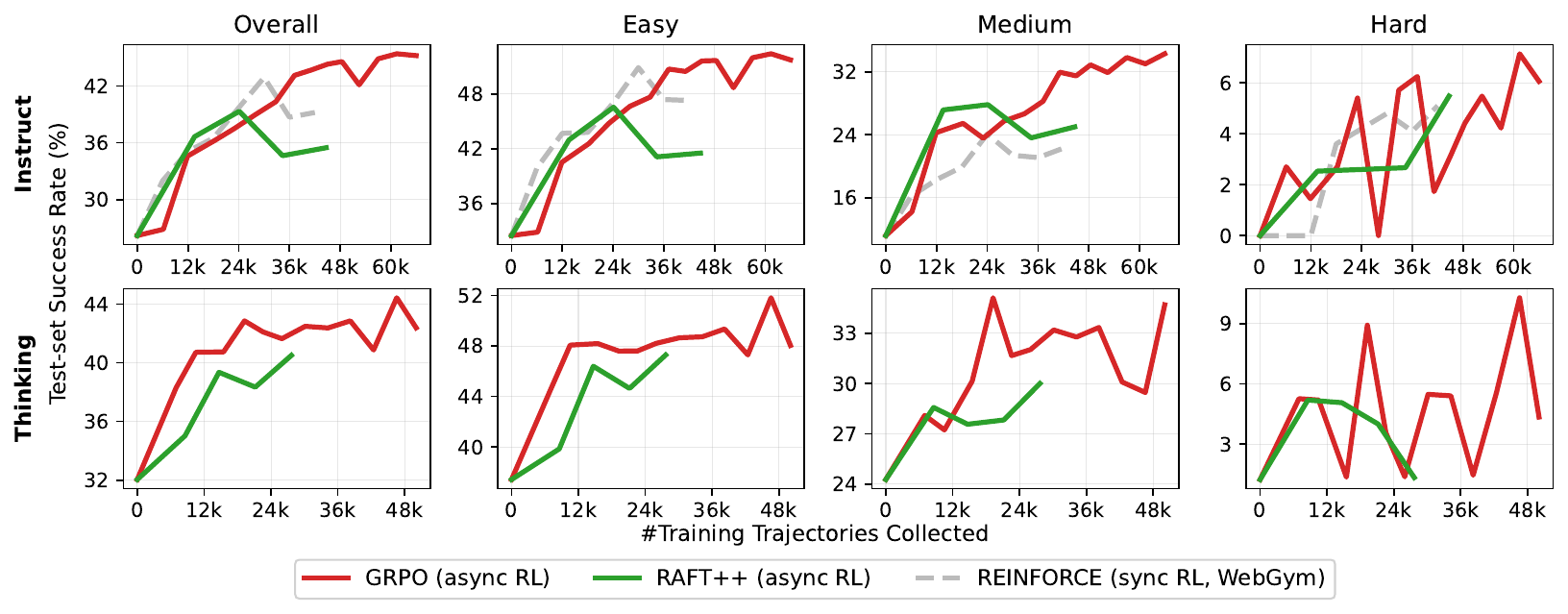}
    \caption{\small Test success rate vs.\ training trajectories collected on the WebGym OOD test split. Solid colored curves are our runs under \ours{}: \textcolor{mypurple}{\ours{} (full)} and \textcolor{blue!70!black}{\ours{}-RAFT++}. The gray dashed curve is the prior \prev{} sync REINFORCE baseline (values from \citet{webgym}). \textit{Top:} Instruct. \textit{Bottom:} Thinking. No \prev{} curve is shown on Thinking because its baseline was not trained under our $(10,20,30)$ per difficulty horizons. \ours{} (full) achieves both higher final test reward and higher sample efficiency per training trajectory collected.}
    \label{fig:main-results}
\end{figure*}

\textbf{\ours{} (full) sets a new SOTA over \prev{}, with the largest relative gains on harder slices.} On Instruct, \ours{} (full) reaches $45.4\%$ Avg against \prev{}'s $42.9\%$, a $\bm{+5.8\%}$ \textbf{relative} improvement (\Cref{tab:main-results-summary}), but Avg understates where the gain lives: \textbf{Easy} barely moves ($50.9 \to 52.4$, $+2.9\%$ relative), \textbf{Medium} widens to $\bm{+42\%}$ \textbf{relative} ($24.1 \to 34.3$), and \textbf{Hard} reaches $\bm{+48\%}$ \textbf{relative} ($4.8 \to 7.1$). The same harder-slice pattern shows up on Thinking in \Cref{fig:main-results} (bottom row), where \prev{}'s reported numbers are not available at our training horizons. Medium and Hard are the slices on which prior REINFORCE-style baselines have the most room to grow, because successful trajectories there are rarer and a behavior-cloning loss on a slowly changing positives buffer cannot push down the dominant failure modes. GRPO's group-normalized advantage does (\Cref{app:grpo-vs-raft}, \Cref{app:raft-lr}).

\begin{figure}[t]
\centering
\includegraphics[width=\linewidth]{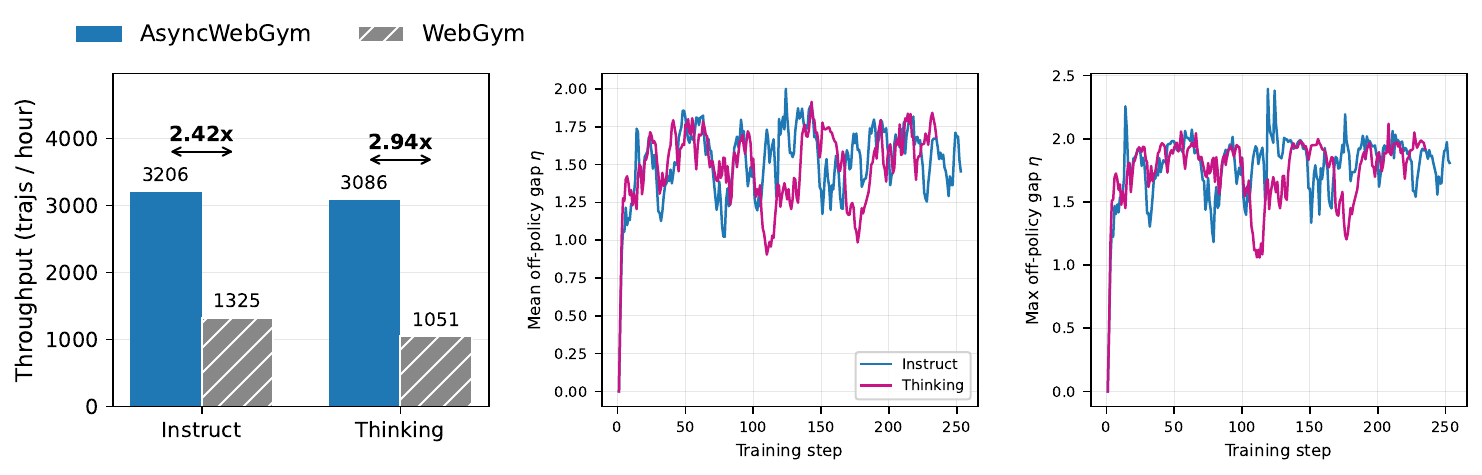}
\caption{\small \textbf{Left:} End to end training trajectory throughput. \textbf{Right:} Off-policyness during GRPO training for the Instruct (blue) and Thinking (red) Qwen3-VL-8B runs: mean and max of the per-token off-policy gap $g$ across a training batch.}
\label{fig:throughput-offpolicy}
\end{figure}

\textbf{Reproducing the \prev{} sync REINFORCE pipeline under async.} Before comparing algorithms, we use \ours{}-RAFT++ as a like-for-like check against the published \prev{} sync REINFORCE numbers (same loss family, async framework substituted for sync). \ours{}-RAFT++ on Instruct reaches 39.3\%, against 42.9\% for the prior \prev{} pipeline (\Cref{tab:main-results-summary}). The $3.6\%$ gap is consistent with the importance sampling overhead any async framework has to pay, and the two methods land in the same neighborhood on the same loss family, with the gain reported above being net of this cost. Wall-clock cost moves the other way: \ours{} produces about 3,100 training trajectories per hour on both variants against approximately 1,300 and 1,050 trajectories per hour for the synchronous \prev{} baseline, a $\bm{2.4}$\textbf{ to }$\bm{2.9\times}$\textbf{ end-to-end speedup} (\Cref{fig:throughput-offpolicy}, left). The speedup also incorporates smaller engineering improvements over \prev{}, including the distributed inference server and in-memory weight broadcast. Since \prev{}~\citep{webgym} itself reports the fastest open-source rollout system for multi-step visual web agents to date, the $2.4$ to $2.9\times$ end-to-end speedup on top of it makes \ours{}, to our knowledge, \textit{the fastest reported open multi-step RL framework for visual web agents}. We qualify with ``reported'' because among open multi-step VLM RL frameworks in this category, only \prev{} publishes throughput numbers.

\textbf{Off-policyness stays small.} A natural worry under fully async RL is that the off-policy gap might be large. AReaL~\citep{areal}, the closest async text reasoning analog, reports running with a max staleness of $\eta{=}4$ for coding and $\eta{=}8$ for math. In our setting, with the max staleness set to $\eta=2$, the mean per-token off-policy gap $\eta$ stays near 1.5 and the max near 2.0 throughout training (\Cref{fig:throughput-offpolicy}, right two panels), well below the cap, and GPUs stay busy throughout. Off-policyness is naturally smaller than in coding and math because web-agent responses are much shorter~\citep{tti}, giving a higher rollout/train speed ratio and leaving few stale trajectories in the training batch.

\subsection{Efficiency} \label{sec:exp-efficiency}

\begin{figure*}[t]
\centering
\includegraphics[width=\linewidth]{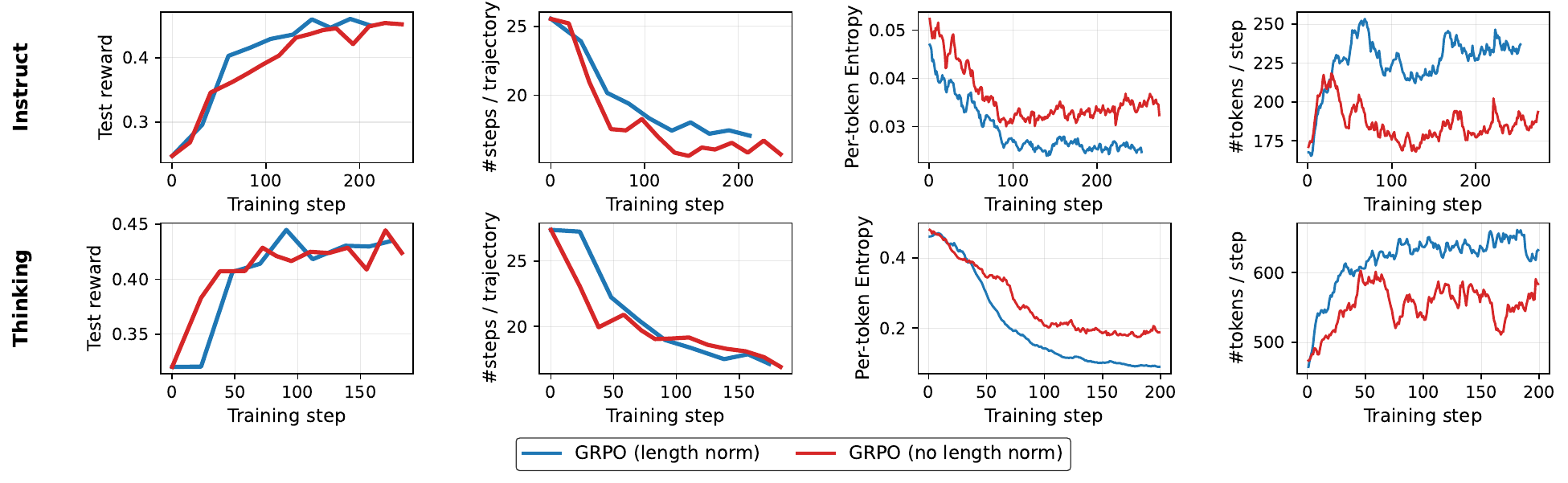}
\caption{\small Effect of the $1/|\tau_i|$ normalizer on GRPO training dynamics. Rows are the two Qwen3-VL-8B variants (top: Instruct, bottom: Thinking), columns are, from left to right, test reward, \#steps per trajectory, per-token entropy, and \#tokens per step. Test reward is essentially tied between the two losses, but the $1/|\tau_i|$ run produces longer trajectories, longer per-step responses, and lower per-token entropy.}
\label{fig:length-norm-comparison}
\end{figure*}

\textbf{Removing $1/|\tau_i|$ preserves performance and shortens trajectories.} The constant $1/k$ replacement does not change what the policy can solve: test reward is essentially tied throughout training on both variants (\Cref{fig:length-norm-comparison}, first column). What changes is trajectory and per-step response length. Under the standard $1/|\tau_i|$ normalizer the average trajectory is consistently longer, and the per-step response grows to around $240$ tokens; the constant $1/k$ run uses fewer steps per trajectory and fewer tokens per step at matched test reward. Measured within each run (same hardware throughout), the trajectory contraction translates into a per-step gradient-update time reduction of $\bm{11}$--$\bm{15\%}$ from the first $20$ to the last $20$ training steps under the constant $1/k$ fix, against only $3$--$5\%$ under the standard $1/|\tau_i|$ loss; the corresponding total per-step wall-clock reductions are $\bm{18}$--$\bm{19\%}$ vs.\ $4$--$5\%$ (full table in \Cref{app:train-step-contraction}). The mechanism behind this, length-coupled memory drift, is analyzed in \Cref{sec:memory-drift}.

\begin{figure}[t]
    \centering
    \includegraphics[width=\linewidth]{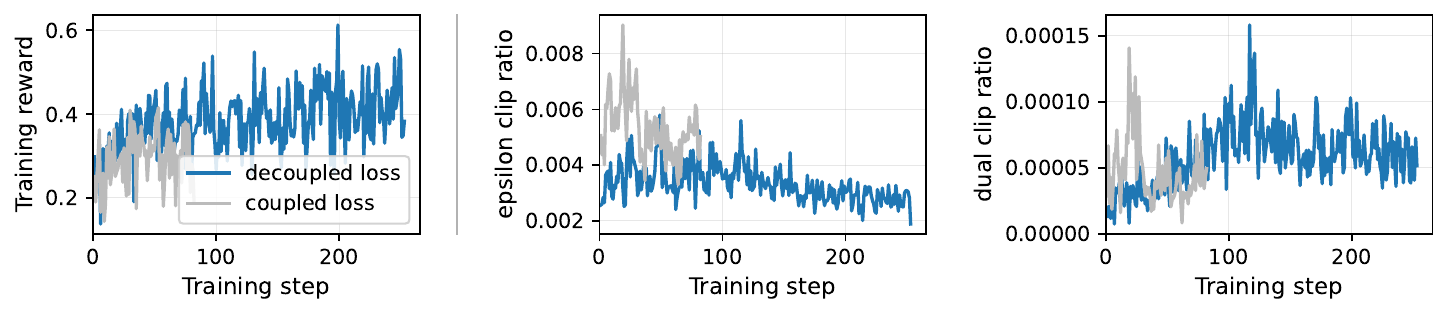}
    \caption{\small Coupled vs.\ decoupled importance sampling under the async RL GRPO loss. From left to right: per update mean of training reward, fraction of tokens hit by the $\epsilon$-clip, and fraction of tokens hit by the dual clip.}
    \label{fig:coupled-loss-clip}
\end{figure}

\textbf{Decoupled importance sampling halves the clip rate.} A separate ablation isolates the off-policy correction described in \Cref{sec:method-algorithm}. We run a coupled loss baseline that folds rollout staleness and current-update movement into a single $\pi_\theta/\pi_{\text{behave}}$ ratio with PPO-style clipping centered on the coupled ratio. The coupled run sits at roughly $2\times$ the $\epsilon$-clip-trigger rate of the decoupled run at start and improves training reward substantially more slowly (\Cref{fig:coupled-loss-clip}). We therefore keep the decoupled factorization in every \ours{} configuration above.

\section{Dynamics Analysis} \label{sec:memory-drift}

\subsection{Main Analysis}

The $1/|\tau_i|$ factor in multi-step GRPO is the step-level analog of the token-level normalizer Dr.~GRPO~\citep{dr.grpo} identifies as biased: each trajectory enters the loss with total weight $1$ regardless of length, so any token in a long trajectory carries a $1/|\tau_i|$ share. In a setting like ours where failure is dominated by horizon exhaustion rather than clearly wrong actions, this attenuates the penalty on exactly the trajectories the policy needs to learn to avoid. On the base model, only around $30\%$ to $40\%$ of rollouts succeed in the DAPO-filtered training batch, and failures average $12.5$ steps against $5.1$ for successes (a $2.4\times$ gap) when RL starts. The short successful population is both rarer and shorter and cannot counterbalance the effect of long failed trajectories, so early in training, the policy is pulled toward longer rollouts (\Cref{fig:length-norm-comparison}, second column).

\begin{figure*}[t]
\centering
\includegraphics[width=\linewidth]{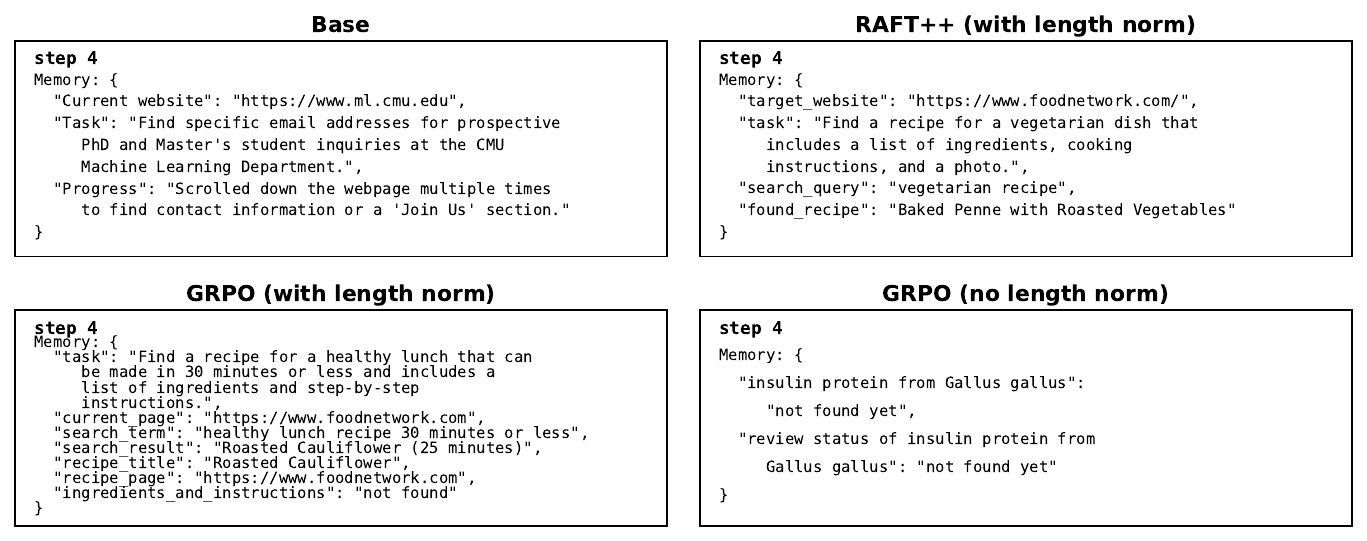}
\caption{\small \texttt{Memory} at step $4$ of one representative 10/20/30-horizon rollout per checkpoint. $1/|\tau_i|$ accumulates verbose generic keys; the constant $1/k$ fix re-keys \texttt{Memory} to task sub-questions.}
\label{fig:memory-evolution}
\end{figure*}

\begin{figure*}[t]
\centering
\includegraphics[width=\linewidth]{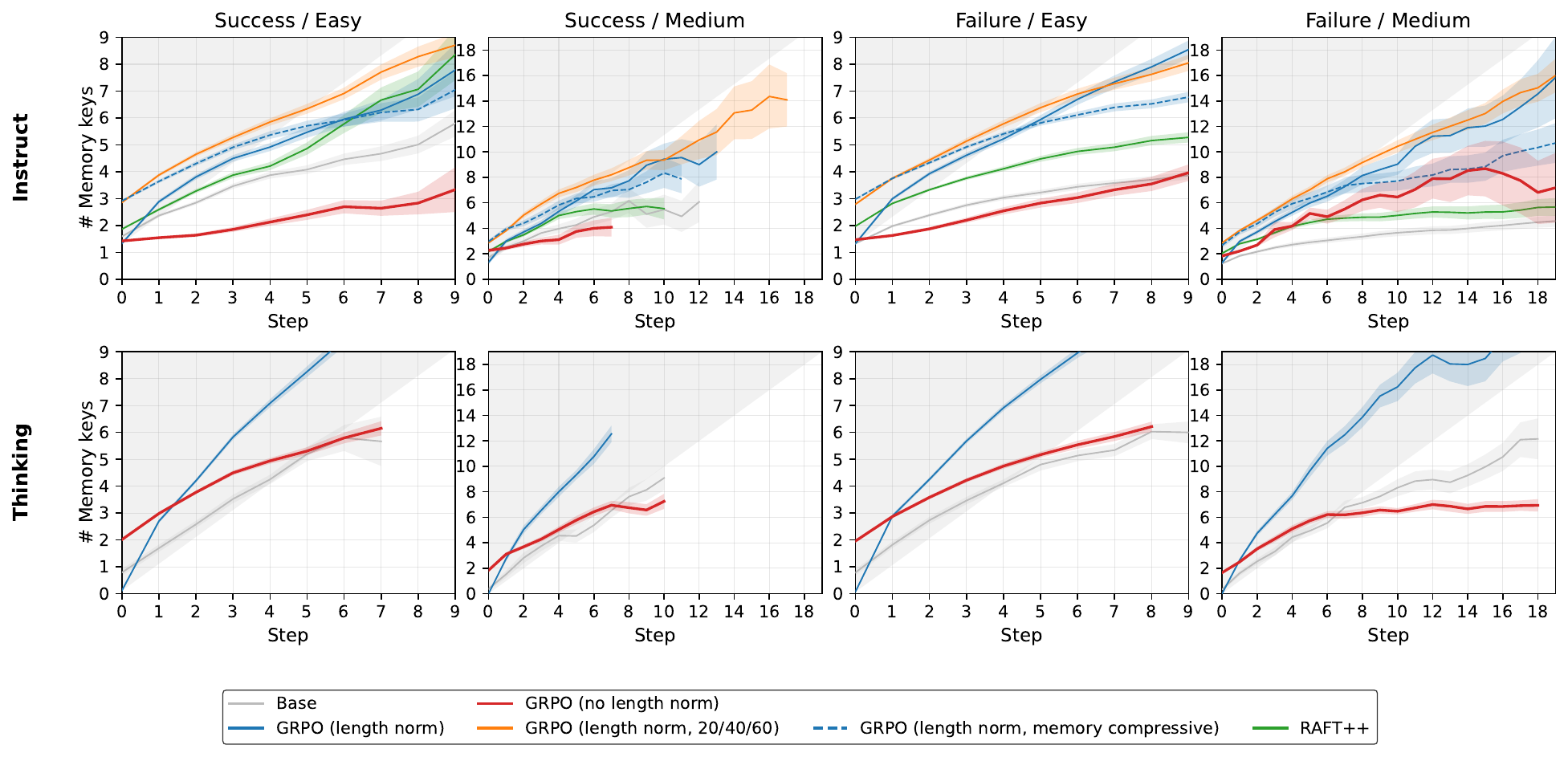}
\caption{\small Trajectory-mean number of \texttt{Memory} JSON keys per agent step, split by outcome (Success, Failure) and difficulty (Easy, Medium). $1/|\tau_i|$ tracks the one-new-key-per-step diagonal; the constant $1/k$ fix stays close to Base.}
\label{fig:response-algorithm-curve}
\end{figure*}

This lengthening of trajectories then propagates into longer per-step responses through the prompt structure. Under the WebGym additive-memory prompt against a multi-step visual environment~\citep{webgym}, each step appends to \texttt{Memory}, and each new screenshot exposes new information the policy can legitimately log, so \texttt{Memory} grows roughly linearly in step count. Under $1/|\tau_i|$ the per-step \texttt{Memory} key count tracks the one-new-key-per-step diagonal, while under the constant $1/k$ fix it stays close to the base-model baseline (\Cref{fig:response-algorithm-curve}). Since \texttt{Memory} is part of the response, the larger step count drags per-step response length up by $\approx 33\%$ (\Cref{fig:length-norm-comparison}, fourth column), and the entropy drop (third column) follows because the extra response length is mostly low-entropy \texttt{Memory} boilerplate (per-token entropy annotations on representative responses in \Cref{sec:token-entropy-length-normalized}). The schema itself also drifts: under $1/|\tau_i|$ the typical trajectory rewrites \texttt{Memory} every step with generic placeholders like \texttt{task\_1} or \texttt{current\_step} ($34\%$ of all observed key-occurrences are generic-slot, only $7\%$ of trajectories preserve their key set end-to-end, only $36\%$ of step pairs leave the \texttt{Memory} key set unchanged), while under the constant $1/k$ fix the policy declares a small set of task-anchored keys at step $0$ and holds them throughout ($65\%$, $76\%$, generic slots drop to $11\%$, \Cref{tab:no-ln-schema} and \Cref{fig:memory-evolution}; token-level last-step responses in \Cref{sec:last-step-responses-instruct,sec:last-step-responses-thinking}). The constant $1/k$ fix restores full per-token weight on long failures, and every downstream symptom above (memory bloat, longer responses, lower entropy, generic schemas) disappears.

\begin{table}[t]
\centering
\small
\setlength{\tabcolsep}{4pt}
\begin{tabular}{l c c}
\toprule
                                          & GRPO   & No LN \\
\midrule
keys$_0=$keys$_\text{last}$               & $7\%$  & $\mathbf{65\%}$ \\
No-edit step pairs                        & $36\%$ & $\mathbf{76\%}$ \\
Generic-slot keys                         & $34\%$ & $\mathbf{11\%}$  \\
\bottomrule
\end{tabular}
\caption{\small Schema signatures of \texttt{Memory} under GRPO with $1/|\tau_i|$ vs.\ the constant $1/k$ fix (Instruct, additive prompt, first 30 steps per trajectory). Rows: fraction of trajectories with first-step and last-step key sets equal; fraction of consecutive step pairs with the same key set; fraction of key occurrences matching a generic placeholder regex (e.g.\ \texttt{task}, \texttt{current\_step}, \texttt{search\_query}).}
\label{tab:no-ln-schema}
\end{table}

\subsection{Ablations}

We run three ablations to confirm that the diagnosis above identifies the right cause. The first checks that a closely related algorithm (RAFT++) exhibits the same pathology when it shares the same loss factor, ruling out group-relative advantage as the driver. The second checks that the cause lives in the loss rather than the prompt, by trying to fix the symptom from the prompt side and observing that it does not work. The third checks that the effect scales as the mechanism predicts when we make long failures even longer.

\textbf{RAFT++ exhibits the same pathology, more mildly.} RAFT++ also normalizes per-trajectory loss by $1/|\tau_i|$, and the diagnosis predicts the same memory bloat in kind. \Cref{fig:response-algorithm-curve} (Instruct row) confirms it: the RAFT++ per-step \texttt{Memory} key-count curve drifts upward in the same direction as GRPO with $1/|\tau_i|$, closely shadowing the GRPO length-norm curve on Easy successful trajectories and sitting well above the constant-$1/k$ run. The effect is muted on failure trajectories, consistent with RAFT++'s success-conditioned filter ($r>0$) providing no direct gradient on failed rollouts: the same length-coupled attenuation is present in the loss but only propagates through the kept-positives. The matched per-step Add/Del/Mod edit-op decomposition (\Cref{app:edit-ops}) tells the same story at a smaller magnitude than full GRPO. RAFT++ and GRPO differ in nearly every other respect, yet both show the same length-coupled drift whenever they share the $1/|\tau_i|$ factor. The factor itself, not the surrounding algorithm, is the driver.

\textbf{Prompt-level intervention fails.} If the loss is what drives the pathology, not the prompt, then intervening at the prompt level should not help. Swapping the additive-memory prompt for a compressive one that explicitly instructs the model to compress \texttt{Memory} at every step, while keeping $1/|\tau_i|$ in the loss, leaves per-step Add/Del/Mod rates elevated relative to the Base run (\Cref{tab:edit-ops}, ``GRPO, compressive''). The prompt change does not reach the cause.

\textbf{Scaled horizons amplify the effect.} The mechanism predicts that anything making long failures even longer should make the symptoms worse. Scaling per-difficulty horizons from 10/20/30 to 20/40/60 while keeping $1/|\tau_i|$ does exactly that, and we see the predicted amplification in \Cref{fig:response-algorithm-curve}: the \emph{GRPO (length norm, 20/40/60)} curve tracks the one-new-key-per-step diagonal across both Easy and Medium panels, exceeding the default $10/20/30$ length-norm run.

\section{Conclusion}

We presented \ours{}, a fully asynchronous multi-step RL framework for visual web agents. An everlasting rollout pool and lightweight screenshot handling give a $2.4$--$2.9\times$ end-to-end speedup over the fastest prior open pipeline; replacing $1/|\tau_i|$ with a constant $1/k$ then contracts trajectories at matched reward for a further $1.80\times$ per-step speedup. A decoupled importance-sampling correction sets a new open SOTA on \prev{}'s OOD split ($45.4\%$ vs.\ $42.9\%$).

\section*{Acknowledgements}

We thank Nan Jiang, Anikait Singh, and Aravind Rajeswaran for their useful discussions on algorithm designs and presentation. This work is partially supported by NSF under Grant No.\ 2416897, Grant No.\ 2505932, and by ONR under Grant No.\ N000142512318. This research used both Delta (NSF award OAC 2005572) and DeltaAI (NSF award OAC 2320345) advanced computing systems, and computing resources provided by Illinois Computes and NAIRR Pilot NAIRR250157.

\bibliography{main}

\newpage
\appendix

\section{Additional Analyses} \label{sec:ablations}

This appendix collects supporting analyses for the main-text results. \Cref{app:raft-lr} reports a learning-rate ablation on RAFT++. \Cref{app:grpo-vs-raft} unpacks the GRPO-vs-RAFT++ gap with entropy and clip-rate diagnostics. \Cref{app:edit-ops} provides the per-step \texttt{Memory} Add/Del/Mod table referenced by the \Cref{sec:memory-drift} ablations. \Cref{app:batch-size} checks batch-size sensitivity.

\subsection{RAFT++ Learning Rate} \label{app:raft-lr}

\begin{figure*}[t]
    \centering
    \includegraphics[width=\linewidth]{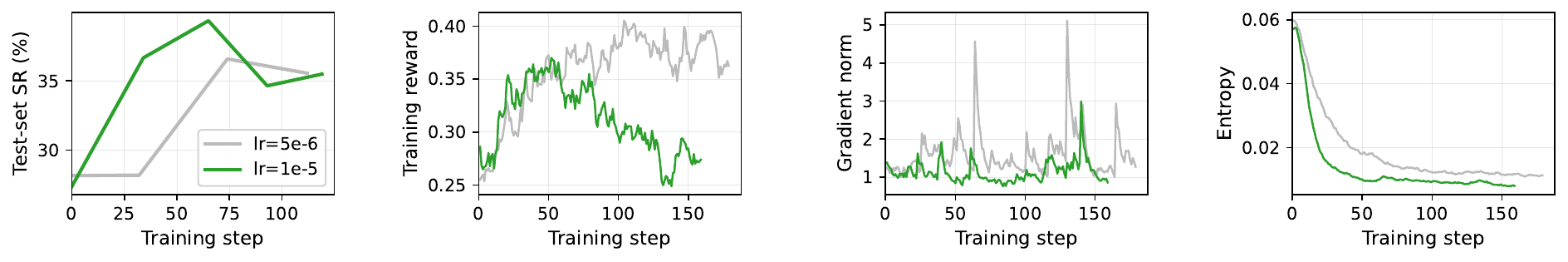}
    \caption{\small Learning-rate ablation on RAFT++
    (\texttt{Qwen3-VL-8B-Instruct}, $b{=}120$, off-policy~$=$~2, no~KL).
    From left to right: held-out test success rate (the larger-LR run
    wins by a wide margin at the peak), training reward, where the trend
    reverses, pre-clipping gradient L2 norm (consistently bounded for
    the larger LR), and per-token policy entropy, which decays faster
    under the larger LR. The train/test flip suggests low-LR RAFT++
    overfits the filtered positives buffer, noisier optimization
    generalizes better.}
    \label{fig:raft-lr}
\end{figure*}

Two RAFT++ runs on \texttt{Qwen3-VL-8B-Instruct} that share dataset, batch size ($b{=}120$), and every other hyperparameter except the learning rate ($5{\times}10^{-6}$ vs.\ $1{\times}10^{-5}$) move in \emph{opposite} directions on training and held-out reward (\Cref{fig:raft-lr}): the smaller-LR run reaches a higher peak training reward but a markedly lower peak test success rate; the larger-LR run gives up some training fit and gains substantially on the test set. We do not have a controlled experiment to attribute the flip causally, but the pattern is consistent with the RAFT++ objective behaving like behavior cloning on a slowly-changing positives buffer, where many small gradient steps can over-fit the kept successful trajectories. This observation motivates our choice of the higher learning rate ($1\times 10^{-5}$) for the Thinking RAFT++ baseline reported in \Cref{tab:hyperparameters}.

\subsection{Why GRPO Outperforms RAFT++} \label{app:grpo-vs-raft}

\begin{figure*}[t]
    \centering
    \includegraphics[width=\linewidth]{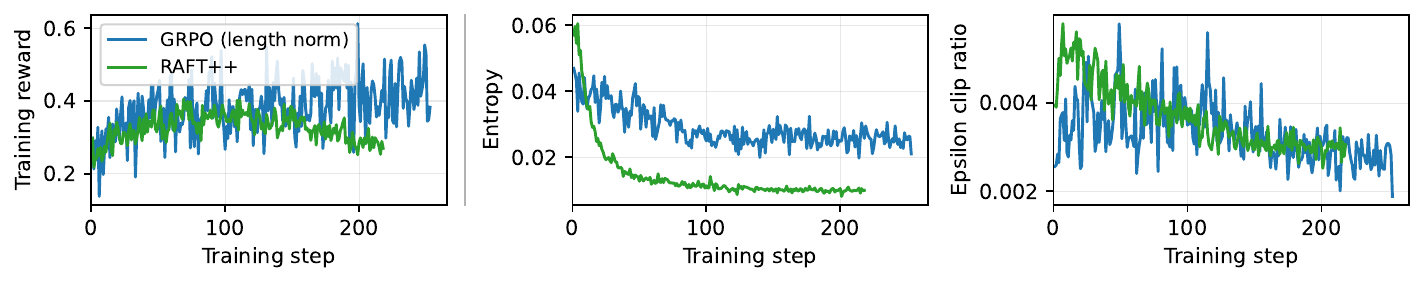}
    \caption{\small GRPO vs.\ RAFT++ on \texttt{Qwen3-VL-8B-Instruct}.
    \textit{Left:} training reward per optimizer step. \textit{Middle:}
    per-token policy entropy averaged over the loss mask, the step-zero
    offset is induced by the conditional vs.\ unconditional denominator
    (see text). \textit{Right:} fraction of tokens hitting the
    $\epsilon$-clip, both runs sit near 0.3\%, ruling out clip
    mechanics as the driver of the entropy gap.}
    \label{fig:grpo-vs-raft}
\end{figure*}

RAFT++ falls short of GRPO on both variants. We attribute this primarily to RAFT++ providing no contrastive signal on below-average rollouts: its $r>0$ filter only pulls mass \emph{onto} successful trajectories, while GRPO's group-normalized advantage simultaneously pushes mass \emph{down} on below-average trajectories within the same group. The entropy gap on Instruct in \Cref{fig:grpo-vs-raft} is consistent with this picture: RAFT++'s entropy collapses from $0.057$ to $0.010$ while GRPO's settles near $0.025$, and $\epsilon$-clip rates are essentially identical ($\approx 0.3\%$ in both runs), so the gap is not driven by clipping mechanics. We do not run a controlled ablation that isolates the contrastive signal from other differences between the two losses.

\subsection{Memory Edit-Op Rates} \label{app:edit-ops}

\begin{table*}[t]
\centering
\caption{\small Per-step edit operations on \texttt{Memory} (instruct prompt, first 30 steps), split by trajectory outcome. Add / Del / Mod count per consecutive step pair the inserted / removed / value-changed keys.}
\label{tab:edit-ops}
\small
\begin{tabular}{l l l c c c c c c}
\toprule
           &            &              & \multicolumn{3}{c}{Success} & \multicolumn{3}{c}{Failure} \\
\cmidrule(lr){4-6} \cmidrule(lr){7-9}
Algorithm  & Prompt     & Step-Num. Norm? & Add  & Del  & Mod                                          & Add  & Del  & Mod  \\
\midrule
Base & additive & N/A & 0.65 & 0.08 & 0.41 & 0.22 & 0.02 & 0.23 \\
RAFT++ & additive & Yes & 0.64 & 0.08 & 0.47 & 0.44 & 0.06 & 0.43 \\
GRPO & additive & Yes & 1.11 & 0.26 & 0.83 & 1.13 & 0.31 & 1.10 \\
GRPO & compressive & Yes & 0.84 & 0.28 & 1.14 & 0.71 & 0.25 & 1.46 \\
\textbf{GRPO} & \textbf{additive} & \textbf{\textcolor{red}{No}} & \textbf{0.24} & \textbf{0.04} & \textbf{0.43} & \textbf{0.39} & \textbf{0.06} & \textbf{0.34} \\
\bottomrule
\end{tabular}
\end{table*}

\Cref{tab:edit-ops} reports per-step Add/Del/Mod edit-op rates on the agent's \texttt{Memory} field, split by trajectory outcome, for the algorithm-prompt-loss combinations studied in \Cref{sec:memory-drift}. The matched per-step Memory key-count curves are plotted in \Cref{fig:response-algorithm-curve}; this table reports the operation-level decomposition, and tells four qualitative stories. RAFT++ edits \texttt{Memory} more aggressively than Base, consistent with the shared $1/|\tau_i|$ factor pushing the policy toward longer memory schemas. GRPO with $1/|\tau_i|$ edits even more aggressively than RAFT++, because the group-relative advantage adds direct negative gradient on below-average trajectories that RAFT++'s success-conditioned filter does not provide. The compressive prompt under $1/|\tau_i|$ brings rates down from full additive GRPO but stays above Base, indicating that prompt-level intervention only mitigates the loss-driven effect rather than removing it. Finally, replacing $1/|\tau_i|$ with the constant $1/k$, even using the additive memory prompt, brings the rates back to roughly Base levels (and below on Success), confirming that the loss aggregation is the root cause analyzed in \Cref{sec:memory-drift}.

\subsection{Batch Size Sensitivity} \label{app:batch-size}

\begin{figure*}[t]
    \centering
    \includegraphics[width=\linewidth]{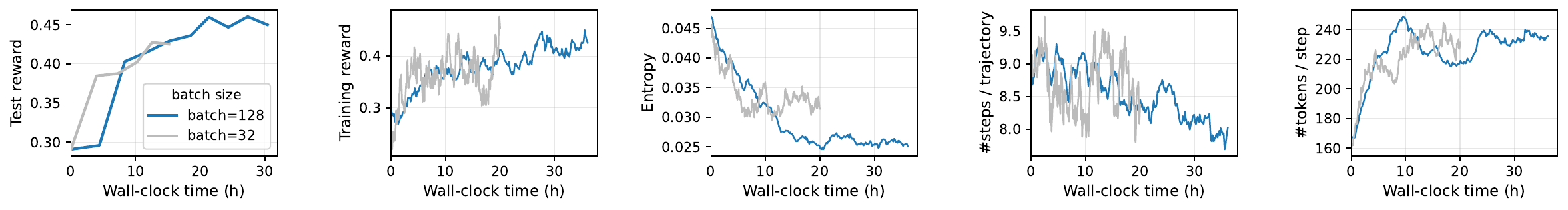}
    \caption{\small Effect of consumer batch size on GRPO Instruct training, plotted against wall-clock hours: batch=128 (the canonical setting used throughout the paper) versus batch=32, all other hyperparameters held fixed. From left: test reward, training reward, policy entropy, average steps per trajectory, and average response tokens per step.}
    \label{fig:batch-size-comparison}
\end{figure*}

We chose a consumer batch size of $128$ for the canonical Instruct GRPO run (\Cref{app:hparams}). To check whether our conclusions depend on this choice, we re-ran the same configuration with batch size $32$, holding learning rate, group size, off-policy lag cap, and all other hyperparameters fixed (\Cref{fig:batch-size-comparison}). The small-batch run improves faster early (reaching $\approx 0.38$ test reward several hours ahead of batch-128, since a four-times-higher optimizer-update rate cashes in when the gradient signal is large) but intersects with the batch-128 run around 10 hours (reaching $\approx 0.40$), so over the full training horizon the choice is essentially neutral on test reward. The early gain comes with visibly higher per-iteration variance: batch-32 training reward, entropy, and steps per trajectory all swing within a wider envelope than batch-128 at every wall-clock time. Crucially, the qualitative behaviors analyzed in \Cref{sec:memory-drift} (entropy collapse, falling steps per trajectory, rising tokens per step) still reproduce at batch size $32$, so the length-coupled memory drift mechanism is a property of the loss shape rather than of the batch-size choice. The canonical batch size of $128$ is therefore a sensible default, but the main qualitative claims of the paper do not depend on it.

\subsection{Within-Run Per-Step Time Contraction} \label{app:train-step-contraction}

To quantify per-step compute savings from the constant $1/k$ fix without cross-run hardware confounds, we report \emph{within-run} ratios: the average time per training step over the last $20$ steps divided by the average over the first $20$ (\Cref{tab:within-run-time-contraction}). Each run executes on the same hardware throughout, so any change reflects the policy contracting (or expanding) its trajectories over training. We report two measurements. The first is the gradient-update time only (forward, backward, optimizer), which the trainer records directly for every training step. The second is the total time elapsed per training step, also including rollout collection, evaluation, and checkpointing. For the total time, we skip a small number of initial steps whose recorded timing reflects setup or resume overhead rather than steady-state training; for one of the Thinking $1/k$ fragments this means the first window starts a few dozen training steps in.

\begin{table}[h]
\centering
\caption{Within-run ratio of last-20 to first-20 per-step time. Lower is more contraction.}
\label{tab:within-run-time-contraction}
\small
\begin{tabular}{lcc}
\toprule
Run & Gradient & Total \\
\midrule
Instruct, $1/|\tau_i|$ & $0.97$ & $0.96$ \\
Instruct, $1/k$ & $\bm{0.85}$ & $\bm{0.81}$ \\
Thinking, $1/|\tau_i|$ & $0.95$ & $0.95$ \\
Thinking, $1/k$ & $\bm{0.89}$ & $\bm{0.82}$ \\
\bottomrule
\end{tabular}
\end{table}

Both measurements agree. Under the standard $1/|\tau_i|$ loss, per-step time is essentially flat over the run ($0.95$--$0.97$): the policy's trajectory-length distribution drifts slowly enough that per-step cost stays roughly constant. Under the constant $1/k$ fix, per-step time contracts noticeably within a single run ($0.81$--$0.89$, an $11$--$19\%$ reduction from start to end of training). This is the within-run signature of the trajectory contraction analyzed in \Cref{sec:memory-drift}: as training proceeds, the $1/k$ policy emits shorter trajectories and fewer tokens per step, so each gradient update consumes less compute.

\section{Implementation Details} \label{app:impl-details}

This appendix documents the system, loss, prompt, and hyperparameter details needed to reproduce our runs. The ``System'' subsection below describes the lightweight screenshot handling promised in \Cref{sec:method-system}. \Cref{app:extra-optimization} defines the dual-clip variant used by every \ours{} GRPO run. \Cref{app:prompts} provides the full agent prompt template (Instruct and Thinking share it). \Cref{app:hparams} lists optimizer, loss, rollout, and hardware-allocation settings for all three main-table runs.

\subsection{System}

\begin{figure*}[t]
    \centering
    \includegraphics[width=\linewidth]{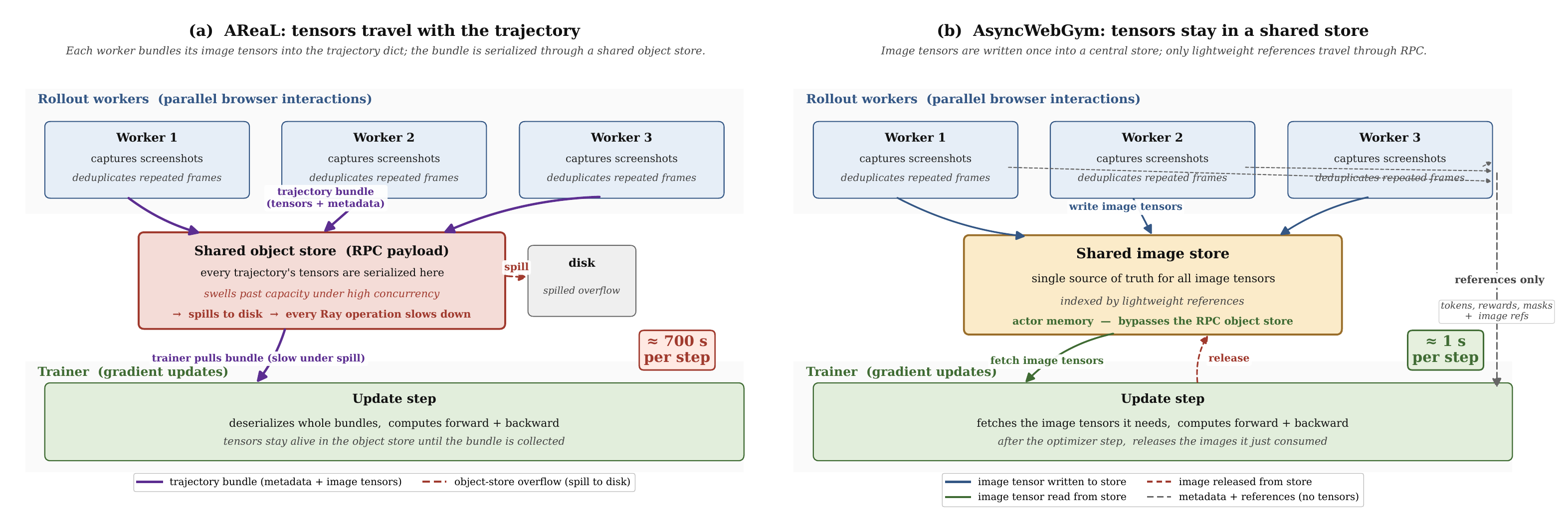}
    \caption{\small Lightweight Screenshot Handling. Compared to \prev{}, which serializes every screenshot through the shared RPC object store and spills to disk under concurrent rollouts, \ours{} keeps all image tensors in a dedicated in-memory actor and routes only lightweight \textit{references} through RPC, eliminating the per-step object-store bottleneck.}
    \label{fig:screenshot-management}
\end{figure*}

\textbf{Lightweight screenshot handling.} Different from \prev{}, which embeds raw image tensors directly into each trajectory dictionary and ships the entire bundle through the shared RPC object store, \ours{} stores all screenshots in a \textbf{dedicated in-memory actor} and routes only lightweight references between rollout workers and the trainer, as shown in~\Cref{fig:screenshot-management}. This redesign is forced by the multi-step web-agent setting: each trajectory carries tens of high-resolution screenshots, and at hundreds of concurrent rollouts the baseline pipeline pushes well over 100\,GiB of pixel tensors through the object store per training step. Once the object store exceeds its memory budget, the framework spills to disk and \emph{every} downstream operation, trajectory deserialization, RPC dispatch, and tensor localization, degrades by orders of magnitude. By keeping image tensors out of the RPC payload entirely, \ours{} confines the screenshot footprint to a single bounded actor, the trainer fetches only the slices it needs at gradient-update time and immediately releases them once the optimizer step completes.

\subsection{Optimization} \label{app:extra-optimization}

There are two considerations of algorithm design that we adopt in our experiments.

\textbf{Dual-clipped GRPO loss.} Let $f(\rho)\hat A_{i,t}$ denote the inner $\min$ argument of \Cref{equ:multi-step-grpo}. This advantage-based clipping framework bounds the surrogate objective when $\hat A_{i,t}\geq 0$, but fails to do so when $\hat A_{i,t}<0$ and $\rho\gg 1+\epsilon$: in this regime the inner $\min$ selects the unclipped term $\rho\hat A_{i,t}$, which is unbounded below in $\rho$. A single token assigned a low probability by the rollout policy can therefore drive the per-token contribution arbitrarily negative. Following the dual-clip extension of PPO~\citep{ppo-dual-clipping}, we cap $f(\rho)$ from below whenever the advantage is negative:
\begin{smalldisplay}
\begin{equation} \label{equ:dual-clip}
    f_\text{dual}(\rho)=\begin{cases}f(\rho), & \hat A_{i,t}\geq 0,\\[2pt] \max\!\left[f(\rho),\; c\cdot\hat A_{i,t}\right], & \hat A_{i,t}<0,\end{cases}
\end{equation}
\end{smalldisplay}
with constant $c>1$. The dual-clipped GRPO loss is obtained by substituting $f_\text{dual}(\rho)$ for $f(\rho)$ in~\Cref{equ:multi-step-grpo}, providing an absolute lower bound on the per-token contribution and eliminating the tail risk introduced by rare low-probability tokens with exploded importance sampling ratios.

\textbf{On not penalizing overlong trajectories specially.} A trajectory that exhausts its horizon is marked as a failure. DAPO~\citep{dapo} notes that this loses some potentially useful signal. We accept the loss: given the horizon, a trajectory that exhausted it means some of its actions were suboptimal, at least when another trajectory in the same group did succeed. Actions like \textit{goback} that appear more often in long trajectories are something we want the policy to take only when necessary, not on every long rollout. When every trajectory in the group fails by exhausting the horizon, either those \textit{goback}-style actions were genuinely unhelpful, or the horizon itself is too tight; either way, treating overlong rollouts as failures is the correct default.

\section{Qualitative Examples}

This appendix shows sample-level evidence for the length-coupled memory drift mechanism analyzed in \Cref{sec:memory-drift}. \Cref{sec:last-step-responses-instruct} and \Cref{sec:last-step-responses-thinking} contrast last-step responses produced by the $1/|\tau_i|$ run against the constant $1/k$ run on the two Qwen3-VL-8B variants, making the generic-slot memory schema versus task-anchored schema split visible at the token level. \Cref{sec:token-entropy-length-normalized} overlays the same responses with per-token entropy annotations, which lets the reader see directly that the extra tokens emitted under $1/|\tau_i|$ sit in low-entropy boilerplate regions rather than in decision-bearing positions.

\setlength{\parskip}{0pt}

\subsection{Instruct Last-Step Response} \label{sec:last-step-responses-instruct}

These examples illustrate the qualitative difference described in~\Cref{sec:memory-drift}.

\input{figures/vis_loss}

\subsection{Thinking Last-Step Response} \label{sec:last-step-responses-thinking}

\input{figures/vis_thinking}

\subsection{Token-level Entropy} \label{sec:token-entropy-length-normalized}

Each high-entropy token is annotated as \texttt{token ($P_{\mathrm{sampled}}$) [alt$_1$:P$_1$, alt$_2$:P$_2$, \dots]}, where $P_{\mathrm{sampled}}$ is the probability the policy assigned to the token it actually sampled and the bracket lists the top-$k$ alternative tokens with their probabilities.

Qualitative results below shows the per-token entropy of one rollout from the Qwen3-VL-8B-Instruct policy trained with the $1/|\tau_i|$ normalizer, on the first and last turn of the trajectory. Token background opacity is proportional to entropy, high-entropy tokens carry a subscript listing the chosen token's probability and the top alternatives. Mean entropy collapses from 0.0297 nats at the first turn to 0.0077 nats at the last, with the few remaining high-entropy positions concentrated on coordinate slots and free-form summary fields that must integrate fresh screenshot evidence.

\input{figures/viz_grid_body}

\subsection{Prompts} \label{app:prompts}

Both the Instruct and Thinking checkpoints share the same agent prompt
template, the only configurable variation is the single rule line that
controls whether \texttt{Memory} is updated additively (default) or
compressively (ablation). Both variants are shown inline below,
color-coded.

\begin{tcblisting}{title={Prompt: AsyncWebRL Agent},
  listing only, breakable,
  shrink break goal=0.5\baselineskip,
  left=2mm, right=2mm, top=1mm, bottom=1mm,
  listing options={style=promptroman, escapeinside={(*@}{@*)}}
}
--- MESSAGE 1: SYSTEM ---
You are a helpful assistant.

# Tools

You may call one or more functions to assist with the user query.

You are provided with function signatures within <tools></tools> XML tags:
<tools>
[computer_use_tool_def]
</tools>

For each function call, return a JSON object with function name and arguments within <tool_call></tool_call> XML tags:
<tool_call>
{"name": <function-name>, "arguments": <args-json-object>}
</tool_call>

# Response format

Response format for every step:
1) Memory: facts you would like to memorize for future actions in json format. Include the current step.
2) Progress: Decompose the task into subtasks and what has been finished so far with json format. Include progress of the current step.
3) Intention: clearly state which subtask you're working on at this step with the json key.
4) Action: a short sentence describing what to do in the UI to accomplish the next subtask.
5) A single <tool_call>...</tool_call> block containing only the JSON: {"name": <function-name>, "arguments": <args-json-object>}.

Rules:
- Output exactly in the order: Memory, Progress, Intention, Action, <tool_call>.
- You MUST use json format for the Memory and Progress parts.
- Example Task: "Search and compare the prices and locations of product 1 and product 2 on Amazon."
  - Example of Memory json format: {"Price of product 1": "10.00", "Location of product 1": "10.00", "Price of product 2": "12.00"}.
  - Example of Progress json format: {"Go to Amazon.com": "finished", "Search for price of product 1": "finished", "Search for location of product 1": "finished", "Search for price of product 2": "finished", "Search for location of product 2": "not finished", "Compare product 1 and product 2": "not finished"}.
  - Example of Intention json key format: "Search for location of product 2".
(*@\textcolor{blue!60!black}{- (default, additive memory prompt) You CANNOT modify previous Memory. Only append to it.}@*)
(*@\textcolor{red!60!black}{- (ablation, compressive memory prompt) Each step, you should COMPRESS previous memory by identifying only those relevant.}@*)
- You CAN modify Progress from previous conversation to further decompose the task and guide your next action.
  - For example, if the previous assistant message specifies Progress: {"Go to Amazon.com": "finished", "Search for product 1": "finished", "Search for product 2": "not finished", "Compare product 1 and 2": "not finished"},
  - You should further decompose "Search for product 1" and "Search for product 2" into "Search for price of product 1" and "Search for location of product 1", and "Search for price of product 2" and "Search for location of product 2".
- Do not output anything else outside those five parts.

--- MESSAGE 2: USER ---
Please generate the next action according to the UI screenshot and task.

Task: [task_description]

Initial website: [initial_website_url]

Generate the next action to complete the task.

[current_screenshot]
\end{tcblisting}

\subsection{Hyperparameters} \label{app:hparams}

\Cref{tab:hyperparameters} reports the hyperparameters used for the three main runs on the Instruct variant: GRPO with the $1/|\tau_i|$ normalizer, GRPO with the constant-$1/k$ fix, and RAFT++. The upper sub-table lists settings shared across all three runs; the lower sub-table lists the settings that differ.

RAFT++ uses a consumer batch size of $120$ rather than $128$ because we find that using 6 GPUs for training is more speed-optimal for RAFT++, while because we use a group size of $8$ we naturally use a consumer batch size of $128$ for GRPO. This is purely a scheduling choice; no other hyperparameter was retuned.

The Thinking-variant GRPO runs share the same prompt template (\Cref{app:prompts}) and the same loss, optimizer, and rollout settings as the Instruct GRPO runs; only the base model differs (Qwen3-VL-8B-Thinking) and, at runtime, the previous step's thinking tokens are hidden from history. The Thinking-variant RAFT++ run additionally uses a higher learning rate ($1\times 10^{-5}$), a single update epoch per batch, and a relaxed off-policy lag cap of $4$.

\begin{table*}[t]
  \caption{\textbf{Hyperparameters for the three main Qwen3-VL-8B-Instruct runs.} Values are taken from each run's wandb config.}
  \label{tab:hyperparameters}
  \centering
  \small
  \setlength{\tabcolsep}{6pt}
  \begin{tabular}{l l}
  \toprule
  \multicolumn{2}{c}{\textit{Settings common to all three runs}} \\
  \midrule
  Base model                     & \texttt{Qwen/Qwen3-VL-8B-Instruct}, bfloat16 \\
  Cluster                        & $2$ nodes $\times\, 8$ GPUs (B200) \\
  Optimizer                      & AdamW ($\beta_1{=}0.9$, $\beta_2{=}0.999$, $\epsilon{=}10^{-8}$) \\
  Learning rate                  & $5\times 10^{-6}$ \\
  LR scheduler                   & cosine, max steps $300$, min ratio $0.1$, warmup proportion $0.0167$ \\
  Weight decay                   & $0.01$ \\
  Gradient clipping              & $1.0$ \\
  Update epochs per batch        & $3$ \\
  Minibatches per update         & $2$ \\
  Importance-sampling level      & token \\
  Off-policy correction          & decoupled PPO, proximal logprobs recomputed \\
  $\epsilon$-clip                & $0.2$ \\
  Max off-policy lag ($\eta$)    & $2$ \\
  KL coefficient (to reference)  & $0$ \\
  Disable dropout                & true \\
  Gradient checkpointing         & true \\
  \midrule
  Rollout temperature            & $1.0$ \\
  Rollout top-$p$ / top-$k$ / freq.\ penalty & $1.0$ / $10^{8}$ / $0$ (effectively disabled) \\
  Max new tokens per step        & $4096$ \\
  Inference engine               & SGLang, flashinfer backend, context length $16{,}384$ \\
  Evaluator backbone             & \texttt{GPT-4o} \\
  SGLang \texttt{mem\_fraction\_static} & $0.75$ \\
  Max concurrent inference requests & $48$ \\
  \midrule
  WebGym prompt version          & complete \\
  Interaction mode               & coordinates-only \\
  History window                 & previous step only \\
  Train horizons (easy, medium, hard) & $(10, 20, 30)$ \\
  Test horizons (easy, medium, hard)  & $(30, 50, 70)$ \\
  Test-time decoding             & greedy \\
  \bottomrule
  \end{tabular}

  \vspace{0.8em}

  \begin{tabular}{l c c c}
  \toprule
  Hyperparameter & GRPO (with LN) & GRPO (no LN) & RAFT++ \\
  \midrule
  Loss (\texttt{actor\_loss})              & \texttt{grpo}             & \texttt{grpo}                       & \texttt{fbc} \\
  Step-number normalizer                   & $1/|\tau_i|$              & constant $1/k$ with $k{=}10$        & $1/|\tau_i|$ \\
  Group mean/std normalization             & group, size $8$           & group, size $8$                     & disabled \\
  Dual-clip $c$ (\Cref{equ:dual-clip})     & $3.0$                     & $3.0$                               & disabled \\
  Samples per task (group size $G$)        & $8$                       & $8$                                 & $1$ \\
  Consumer batch size                      & $128$                     & $128$                               & $120$ \\
  Filter steps with unchanged screenshot   & no                        & no                                  & yes \\
  \bottomrule
  \end{tabular}
\end{table*}

\end{document}

%% file: figures/vis_loss.tex
\begin{panelmd}{Base $\cdot$ n\_steps=10 $\cdot$ reward=1.0 $\cdot$ 178 tokens}
\tokent{255}{255}{255}{\textbf{Memory}}%
\tokent{255}{255}{255}{\textbf{:}}%
\tokent{255}{255}{255}{\ \{"}%
\tokent{255}{255}{255}{Task}%
\tokent{255}{255}{255}{":}%
\tokent{255}{255}{255}{\ "}%
\tokent{255}{255}{255}{Find}%
\tokent{255}{255}{255}{\ out}%
\tokent{255}{255}{255}{\ when}%
\tokent{255}{255}{255}{\ the}%
\tokent{255}{255}{255}{\ next}%
\tokent{255}{255}{255}{\ full}%
\tokent{255}{255}{255}{\ moon}%
\tokent{255}{255}{255}{\ is}%
\tokent{255}{255}{255}{\ scheduled}%
\tokent{255}{255}{255}{\ according}%
\tokent{255}{255}{255}{\ to}%
\tokent{255}{255}{255}{\ Space}%
\tokent{255}{255}{255}{.com}%
\tokent{255}{255}{255}{'s}%
\tokent{255}{255}{255}{\ '}%
\tokent{255}{255}{255}{Space}%
\tokent{255}{255}{255}{\ Calendar}%
\tokent{255}{255}{255}{'}%
\tokent{255}{255}{255}{."}%
\tokent{255}{255}{255}{\}\newline }%
\tokent{255}{255}{255}{\textbf{Progress}}%
\tokent{255}{255}{255}{\textbf{:}}%
\tokent{255}{255}{255}{\ \{"}%
\tokent{255}{255}{255}{Go}%
\tokent{255}{255}{255}{\ to}%
\tokent{255}{255}{255}{\ Space}%
\tokent{255}{255}{255}{.com}%
\tokent{255}{255}{255}{":}%
\tokent{255}{255}{255}{\ "}%
\tokent{255}{255}{255}{finished}%
\tokent{255}{255}{255}{",}%
\tokent{255}{255}{255}{\ "}%
\tokent{255}{255}{255}{Navigate}%
\tokent{255}{255}{255}{\ to}%
\tokent{255}{255}{255}{\ '}%
\tokent{255}{255}{255}{Space}%
\tokent{255}{255}{255}{\ Calendar}%
\tokent{255}{255}{255}{'}%
\tokent{255}{255}{255}{":}%
\tokent{255}{255}{255}{\ "}%
\tokent{255}{255}{255}{finished}%
\tokent{255}{255}{255}{",}%
\tokent{255}{255}{255}{\ "}%
\tokent{255}{255}{255}{Find}%
\tokent{255}{255}{255}{\ next}%
\tokent{255}{255}{255}{\ full}%
\tokent{255}{255}{255}{\ moon}%
\tokent{255}{255}{255}{\ date}%
\tokent{255}{255}{255}{":}%
\tokent{255}{255}{255}{\ "}%
\tokent{255}{255}{255}{finished}%
\tokent{255}{255}{255}{"\}\newline }%
\tokent{255}{255}{255}{Int}%
\tokent{255}{255}{255}{ention}%
\tokent{255}{255}{255}{:}%
\tokent{255}{255}{255}{\ The}%
\tokent{255}{255}{255}{\ next}%
\tokent{255}{255}{255}{\ full}%
\tokent{255}{255}{255}{\ moon}%
\tokent{255}{255}{255}{\ is}%
\tokent{255}{255}{255}{\ scheduled}%
\tokent{255}{255}{255}{\ for}%
\tokent{255}{255}{255}{\ May}%
\tokent{255}{255}{255}{\ }%
\tokent{255}{255}{255}{1}%
\tokent{255}{255}{255}{,}%
\tokent{255}{255}{255}{\ }%
\tokent{255}{255}{255}{2}%
\tokent{255}{255}{255}{0}%
\tokent{255}{255}{255}{2}%
\tokent{255}{255}{255}{6}%
\tokent{255}{255}{255}{,}%
\tokent{255}{255}{255}{\ at}%
\tokent{255}{255}{255}{\ }%
\tokent{255}{255}{255}{1}%
\tokent{255}{255}{255}{:}%
\tokent{255}{255}{255}{2}%
\tokent{255}{255}{255}{3}%
\tokent{255}{255}{255}{\ p}%
\tokent{255}{255}{255}{.m}%
\tokent{255}{255}{255}{.}%
\tokent{255}{255}{255}{\ (}%
\tokent{255}{255}{255}{ED}%
\tokent{255}{255}{255}{T}%
\tokent{255}{255}{255}{).\newline }%
\tokent{255}{255}{255}{\textbf{Action}}%
\tokent{255}{255}{255}{\textbf{:}}%
\tokent{255}{255}{255}{\ The}%
\tokent{255}{255}{255}{\ next}%
\tokent{255}{255}{255}{\ full}%
\tokent{255}{255}{255}{\ moon}%
\tokent{255}{255}{255}{\ is}%
\tokent{255}{255}{255}{\ scheduled}%
\tokent{255}{255}{255}{\ for}%
\tokent{255}{255}{255}{\ May}%
\tokent{255}{255}{255}{\ }%
\tokent{255}{255}{255}{1}%
\tokent{255}{255}{255}{,}%
\tokent{255}{255}{255}{\ }%
\tokent{255}{255}{255}{2}%
\tokent{255}{255}{255}{0}%
\tokent{255}{255}{255}{2}%
\tokent{255}{255}{255}{6}%
\tokent{255}{255}{255}{,}%
\tokent{255}{255}{255}{\ at}%
\tokent{255}{255}{255}{\ }%
\tokent{255}{255}{255}{1}%
\tokent{255}{255}{255}{:}%
\tokent{255}{255}{255}{2}%
\tokent{255}{255}{255}{3}%
\tokent{255}{255}{255}{\ p}%
\tokent{255}{255}{255}{.m}%
\tokent{255}{255}{255}{.}%
\tokent{255}{255}{255}{\ (}%
\tokent{255}{255}{255}{ED}%
\tokent{255}{255}{255}{T}%
\tokent{255}{255}{255}{).\newline }%
\tokent{255}{255}{255}{\textbf{\textless{}tool\_call\textgreater{}}}%
\tokent{255}{255}{255}{\newline }%
\tokent{255}{255}{255}{\{"}%
\tokent{255}{255}{255}{name}%
\tokent{255}{255}{255}{":}%
\tokent{255}{255}{255}{\ "}%
\tokent{255}{255}{255}{computer}%
\tokent{255}{255}{255}{\_use}%
\tokent{255}{255}{255}{",}%
\tokent{255}{255}{255}{\ "}%
\tokent{255}{255}{255}{arguments}%
\tokent{255}{255}{255}{":}%
\tokent{255}{255}{255}{\ \{"}%
\tokent{255}{255}{255}{action}%
\tokent{255}{255}{255}{":}%
\tokent{255}{255}{255}{\ "}%
\tokent{255}{255}{255}{answer}%
\tokent{255}{255}{255}{",}%
\tokent{255}{255}{255}{\ "}%
\tokent{255}{255}{255}{text}%
\tokent{255}{255}{255}{":}%
\tokent{255}{255}{255}{\ "}%
\tokent{255}{255}{255}{The}%
\tokent{255}{255}{255}{\ next}%
\tokent{255}{255}{255}{\ full}%
\tokent{255}{255}{255}{\ moon}%
\tokent{255}{255}{255}{\ is}%
\tokent{255}{255}{255}{\ scheduled}%
\tokent{255}{255}{255}{\ for}%
\tokent{255}{255}{255}{\ May}%
\tokent{255}{255}{255}{\ }%
\tokent{255}{255}{255}{1}%
\tokent{255}{255}{255}{,}%
\tokent{255}{255}{255}{\ }%
\tokent{255}{255}{255}{2}%
\tokent{255}{255}{255}{0}%
\tokent{255}{255}{255}{2}%
\tokent{255}{255}{255}{6}%
\tokent{255}{255}{255}{,}%
\tokent{255}{255}{255}{\ at}%
\tokent{255}{255}{255}{\ }%
\tokent{255}{255}{255}{1}%
\tokent{255}{255}{255}{:}%
\tokent{255}{255}{255}{2}%
\tokent{255}{255}{255}{3}%
\tokent{255}{255}{255}{\ p}%
\tokent{255}{255}{255}{.m}%
\tokent{255}{255}{255}{.}%
\tokent{255}{255}{255}{\ (}%
\tokent{255}{255}{255}{ED}%
\tokent{255}{255}{255}{T}%
\tokent{255}{255}{255}{).}%
\tokent{255}{255}{255}{"\}\}\newline }%
\tokent{255}{255}{255}{\textbf{\textless{}/tool\_call\textgreater{}}}%
\tokent{255}{255}{255}{\textbf{\textless{}|im\_end|\textgreater{}}}
\end{panelmd}

\medskip

\begin{panelmd}{RAFT++ $\cdot$ n\_steps=10 $\cdot$ reward=1.0 $\cdot$ 174 tokens}
\tokent{255}{255}{255}{\textbf{Memory}}%
\tokent{255}{255}{255}{\textbf{:}}%
\tokent{255}{255}{255}{\ \{"}%
\tokent{255}{255}{255}{target}%
\tokent{255}{255}{255}{\_website}%
\tokent{255}{255}{255}{":}%
\tokent{255}{255}{255}{\ "}%
\tokent{255}{255}{255}{https}%
\tokent{255}{255}{255}{://}%
\tokent{255}{255}{255}{www}%
\tokent{255}{255}{255}{.food}%
\tokent{255}{255}{255}{network}%
\tokent{255}{255}{255}{.com}%
\tokent{255}{255}{255}{/",}%
\tokent{255}{255}{255}{\ "}%
\tokent{255}{255}{255}{task}%
\tokent{255}{255}{255}{":}%
\tokent{255}{255}{255}{\ "}%
\tokent{255}{255}{255}{Search}%
\tokent{255}{255}{255}{\ for}%
\tokent{255}{255}{255}{\ a}%
\tokent{255}{255}{255}{\ recipe}%
\tokent{255}{255}{255}{\ for}%
\tokent{255}{255}{255}{\ a}%
\tokent{255}{255}{255}{\ dessert}%
\tokent{255}{255}{255}{\ that}%
\tokent{255}{255}{255}{\ is}%
\tokent{255}{255}{255}{\ suitable}%
\tokent{255}{255}{255}{\ for}%
\tokent{255}{255}{255}{\ a}%
\tokent{255}{255}{255}{\ romantic}%
\tokent{255}{255}{255}{\ dinner}%
\tokent{255}{255}{255}{,}%
\tokent{255}{255}{255}{\ such}%
\tokent{255}{255}{255}{\ as}%
\tokent{255}{255}{255}{\ a}%
\tokent{255}{255}{255}{\ chocolate}%
\tokent{255}{255}{255}{\ sou}%
\tokent{255}{255}{255}{ff}%
\tokent{255}{255}{255}{lé}%
\tokent{255}{255}{255}{\ or}%
\tokent{255}{255}{255}{\ a}%
\tokent{255}{255}{255}{\ fruit}%
\tokent{255}{255}{255}{\ tart}%
\tokent{255}{255}{255}{.",}%
\tokent{255}{255}{255}{\ "}%
\tokent{255}{255}{255}{search}%
\tokent{255}{255}{255}{\_term}%
\tokent{255}{255}{255}{":}%
\tokent{255}{255}{255}{\ "}%
\tokent{255}{255}{255}{fruit}%
\tokent{255}{255}{255}{\ tart}%
\tokent{255}{255}{255}{",}%
\tokent{255}{255}{255}{\ "}%
\tokent{255}{255}{255}{results}%
\tokent{255}{255}{255}{\_found}%
\tokent{255}{255}{255}{":}%
\tokent{255}{255}{255}{\ "}%
\tokent{255}{255}{255}{yes}%
\tokent{255}{255}{255}{",}%
\tokent{255}{255}{255}{\ "}%
\tokent{255}{255}{255}{recipe}%
\tokent{255}{255}{255}{\_display}%
\tokent{255}{255}{255}{ed}%
\tokent{255}{255}{255}{":}%
\tokent{255}{255}{255}{\ "}%
\tokent{255}{255}{255}{F}%
\tokent{255}{255}{255}{ruit}%
\tokent{255}{255}{255}{-T}%
\tokent{255}{255}{255}{art}%
\tokent{255}{255}{255}{\ Flag}%
\tokent{255}{255}{255}{"\}\newline }%
\tokent{255}{255}{255}{\textbf{Progress}}%
\tokent{255}{255}{255}{\textbf{:}}%
\tokent{255}{255}{255}{\ \{"}%
\tokent{255}{255}{255}{Go}%
\tokent{255}{255}{255}{\ to}%
\tokent{255}{255}{255}{\ food}%
\tokent{255}{255}{255}{network}%
\tokent{255}{255}{255}{.com}%
\tokent{255}{255}{255}{":}%
\tokent{255}{255}{255}{\ "}%
\tokent{255}{255}{255}{finished}%
\tokent{255}{255}{255}{",}%
\tokent{255}{255}{255}{\ "}%
\tokent{255}{255}{255}{Search}%
\tokent{255}{255}{255}{\ for}%
\tokent{255}{255}{255}{\ a}%
\tokent{255}{255}{255}{\ romantic}%
\tokent{255}{255}{255}{\ dinner}%
\tokent{255}{255}{255}{\ dessert}%
\tokent{255}{255}{255}{\ recipe}%
\tokent{255}{255}{255}{":}%
\tokent{255}{255}{255}{\ "}%
\tokent{255}{255}{255}{finished}%
\tokent{255}{255}{255}{"\}\newline }%
\tokent{255}{255}{255}{Int}%
\tokent{255}{255}{255}{ention}%
\tokent{255}{255}{255}{:}%
\tokent{255}{255}{255}{\ The}%
\tokent{255}{255}{255}{\ task}%
\tokent{255}{255}{255}{\ is}%
\tokent{255}{255}{255}{\ completed}%
\tokent{255}{255}{255}{\ as}%
\tokent{255}{255}{255}{\ a}%
\tokent{255}{255}{255}{\ suitable}%
\tokent{255}{255}{255}{\ recipe}%
\tokent{255}{255}{255}{\ for}%
\tokent{255}{255}{255}{\ a}%
\tokent{255}{255}{255}{\ romantic}%
\tokent{255}{255}{255}{\ dinner}%
\tokent{255}{255}{255}{\ has}%
\tokent{255}{255}{255}{\ been}%
\tokent{255}{255}{255}{\ found}%
\tokent{255}{255}{255}{.\newline }%
\tokent{255}{255}{255}{\textbf{Action}}%
\tokent{255}{255}{255}{\textbf{:}}%
\tokent{255}{255}{255}{\ The}%
\tokent{255}{255}{255}{\ task}%
\tokent{255}{255}{255}{\ is}%
\tokent{255}{255}{255}{\ completed}%
\tokent{255}{255}{255}{\ as}%
\tokent{255}{255}{255}{\ a}%
\tokent{255}{255}{255}{\ suitable}%
\tokent{255}{255}{255}{\ recipe}%
\tokent{255}{255}{255}{\ for}%
\tokent{255}{255}{255}{\ a}%
\tokent{255}{255}{255}{\ romantic}%
\tokent{255}{255}{255}{\ dinner}%
\tokent{255}{255}{255}{\ has}%
\tokent{255}{255}{255}{\ been}%
\tokent{255}{255}{255}{\ found}%
\tokent{255}{255}{255}{.\newline }%
\tokent{255}{255}{255}{\textbf{\textless{}tool\_call\textgreater{}}}%
\tokent{255}{255}{255}{\newline }%
\tokent{255}{255}{255}{\{"}%
\tokent{255}{255}{255}{name}%
\tokent{255}{255}{255}{":}%
\tokent{255}{255}{255}{\ "}%
\tokent{255}{255}{255}{computer}%
\tokent{255}{255}{255}{\_use}%
\tokent{255}{255}{255}{",}%
\tokent{255}{255}{255}{\ "}%
\tokent{255}{255}{255}{arguments}%
\tokent{255}{255}{255}{":}%
\tokent{255}{255}{255}{\ \{"}%
\tokent{255}{255}{255}{action}%
\tokent{255}{255}{255}{":}%
\tokent{255}{255}{255}{\ "}%
\tokent{255}{255}{255}{answer}%
\tokent{255}{255}{255}{",}%
\tokent{255}{255}{255}{\ "}%
\tokent{255}{255}{255}{text}%
\tokent{255}{255}{255}{":}%
\tokent{255}{255}{255}{\ "}%
\tokent{255}{255}{255}{A}%
\tokent{255}{255}{255}{\ suitable}%
\tokent{255}{255}{255}{\ recipe}%
\tokent{255}{255}{255}{\ for}%
\tokent{255}{255}{255}{\ a}%
\tokent{255}{255}{255}{\ romantic}%
\tokent{255}{255}{255}{\ dinner}%
\tokent{255}{255}{255}{\ has}%
\tokent{255}{255}{255}{\ been}%
\tokent{255}{255}{255}{\ found}%
\tokent{255}{255}{255}{:}%
\tokent{255}{255}{255}{\ Fruit}%
\tokent{255}{255}{255}{-T}%
\tokent{255}{255}{255}{art}%
\tokent{255}{255}{255}{\ Flag}%
\tokent{255}{255}{255}{."}%
\tokent{255}{255}{255}{\}\}\newline }%
\tokent{255}{255}{255}{\textbf{\textless{}/tool\_call\textgreater{}}}%
\tokent{255}{255}{255}{\textbf{\textless{}|im\_end|\textgreater{}}}
\end{panelmd}

\medskip

\begin{panelmd}{GRPO without length norm $\cdot$ n\_steps=10 $\cdot$ reward=1.0 $\cdot$ 150 tokens}
\tokent{255}{255}{255}{\textbf{Memory}}%
\tokent{255}{255}{255}{\textbf{:}}%
\tokent{255}{255}{255}{\ \{"}%
\tokent{255}{255}{255}{accepted}%
\tokent{255}{255}{255}{\_cookies}%
\tokent{255}{255}{255}{":}%
\tokent{255}{255}{255}{\ "}%
\tokent{255}{255}{255}{finished}%
\tokent{255}{255}{255}{",}%
\tokent{255}{255}{255}{\ "}%
\tokent{255}{255}{255}{n}%
\tokent{255}{255}{255}{avig}%
\tokent{255}{255}{255}{ated}%
\tokent{255}{255}{255}{\_to}%
\tokent{255}{255}{255}{\_j}%
\tokent{255}{255}{255}{ackets}%
\tokent{255}{255}{255}{":}%
\tokent{255}{255}{255}{\ "}%
\tokent{255}{255}{255}{finished}%
\tokent{255}{255}{255}{",}%
\tokent{255}{255}{255}{\ "}%
\tokent{255}{255}{255}{found}%
\tokent{255}{255}{255}{\_le}%
\tokent{255}{255}{255}{ather}%
\tokent{255}{255}{255}{\_j}%
\tokent{255}{255}{255}{acket}%
\tokent{255}{255}{255}{\_price}%
\tokent{255}{255}{255}{":}%
\tokent{255}{255}{255}{\ "}%
\tokent{255}{255}{255}{finished}%
\tokent{255}{255}{255}{"\}\newline }%
\tokent{255}{255}{255}{\textbf{Progress}}%
\tokent{255}{255}{255}{\textbf{:}}%
\tokent{255}{255}{255}{\ \{"}%
\tokent{255}{255}{255}{Go}%
\tokent{255}{255}{255}{\ to}%
\tokent{255}{255}{255}{\ Diesel}%
\tokent{255}{255}{255}{.com}%
\tokent{255}{255}{255}{":}%
\tokent{255}{255}{255}{\ "}%
\tokent{255}{255}{255}{finished}%
\tokent{255}{255}{255}{",}%
\tokent{255}{255}{255}{\ "}%
\tokent{255}{255}{255}{Accept}%
\tokent{255}{255}{255}{\ cookies}%
\tokent{255}{255}{255}{":}%
\tokent{255}{255}{255}{\ "}%
\tokent{255}{255}{255}{finished}%
\tokent{255}{255}{255}{",}%
\tokent{255}{255}{255}{\ "}%
\tokent{255}{255}{255}{Navigate}%
\tokent{255}{255}{255}{\ to}%
\tokent{255}{255}{255}{\ '}%
\tokent{255}{255}{255}{J}%
\tokent{255}{255}{255}{ackets}%
\tokent{255}{255}{255}{'}%
\tokent{255}{255}{255}{\ category}%
\tokent{255}{255}{255}{\ for}%
\tokent{255}{255}{255}{\ men}%
\tokent{255}{255}{255}{":}%
\tokent{255}{255}{255}{\ "}%
\tokent{255}{255}{255}{finished}%
\tokent{255}{255}{255}{",}%
\tokent{255}{255}{255}{\ "}%
\tokent{255}{255}{255}{Find}%
\tokent{255}{255}{255}{\ price}%
\tokent{255}{255}{255}{\ of}%
\tokent{255}{255}{255}{\ any}%
\tokent{255}{255}{255}{\ leather}%
\tokent{255}{255}{255}{\ jacket}%
\tokent{255}{255}{255}{":}%
\tokent{255}{255}{255}{\ "}%
\tokent{255}{255}{255}{finished}%
\tokent{255}{255}{255}{"\}\newline }%
\tokent{255}{255}{255}{Int}%
\tokent{255}{255}{255}{ention}%
\tokent{255}{255}{255}{:}%
\tokent{255}{255}{255}{\ Complete}%
\tokent{255}{255}{255}{\ the}%
\tokent{255}{255}{255}{\ task}%
\tokent{255}{255}{255}{\ by}%
\tokent{255}{255}{255}{\ providing}%
\tokent{255}{255}{255}{\ the}%
\tokent{255}{255}{255}{\ price}%
\tokent{255}{255}{255}{\ of}%
\tokent{255}{255}{255}{\ a}%
\tokent{255}{255}{255}{\ leather}%
\tokent{255}{255}{255}{\ jacket}%
\tokent{255}{255}{255}{\ found}%
\tokent{255}{255}{255}{\ in}%
\tokent{255}{255}{255}{\ the}%
\tokent{255}{255}{255}{\ '}%
\tokent{255}{255}{255}{J}%
\tokent{255}{255}{255}{ackets}%
\tokent{255}{255}{255}{'}%
\tokent{255}{255}{255}{\ category}%
\tokent{255}{255}{255}{\newline }%
\tokent{255}{255}{255}{\textbf{Action}}%
\tokent{255}{255}{255}{\textbf{:}}%
\tokent{255}{255}{255}{\ The}%
\tokent{255}{255}{255}{\ task}%
\tokent{255}{255}{255}{\ is}%
\tokent{255}{255}{255}{\ completed}%
\tokent{255}{255}{255}{.}%
\tokent{255}{255}{255}{\ The}%
\tokent{255}{255}{255}{\ price}%
\tokent{255}{255}{255}{\ of}%
\tokent{255}{255}{255}{\ a}%
\tokent{255}{255}{255}{\ leather}%
\tokent{255}{255}{255}{\ jacket}%
\tokent{255}{255}{255}{\ is}%
\tokent{255}{255}{255}{\ \$}%
\tokent{255}{255}{255}{3}%
\tokent{255}{255}{255}{9}%
\tokent{255}{255}{255}{5}%
\tokent{255}{255}{255}{.}%
\tokent{255}{255}{255}{0}%
\tokent{255}{255}{255}{0}%
\tokent{255}{255}{255}{.\newline }%
\tokent{255}{255}{255}{\textbf{\textless{}tool\_call\textgreater{}}}%
\tokent{255}{255}{255}{\newline }%
\tokent{255}{255}{255}{\{"}%
\tokent{255}{255}{255}{name}%
\tokent{255}{255}{255}{":}%
\tokent{255}{255}{255}{\ "}%
\tokent{255}{255}{255}{computer}%
\tokent{255}{255}{255}{\_use}%
\tokent{255}{255}{255}{",}%
\tokent{255}{255}{255}{\ "}%
\tokent{255}{255}{255}{arguments}%
\tokent{255}{255}{255}{":}%
\tokent{255}{255}{255}{\ \{"}%
\tokent{255}{255}{255}{action}%
\tokent{255}{255}{255}{":}%
\tokent{255}{255}{255}{\ "}%
\tokent{255}{255}{255}{answer}%
\tokent{255}{255}{255}{",}%
\tokent{255}{255}{255}{\ "}%
\tokent{255}{255}{255}{text}%
\tokent{255}{255}{255}{":}%
\tokent{255}{255}{255}{\ "\$}%
\tokent{255}{255}{255}{3}%
\tokent{255}{255}{255}{9}%
\tokent{255}{255}{255}{5}%
\tokent{255}{255}{255}{.}%
\tokent{255}{255}{255}{0}%
\tokent{255}{255}{255}{0}%
\tokent{255}{255}{255}{"\}\}\newline }%
\tokent{255}{255}{255}{\textbf{\textless{}/tool\_call\textgreater{}}}%
\tokent{255}{255}{255}{\textbf{\textless{}|im\_end|\textgreater{}}}
\end{panelmd}
\medskip

\begin{panelmd}{GRPO with length norm $\cdot$ n\_steps=10 $\cdot$ reward=1.0 $\cdot$ 589 tokens}
\tokent{255}{255}{255}{\textbf{Memory}}%
\tokent{255}{255}{255}{\textbf{:}}%
\tokent{255}{255}{255}{\ \{"}%
\tokent{255}{255}{255}{task}%
\tokent{255}{255}{255}{":}%
\tokent{255}{255}{255}{\ "}%
\tokent{255}{255}{255}{Search}%
\tokent{255}{255}{255}{\ for}%
\tokent{255}{255}{255}{\ news}%
\tokent{255}{255}{255}{\ articles}%
\tokent{255}{255}{255}{\ about}%
\tokent{255}{255}{255}{\ the}%
\tokent{255}{255}{255}{\ latest}%
\tokent{255}{255}{255}{\ technological}%
\tokent{255}{255}{255}{\ advancements}%
\tokent{255}{255}{255}{\ in}%
\tokent{255}{255}{255}{\ the}%
\tokent{255}{255}{255}{\ field}%
\tokent{255}{255}{255}{\ of}%
\tokent{255}{255}{255}{\ medicine}%
\tokent{255}{255}{255}{\ and}%
\tokent{255}{255}{255}{\ summarize}%
\tokent{255}{255}{255}{\ the}%
\tokent{255}{255}{255}{\ findings}%
\tokent{255}{255}{255}{\ of}%
\tokent{255}{255}{255}{\ the}%
\tokent{255}{255}{255}{\ articles}%
\tokent{255}{255}{255}{.",}%
\tokent{255}{255}{255}{\ "}%
\tokent{255}{255}{255}{search}%
\tokent{255}{255}{255}{\_query}%
\tokent{255}{255}{255}{":}%
\tokent{255}{255}{255}{\ "}%
\tokent{255}{255}{255}{latest}%
\tokent{255}{255}{255}{\ technological}%
\tokent{255}{255}{255}{\ advancements}%
\tokent{255}{255}{255}{\ in}%
\tokent{255}{255}{255}{\ the}%
\tokent{255}{255}{255}{\ field}%
\tokent{255}{255}{255}{\ of}%
\tokent{255}{255}{255}{\ medicine}%
\tokent{255}{255}{255}{",}%
\tokent{255}{255}{255}{\ "}%
\tokent{255}{255}{255}{current}%
\tokent{255}{255}{255}{\_page}%
\tokent{255}{255}{255}{":}%
\tokent{255}{255}{255}{\ "}%
\tokent{255}{255}{255}{NI}%
\tokent{255}{255}{255}{H}%
\tokent{255}{255}{255}{\ website}%
\tokent{255}{255}{255}{\ article}%
\tokent{255}{255}{255}{\ page}%
\tokent{255}{255}{255}{",}%
\tokent{255}{255}{255}{\ "}%
\tokent{255}{255}{255}{search}%
\tokent{255}{255}{255}{\_results}%
\tokent{255}{255}{255}{":}%
\tokent{255}{255}{255}{\ ["}%
\tokent{255}{255}{255}{AI}%
\tokent{255}{255}{255}{\ teachers}%
\tokent{255}{255}{255}{\ and}%
\tokent{255}{255}{255}{\ cy}%
\tokent{255}{255}{255}{bert}%
\tokent{255}{255}{255}{net}%
\tokent{255}{255}{255}{ics}%
\tokent{255}{255}{255}{\ -}%
\tokent{255}{255}{255}{\ what}%
\tokent{255}{255}{255}{\ could}%
\tokent{255}{255}{255}{\ the}%
\tokent{255}{255}{255}{\ world}%
\tokent{255}{255}{255}{\ look}%
\tokent{255}{255}{255}{\ like}%
\tokent{255}{255}{255}{\ in}%
\tokent{255}{255}{255}{\ }%
\tokent{255}{255}{255}{2}%
\tokent{255}{255}{255}{0}%
\tokent{255}{255}{255}{5}%
\tokent{255}{255}{255}{0}%
\tokent{255}{255}{255}{?",}%
\tokent{255}{255}{255}{\ "'}%
\tokent{255}{255}{255}{There}%
\tokent{255}{255}{255}{\ is}%
\tokent{255}{255}{255}{\ not}%
\tokent{255}{255}{255}{\ enough}%
\tokent{255}{255}{255}{\ time}%
\tokent{255}{255}{255}{\ or}%
\tokent{255}{255}{255}{\ space}%
\tokent{255}{255}{255}{\ in}%
\tokent{255}{255}{255}{\ the}%
\tokent{255}{255}{255}{\ day}%
\tokent{255}{255}{255}{'}%
\tokent{255}{255}{255}{"],}%
\tokent{255}{255}{255}{\ "}%
\tokent{255}{255}{255}{article}%
\tokent{255}{255}{255}{\_title}%
\tokent{255}{255}{255}{":}%
\tokent{255}{255}{255}{\ "}%
\tokent{255}{255}{255}{AI}%
\tokent{255}{255}{255}{\ teachers}%
\tokent{255}{255}{255}{\ and}%
\tokent{255}{255}{255}{\ cy}%
\tokent{255}{255}{255}{bert}%
\tokent{255}{255}{255}{net}%
\tokent{255}{255}{255}{ics}%
\tokent{255}{255}{255}{\ -}%
\tokent{255}{255}{255}{\ what}%
\tokent{255}{255}{255}{\ could}%
\tokent{255}{255}{255}{\ the}%
\tokent{255}{255}{255}{\ world}%
\tokent{255}{255}{255}{\ look}%
\tokent{255}{255}{255}{\ like}%
\tokent{255}{255}{255}{\ in}%
\tokent{255}{255}{255}{\ }%
\tokent{255}{255}{255}{2}%
\tokent{255}{255}{255}{0}%
\tokent{255}{255}{255}{5}%
\tokent{255}{255}{255}{0}%
\tokent{255}{255}{255}{?",}%
\tokent{255}{255}{255}{\ "}%
\tokent{255}{255}{255}{popup}%
\tokent{255}{255}{255}{":}%
\tokent{255}{255}{255}{\ "}%
\tokent{255}{255}{255}{survey}%
\tokent{255}{255}{255}{\ popup}%
\tokent{255}{255}{255}{",}%
\tokent{255}{255}{255}{\ "}%
\tokent{255}{255}{255}{article}%
\tokent{255}{255}{255}{\_content}%
\tokent{255}{255}{255}{":}%
\tokent{255}{255}{255}{\ "}%
\tokent{255}{255}{255}{AI}%
\tokent{255}{255}{255}{\ teachers}%
\tokent{255}{255}{255}{\ and}%
\tokent{255}{255}{255}{\ cy}%
\tokent{255}{255}{255}{bert}%
\tokent{255}{255}{255}{net}%
\tokent{255}{255}{255}{ics}%
\tokent{255}{255}{255}{\ -}%
\tokent{255}{255}{255}{\ what}%
\tokent{255}{255}{255}{\ could}%
\tokent{255}{255}{255}{\ the}%
\tokent{255}{255}{255}{\ world}%
\tokent{255}{255}{255}{\ look}%
\tokent{255}{255}{255}{\ like}%
\tokent{255}{255}{255}{\ in}%
\tokent{255}{255}{255}{\ }%
\tokent{255}{255}{255}{2}%
\tokent{255}{255}{255}{0}%
\tokent{255}{255}{255}{5}%
\tokent{255}{255}{255}{0}%
\tokent{255}{255}{255}{?",}%
\tokent{255}{255}{255}{\ "}%
\tokent{255}{255}{255}{subscription}%
\tokent{255}{255}{255}{\_popup}%
\tokent{255}{255}{255}{":}%
\tokent{255}{255}{255}{\ "}%
\tokent{255}{255}{255}{active}%
\tokent{255}{255}{255}{",}%
\tokent{255}{255}{255}{\ "}%
\tokent{255}{255}{255}{attempt}%
\tokent{255}{255}{255}{ed}%
\tokent{255}{255}{255}{\_navigation}%
\tokent{255}{255}{255}{":}%
\tokent{255}{255}{255}{\ "}%
\tokent{255}{255}{255}{https}%
\tokent{255}{255}{255}{://}%
\tokent{255}{255}{255}{www}%
\tokent{255}{255}{255}{.b}%
\tokent{255}{255}{255}{bc}%
\tokent{255}{255}{255}{.com}%
\tokent{255}{255}{255}{/news}%
\tokent{255}{255}{255}{/}%
\tokent{255}{255}{255}{health}%
\tokent{255}{255}{255}{-}%
\tokent{255}{255}{255}{6}%
\tokent{255}{255}{255}{4}%
\tokent{255}{255}{255}{8}%
\tokent{255}{255}{255}{1}%
\tokent{255}{255}{255}{0}%
\tokent{255}{255}{255}{6}%
\tokent{255}{255}{255}{9}%
\tokent{255}{255}{255}{8}%
\tokent{255}{255}{255}{",}%
\tokent{255}{255}{255}{\ "}%
\tokent{255}{255}{255}{error}%
\tokent{255}{255}{255}{\_page}%
\tokent{255}{255}{255}{":}%
\tokent{255}{255}{255}{\ "}%
\tokent{255}{255}{255}{5}%
\tokent{255}{255}{255}{0}%
\tokent{255}{255}{255}{0}%
\tokent{255}{255}{255}{\ Internal}%
\tokent{255}{255}{255}{\ Server}%
\tokent{255}{255}{255}{\ Error}%
\tokent{255}{255}{255}{",}%
\tokent{255}{255}{255}{\ "}%
\tokent{255}{255}{255}{new}%
\tokent{255}{255}{255}{\_website}%
\tokent{255}{255}{255}{":}%
\tokent{255}{255}{255}{\ "}%
\tokent{255}{255}{255}{https}%
\tokent{255}{255}{255}{://}%
\tokent{255}{255}{255}{www}%
\tokent{255}{255}{255}{.nih}%
\tokent{255}{255}{255}{.gov}%
\tokent{255}{255}{255}{/",}%
\tokent{255}{255}{255}{\ "}%
\tokent{255}{255}{255}{search}%
\tokent{255}{255}{255}{\_bar}%
\tokent{255}{255}{255}{":}%
\tokent{255}{255}{255}{\ "}%
\tokent{255}{255}{255}{used}%
\tokent{255}{255}{255}{",}%
\tokent{255}{255}{255}{\ "}%
\tokent{255}{255}{255}{search}%
\tokent{255}{255}{255}{\_results}%
\tokent{255}{255}{255}{\_count}%
\tokent{255}{255}{255}{":}%
\tokent{255}{255}{255}{\ "}%
\tokent{255}{255}{255}{2}%
\tokent{255}{255}{255}{,}%
\tokent{255}{255}{255}{1}%
\tokent{255}{255}{255}{4}%
\tokent{255}{255}{255}{1}%
\tokent{255}{255}{255}{\ results}%
\tokent{255}{255}{255}{",}%
\tokent{255}{255}{255}{\ "}%
\tokent{255}{255}{255}{article}%
\tokent{255}{255}{255}{\_found}%
\tokent{255}{255}{255}{":}%
\tokent{255}{255}{255}{\ "}%
\tokent{255}{255}{255}{S}%
\tokent{255}{255}{255}{LEEP}%
\tokent{255}{255}{255}{\ }%
\tokent{255}{255}{255}{2}%
\tokent{255}{255}{255}{0}%
\tokent{255}{255}{255}{2}%
\tokent{255}{255}{255}{4}%
\tokent{255}{255}{255}{,}%
\tokent{255}{255}{255}{\ the}%
\tokent{255}{255}{255}{\ Associated}%
\tokent{255}{255}{255}{\ Professional}%
\tokent{255}{255}{255}{\ Sleep}%
\tokent{255}{255}{255}{\ Soci}%
\tokent{255}{255}{255}{eties}%
\tokent{255}{255}{255}{\ (}%
\tokent{255}{255}{255}{AP}%
\tokent{255}{255}{255}{SS}%
\tokent{255}{255}{255}{)'}%
\tokent{255}{255}{255}{s}%
\tokent{255}{255}{255}{\ }%
\tokent{255}{255}{255}{3}%
\tokent{255}{255}{255}{8}%
\tokent{255}{255}{255}{th}%
\tokent{255}{255}{255}{\ Annual}%
\tokent{255}{255}{255}{\ Meeting}%
\tokent{255}{255}{255}{\ |}%
\tokent{255}{255}{255}{\ NHL}%
\tokent{255}{255}{255}{BI}%
\tokent{255}{255}{255}{,}%
\tokent{255}{255}{255}{\ NIH}%
\tokent{255}{255}{255}{",}%
\tokent{255}{255}{255}{\ "}%
\tokent{255}{255}{255}{article}%
\tokent{255}{255}{255}{\_content}%
\tokent{255}{255}{255}{\_summary}%
\tokent{255}{255}{255}{":}%
\tokent{255}{255}{255}{\ "}%
\tokent{255}{255}{255}{The}%
\tokent{255}{255}{255}{\ S}%
\tokent{255}{255}{255}{LEEP}%
\tokent{255}{255}{255}{\ }%
\tokent{255}{255}{255}{2}%
\tokent{255}{255}{255}{0}%
\tokent{255}{255}{255}{2}%
\tokent{255}{255}{255}{4}%
\tokent{255}{255}{255}{\ meeting}%
\tokent{255}{255}{255}{\ will}%
\tokent{255}{255}{255}{\ discuss}%
\tokent{255}{255}{255}{\ the}%
\tokent{255}{255}{255}{\ latest}%
\tokent{255}{255}{255}{\ advances}%
\tokent{255}{255}{255}{\ in}%
\tokent{255}{255}{255}{\ sleep}%
\tokent{255}{255}{255}{\ science}%
\tokent{255}{255}{255}{\ and}%
\tokent{255}{255}{255}{\ new}%
\tokent{255}{255}{255}{\ developments}%
\tokent{255}{255}{255}{\ in}%
\tokent{255}{255}{255}{\ the}%
\tokent{255}{255}{255}{\ diagnosis}%
\tokent{255}{255}{255}{\ and}%
\tokent{255}{255}{255}{\ treatment}%
\tokent{255}{255}{255}{\ of}%
\tokent{255}{255}{255}{\ sleep}%
\tokent{255}{255}{255}{\ disorders}%
\tokent{255}{255}{255}{,}%
\tokent{255}{255}{255}{\ disse}%
\tokent{255}{255}{255}{minate}%
\tokent{255}{255}{255}{\ cutting}%
\tokent{255}{255}{255}{-edge}%
\tokent{255}{255}{255}{\ research}%
\tokent{255}{255}{255}{,}%
\tokent{255}{255}{255}{\ and}%
\tokent{255}{255}{255}{\ promote}%
\tokent{255}{255}{255}{\ the}%
\tokent{255}{255}{255}{\ translation}%
\tokent{255}{255}{255}{\ of}%
\tokent{255}{255}{255}{\ basic}%
\tokent{255}{255}{255}{\ science}%
\tokent{255}{255}{255}{\ into}%
\tokent{255}{255}{255}{\ clinical}%
\tokent{255}{255}{255}{\ practice}%
\tokent{255}{255}{255}{."}%
\tokent{255}{255}{255}{\}\newline }%
\tokent{255}{255}{255}{\textbf{Progress}}%
\tokent{255}{255}{255}{\textbf{:}}%
\tokent{255}{255}{255}{\ \{"}%
\tokent{255}{255}{255}{Go}%
\tokent{255}{255}{255}{\ to}%
\tokent{255}{255}{255}{\ BBC}%
\tokent{255}{255}{255}{\ News}%
\tokent{255}{255}{255}{":}%
\tokent{255}{255}{255}{\ "}%
\tokent{255}{255}{255}{finished}%
\tokent{255}{255}{255}{",}%
\tokent{255}{255}{255}{\ "}%
\tokent{255}{255}{255}{Search}%
\tokent{255}{255}{255}{\ for}%
\tokent{255}{255}{255}{\ technological}%
\tokent{255}{255}{255}{\ advancements}%
\tokent{255}{255}{255}{\ in}%
\tokent{255}{255}{255}{\ medicine}%
\tokent{255}{255}{255}{":}%
\tokent{255}{255}{255}{\ "}%
\tokent{255}{255}{255}{finished}%
\tokent{255}{255}{255}{",}%
\tokent{255}{255}{255}{\ "}%
\tokent{255}{255}{255}{Review}%
\tokent{255}{255}{255}{\ search}%
\tokent{255}{255}{255}{\ results}%
\tokent{255}{255}{255}{":}%
\tokent{255}{255}{255}{\ "}%
\tokent{255}{255}{255}{in}%
\tokent{255}{255}{255}{\ progress}%
\tokent{255}{255}{255}{",}%
\tokent{255}{255}{255}{\ "}%
\tokent{255}{255}{255}{Read}%
\tokent{255}{255}{255}{\ the}%
\tokent{255}{255}{255}{\ article}%
\tokent{255}{255}{255}{":}%
\tokent{255}{255}{255}{\ "}%
\tokent{255}{255}{255}{interrupt}%
\tokent{255}{255}{255}{ed}%
\tokent{255}{255}{255}{\ by}%
\tokent{255}{255}{255}{\ subscription}%
\tokent{255}{255}{255}{\ popup}%
\tokent{255}{255}{255}{\ and}%
\tokent{255}{255}{255}{\ error}%
\tokent{255}{255}{255}{\ page}%
\tokent{255}{255}{255}{",}%
\tokent{255}{255}{255}{\ "}%
\tokent{255}{255}{255}{Sum}%
\tokent{255}{255}{255}{mar}%
\tokent{255}{255}{255}{ize}%
\tokent{255}{255}{255}{\ the}%
\tokent{255}{255}{255}{\ findings}%
\tokent{255}{255}{255}{":}%
\tokent{255}{255}{255}{\ "}%
\tokent{255}{255}{255}{not}%
\tokent{255}{255}{255}{\ finished}%
\tokent{255}{255}{255}{",}%
\tokent{255}{255}{255}{\ "}%
\tokent{255}{255}{255}{Navigate}%
\tokent{255}{255}{255}{\ to}%
\tokent{255}{255}{255}{\ NIH}%
\tokent{255}{255}{255}{\ website}%
\tokent{255}{255}{255}{":}%
\tokent{255}{255}{255}{\ "}%
\tokent{255}{255}{255}{finished}%
\tokent{255}{255}{255}{",}%
\tokent{255}{255}{255}{\ "}%
\tokent{255}{255}{255}{Search}%
\tokent{255}{255}{255}{\ for}%
\tokent{255}{255}{255}{\ articles}%
\tokent{255}{255}{255}{\ on}%
\tokent{255}{255}{255}{\ NIH}%
\tokent{255}{255}{255}{\ website}%
\tokent{255}{255}{255}{":}%
\tokent{255}{255}{255}{\ "}%
\tokent{255}{255}{255}{finished}%
\tokent{255}{255}{255}{",}%
\tokent{255}{255}{255}{\ "}%
\tokent{255}{255}{255}{Review}%
\tokent{255}{255}{255}{\ NIH}%
\tokent{255}{255}{255}{\ search}%
\tokent{255}{255}{255}{\ results}%
\tokent{255}{255}{255}{":}%
\tokent{255}{255}{255}{\ "}%
\tokent{255}{255}{255}{finished}%
\tokent{255}{255}{255}{",}%
\tokent{255}{255}{255}{\ "}%
\tokent{255}{255}{255}{Select}%
\tokent{255}{255}{255}{\ and}%
\tokent{255}{255}{255}{\ read}%
\tokent{255}{255}{255}{\ an}%
\tokent{255}{255}{255}{\ article}%
\tokent{255}{255}{255}{":}%
\tokent{255}{255}{255}{\ "}%
\tokent{255}{255}{255}{finished}%
\tokent{255}{255}{255}{",}%
\tokent{255}{255}{255}{\ "}%
\tokent{255}{255}{255}{Sum}%
\tokent{255}{255}{255}{mar}%
\tokent{255}{255}{255}{ize}%
\tokent{255}{255}{255}{\ the}%
\tokent{255}{255}{255}{\ findings}%
\tokent{255}{255}{255}{\ from}%
\tokent{255}{255}{255}{\ the}%
\tokent{255}{255}{255}{\ selected}%
\tokent{255}{255}{255}{\ article}%
\tokent{255}{255}{255}{":}%
\tokent{255}{255}{255}{\ "}%
\tokent{255}{255}{255}{in}%
\tokent{255}{255}{255}{\ progress}%
\tokent{255}{255}{255}{"\}\newline }%
\tokent{255}{255}{255}{Int}%
\tokent{255}{255}{255}{ention}%
\tokent{255}{255}{255}{:}%
\tokent{255}{255}{255}{\ Provide}%
\tokent{255}{255}{255}{\ the}%
\tokent{255}{255}{255}{\ final}%
\tokent{255}{255}{255}{\ answer}%
\tokent{255}{255}{255}{\ based}%
\tokent{255}{255}{255}{\ on}%
\tokent{255}{255}{255}{\ the}%
\tokent{255}{255}{255}{\ article}%
\tokent{255}{255}{255}{\ summary}%
\tokent{255}{255}{255}{\newline }%
\tokent{255}{255}{255}{\textbf{Action}}%
\tokent{255}{255}{255}{\textbf{:}}%
\tokent{255}{255}{255}{\ The}%
\tokent{255}{255}{255}{\ task}%
\tokent{255}{255}{255}{\ is}%
\tokent{255}{255}{255}{\ completed}%
\tokent{255}{255}{255}{.}%
\tokent{255}{255}{255}{\ The}%
\tokent{255}{255}{255}{\ latest}%
\tokent{255}{255}{255}{\ technological}%
\tokent{255}{255}{255}{\ advancements}%
\tokent{255}{255}{255}{\ in}%
\tokent{255}{255}{255}{\ the}%
\tokent{255}{255}{255}{\ field}%
\tokent{255}{255}{255}{\ of}%
\tokent{255}{255}{255}{\ medicine}%
\tokent{255}{255}{255}{,}%
\tokent{255}{255}{255}{\ as}%
\tokent{255}{255}{255}{\ discussed}%
\tokent{255}{255}{255}{\ in}%
\tokent{255}{255}{255}{\ the}%
\tokent{255}{255}{255}{\ S}%
\tokent{255}{255}{255}{LEEP}%
\tokent{255}{255}{255}{\ }%
\tokent{255}{255}{255}{2}%
\tokent{255}{255}{255}{0}%
\tokent{255}{255}{255}{2}%
\tokent{255}{255}{255}{4}%
\tokent{255}{255}{255}{\ meeting}%
\tokent{255}{255}{255}{,}%
\tokent{255}{255}{255}{\ include}%
\tokent{255}{255}{255}{\ the}%
\tokent{255}{255}{255}{\ latest}%
\tokent{255}{255}{255}{\ advances}%
\tokent{255}{255}{255}{\ in}%
\tokent{255}{255}{255}{\ sleep}%
\tokent{255}{255}{255}{\ science}%
\tokent{255}{255}{255}{\ and}%
\tokent{255}{255}{255}{\ new}%
\tokent{255}{255}{255}{\ developments}%
\tokent{255}{255}{255}{\ in}%
\tokent{255}{255}{255}{\ the}%
\tokent{255}{255}{255}{\ diagnosis}%
\tokent{255}{255}{255}{\ and}%
\tokent{255}{255}{255}{\ treatment}%
\tokent{255}{255}{255}{\ of}%
\tokent{255}{255}{255}{\ sleep}%
\tokent{255}{255}{255}{\ disorders}%
\tokent{255}{255}{255}{,}%
\tokent{255}{255}{255}{\ disse}%
\tokent{255}{255}{255}{minating}%
\tokent{255}{255}{255}{\ cutting}%
\tokent{255}{255}{255}{-edge}%
\tokent{255}{255}{255}{\ research}%
\tokent{255}{255}{255}{,}%
\tokent{255}{255}{255}{\ and}%
\tokent{255}{255}{255}{\ promoting}%
\tokent{255}{255}{255}{\ the}%
\tokent{255}{255}{255}{\ translation}%
\tokent{255}{255}{255}{\ of}%
\tokent{255}{255}{255}{\ basic}%
\tokent{255}{255}{255}{\ science}%
\tokent{255}{255}{255}{\ into}%
\tokent{255}{255}{255}{\ clinical}%
\tokent{255}{255}{255}{\ practice}%
\tokent{255}{255}{255}{.\newline }%
\tokent{255}{255}{255}{\textbf{\textless{}tool\_call\textgreater{}}}%
\tokent{255}{255}{255}{\newline }%
\tokent{255}{255}{255}{\{"}%
\tokent{255}{255}{255}{name}%
\tokent{255}{255}{255}{":}%
\tokent{255}{255}{255}{\ "}%
\tokent{255}{255}{255}{computer}%
\tokent{255}{255}{255}{\_use}%
\tokent{255}{255}{255}{",}%
\tokent{255}{255}{255}{\ "}%
\tokent{255}{255}{255}{arguments}%
\tokent{255}{255}{255}{":}%
\tokent{255}{255}{255}{\ \{"}%
\tokent{255}{255}{255}{action}%
\tokent{255}{255}{255}{":}%
\tokent{255}{255}{255}{\ "}%
\tokent{255}{255}{255}{answer}%
\tokent{255}{255}{255}{",}%
\tokent{255}{255}{255}{\ "}%
\tokent{255}{255}{255}{text}%
\tokent{255}{255}{255}{":}%
\tokent{255}{255}{255}{\ "}%
\tokent{255}{255}{255}{The}%
\tokent{255}{255}{255}{\ latest}%
\tokent{255}{255}{255}{\ technological}%
\tokent{255}{255}{255}{\ advancements}%
\tokent{255}{255}{255}{\ in}%
\tokent{255}{255}{255}{\ the}%
\tokent{255}{255}{255}{\ field}%
\tokent{255}{255}{255}{\ of}%
\tokent{255}{255}{255}{\ medicine}%
\tokent{255}{255}{255}{,}%
\tokent{255}{255}{255}{\ as}%
\tokent{255}{255}{255}{\ discussed}%
\tokent{255}{255}{255}{\ in}%
\tokent{255}{255}{255}{\ the}%
\tokent{255}{255}{255}{\ S}%
\tokent{255}{255}{255}{LEEP}%
\tokent{255}{255}{255}{\ }%
\tokent{255}{255}{255}{2}%
\tokent{255}{255}{255}{0}%
\tokent{255}{255}{255}{2}%
\tokent{255}{255}{255}{4}%
\tokent{255}{255}{255}{\ meeting}%
\tokent{255}{255}{255}{,}%
\tokent{255}{255}{255}{\ include}%
\tokent{255}{255}{255}{\ the}%
\tokent{255}{255}{255}{\ latest}%
\tokent{255}{255}{255}{\ advances}%
\tokent{255}{255}{255}{\ in}%
\tokent{255}{255}{255}{\ sleep}%
\tokent{255}{255}{255}{\ science}%
\tokent{255}{255}{255}{\ and}%
\tokent{255}{255}{255}{\ new}%
\tokent{255}{255}{255}{\ developments}%
\tokent{255}{255}{255}{\ in}%
\tokent{255}{255}{255}{\ the}%
\tokent{255}{255}{255}{\ diagnosis}%
\tokent{255}{255}{255}{\ and}%
\tokent{255}{255}{255}{\ treatment}%
\tokent{255}{255}{255}{\ of}%
\tokent{255}{255}{255}{\ sleep}%
\tokent{255}{255}{255}{\ disorders}%
\tokent{255}{255}{255}{,}%
\tokent{255}{255}{255}{\ disse}%
\tokent{255}{255}{255}{minating}%
\tokent{255}{255}{255}{\ cutting}%
\tokent{255}{255}{255}{-edge}%
\tokent{255}{255}{255}{\ research}%
\tokent{255}{255}{255}{,}%
\tokent{255}{255}{255}{\ and}%
\tokent{255}{255}{255}{\ promoting}%
\tokent{255}{255}{255}{\ the}%
\tokent{255}{255}{255}{\ translation}%
\tokent{255}{255}{255}{\ of}%
\tokent{255}{255}{255}{\ basic}%
\tokent{255}{255}{255}{\ science}%
\tokent{255}{255}{255}{\ into}%
\tokent{255}{255}{255}{\ clinical}%
\tokent{255}{255}{255}{\ practice}%
\tokent{255}{255}{255}{."}%
\tokent{255}{255}{255}{\}\}\newline }%
\tokent{255}{255}{255}{\textbf{\textless{}/tool\_call\textgreater{}}}%
\tokent{255}{255}{255}{\textbf{\textless{}|im\_end|\textgreater{}}}
\end{panelmd}

%% file: figures/vis_thinking.tex
\begin{panelmd}{Base $\cdot$ n\_steps=10 $\cdot$ reward=1.0 $\cdot$ 387 tokens}
\tokent{255}{255}{255}{\textless{}think\textgreater{}}%
\tokent{255}{255}{255}{\newline }%
\tokent{255}{255}{255}{\textbf{Memory}}%
\tokent{255}{255}{255}{\textbf{:}}%
\tokent{255}{255}{255}{\ We}%
\tokent{255}{255}{255}{\ have}%
\tokent{255}{255}{255}{\ confirmed}%
\tokent{255}{255}{255}{\ that}%
\tokent{255}{255}{255}{\ Think}%
\tokent{255}{255}{255}{Tok}%
\tokent{255}{255}{255}{yo}%
\tokent{255}{255}{255}{.com}%
\tokent{255}{255}{255}{\ is}%
\tokent{255}{255}{255}{\ for}%
\tokent{255}{255}{255}{\ sale}%
\tokent{255}{255}{255}{\ on}%
\tokent{255}{255}{255}{\ Huge}%
\tokent{255}{255}{255}{Domains}%
\tokent{255}{255}{255}{\ at}%
\tokent{255}{255}{255}{\ \$}%
\tokent{255}{255}{255}{3}%
\tokent{255}{255}{255}{,}%
\tokent{255}{255}{255}{0}%
\tokent{255}{255}{255}{9}%
\tokent{255}{255}{255}{5}%
\tokent{255}{255}{255}{.}%
\tokent{255}{255}{255}{\ The}%
\tokent{255}{255}{255}{\ page}%
\tokent{255}{255}{255}{\ shows}%
\tokent{255}{255}{255}{\ purchase}%
\tokent{255}{255}{255}{\ options}%
\tokent{255}{255}{255}{\ including}%
\tokent{255}{255}{255}{\ "}%
\tokent{255}{255}{255}{Buy}%
\tokent{255}{255}{255}{\ now}%
\tokent{255}{255}{255}{"}%
\tokent{255}{255}{255}{\ and}%
\tokent{255}{255}{255}{\ "}%
\tokent{255}{255}{255}{Start}%
\tokent{255}{255}{255}{\ payment}%
\tokent{255}{255}{255}{\ plan}%
\tokent{255}{255}{255}{,"}%
\tokent{255}{255}{255}{\ with}%
\tokent{255}{255}{255}{\ additional}%
\tokent{255}{255}{255}{\ details}%
\tokent{255}{255}{255}{\ about}%
\tokent{255}{255}{255}{\ the}%
\tokent{255}{255}{255}{\ price}%
\tokent{255}{255}{255}{\ and}%
\tokent{255}{255}{255}{\ a}%
\tokent{255}{255}{255}{\ money}%
\tokent{255}{255}{255}{-back}%
\tokent{255}{255}{255}{\ guarantee}%
\tokent{255}{255}{255}{.\newline \newline }%
\tokent{255}{255}{255}{\textbf{Progress}}%
\tokent{255}{255}{255}{\textbf{:}}%
\tokent{255}{255}{255}{\ \newline }%
\tokent{255}{255}{255}{-}%
\tokent{255}{255}{255}{\ Task}%
\tokent{255}{255}{255}{:}%
\tokent{255}{255}{255}{\ Determine}%
\tokent{255}{255}{255}{\ the}%
\tokent{255}{255}{255}{\ current}%
\tokent{255}{255}{255}{\ status}%
\tokent{255}{255}{255}{\ of}%
\tokent{255}{255}{255}{\ the}%
\tokent{255}{255}{255}{\ domain}%
\tokent{255}{255}{255}{\ '}%
\tokent{255}{255}{255}{think}%
\tokent{255}{255}{255}{tok}%
\tokent{255}{255}{255}{yo}%
\tokent{255}{255}{255}{.com}%
\tokent{255}{255}{255}{'}%
\tokent{255}{255}{255}{\ and}%
\tokent{255}{255}{255}{\ if}%
\tokent{255}{255}{255}{\ it}%
\tokent{255}{255}{255}{\ is}%
\tokent{255}{255}{255}{\ for}%
\tokent{255}{255}{255}{\ sale}%
\tokent{255}{255}{255}{,}%
\tokent{255}{255}{255}{\ identify}%
\tokent{255}{255}{255}{\ where}%
\tokent{255}{255}{255}{\ it}%
\tokent{255}{255}{255}{\ is}%
\tokent{255}{255}{255}{\ listed}%
\tokent{255}{255}{255}{\ for}%
\tokent{255}{255}{255}{\ sale}%
\tokent{255}{255}{255}{.\newline }%
\tokent{255}{255}{255}{-}%
\tokent{255}{255}{255}{\ Completed}%
\tokent{255}{255}{255}{:}%
\tokent{255}{255}{255}{\ Conf}%
\tokent{255}{255}{255}{irmed}%
\tokent{255}{255}{255}{\ the}%
\tokent{255}{255}{255}{\ domain}%
\tokent{255}{255}{255}{\ is}%
\tokent{255}{255}{255}{\ for}%
\tokent{255}{255}{255}{\ sale}%
\tokent{255}{255}{255}{\ on}%
\tokent{255}{255}{255}{\ Huge}%
\tokent{255}{255}{255}{Domains}%
\tokent{255}{255}{255}{\ with}%
\tokent{255}{255}{255}{\ price}%
\tokent{255}{255}{255}{\ and}%
\tokent{255}{255}{255}{\ purchase}%
\tokent{255}{255}{255}{\ options}%
\tokent{255}{255}{255}{\ identified}%
\tokent{255}{255}{255}{.\newline }%
\tokent{255}{255}{255}{-}%
\tokent{255}{255}{255}{\ Remaining}%
\tokent{255}{255}{255}{:}%
\tokent{255}{255}{255}{\ No}%
\tokent{255}{255}{255}{\ further}%
\tokent{255}{255}{255}{\ steps}%
\tokent{255}{255}{255}{\ needed}%
\tokent{255}{255}{255}{\ since}%
\tokent{255}{255}{255}{\ the}%
\tokent{255}{255}{255}{\ task}%
\tokent{255}{255}{255}{\ requirements}%
\tokent{255}{255}{255}{\ are}%
\tokent{255}{255}{255}{\ met}%
\tokent{255}{255}{255}{.\newline \newline }%
\tokent{255}{255}{255}{Int}%
\tokent{255}{255}{255}{ention}%
\tokent{255}{255}{255}{:}%
\tokent{255}{255}{255}{\ Complete}%
\tokent{255}{255}{255}{\ the}%
\tokent{255}{255}{255}{\ task}%
\tokent{255}{255}{255}{\ by}%
\tokent{255}{255}{255}{\ providing}%
\tokent{255}{255}{255}{\ the}%
\tokent{255}{255}{255}{\ final}%
\tokent{255}{255}{255}{\ answer}%
\tokent{255}{255}{255}{\ based}%
\tokent{255}{255}{255}{\ on}%
\tokent{255}{255}{255}{\ the}%
\tokent{255}{255}{255}{\ observed}%
\tokent{255}{255}{255}{\ information}%
\tokent{255}{255}{255}{.\newline \newline }%
\tokent{255}{255}{255}{\textbf{Action}}%
\tokent{255}{255}{255}{\textbf{:}}%
\tokent{255}{255}{255}{\ Provide}%
\tokent{255}{255}{255}{\ the}%
\tokent{255}{255}{255}{\ final}%
\tokent{255}{255}{255}{\ answer}%
\tokent{255}{255}{255}{.\newline \newline }%
\tokent{255}{255}{255}{Given}%
\tokent{255}{255}{255}{\ the}%
\tokent{255}{255}{255}{\ information}%
\tokent{255}{255}{255}{\ gathered}%
\tokent{255}{255}{255}{\ from}%
\tokent{255}{255}{255}{\ the}%
\tokent{255}{255}{255}{\ page}%
\tokent{255}{255}{255}{,}%
\tokent{255}{255}{255}{\ the}%
\tokent{255}{255}{255}{\ current}%
\tokent{255}{255}{255}{\ status}%
\tokent{255}{255}{255}{\ of}%
\tokent{255}{255}{255}{\ the}%
\tokent{255}{255}{255}{\ domain}%
\tokent{255}{255}{255}{\ is}%
\tokent{255}{255}{255}{\ clear}%
\tokent{255}{255}{255}{:}%
\tokent{255}{255}{255}{\ it}%
\tokent{255}{255}{255}{\ is}%
\tokent{255}{255}{255}{\ for}%
\tokent{255}{255}{255}{\ sale}%
\tokent{255}{255}{255}{\ on}%
\tokent{255}{255}{255}{\ Huge}%
\tokent{255}{255}{255}{Domains}%
\tokent{255}{255}{255}{,}%
\tokent{255}{255}{255}{\ priced}%
\tokent{255}{255}{255}{\ at}%
\tokent{255}{255}{255}{\ \$}%
\tokent{255}{255}{255}{3}%
\tokent{255}{255}{255}{,}%
\tokent{255}{255}{255}{0}%
\tokent{255}{255}{255}{9}%
\tokent{255}{255}{255}{5}%
\tokent{255}{255}{255}{,}%
\tokent{255}{255}{255}{\ with}%
\tokent{255}{255}{255}{\ specific}%
\tokent{255}{255}{255}{\ purchase}%
\tokent{255}{255}{255}{\ options}%
\tokent{255}{255}{255}{\ available}%
\tokent{255}{255}{255}{.}%
\tokent{255}{255}{255}{\ No}%
\tokent{255}{255}{255}{\ additional}%
\tokent{255}{255}{255}{\ actions}%
\tokent{255}{255}{255}{\ are}%
\tokent{255}{255}{255}{\ necessary}%
\tokent{255}{255}{255}{\ because}%
\tokent{255}{255}{255}{\ the}%
\tokent{255}{255}{255}{\ task}%
\tokent{255}{255}{255}{'s}%
\tokent{255}{255}{255}{\ requirements}%
\tokent{255}{255}{255}{\ have}%
\tokent{255}{255}{255}{\ been}%
\tokent{255}{255}{255}{\ fully}%
\tokent{255}{255}{255}{\ satisfied}%
\tokent{255}{255}{255}{.\newline }%
\tokent{255}{255}{255}{\textless{}/think\textgreater{}}%
\tokent{255}{255}{255}{\newline \newline }%
\tokent{255}{255}{255}{\textbf{Memory}}%
\tokent{255}{255}{255}{\textbf{:}}%
\tokent{255}{255}{255}{\ \{"}%
\tokent{255}{255}{255}{Status}%
\tokent{255}{255}{255}{\ of}%
\tokent{255}{255}{255}{\ Think}%
\tokent{255}{255}{255}{Tok}%
\tokent{255}{255}{255}{yo}%
\tokent{255}{255}{255}{.com}%
\tokent{255}{255}{255}{":}%
\tokent{255}{255}{255}{\ "}%
\tokent{255}{255}{255}{For}%
\tokent{255}{255}{255}{\ sale}%
\tokent{255}{255}{255}{",}%
\tokent{255}{255}{255}{\ "}%
\tokent{255}{255}{255}{Price}%
\tokent{255}{255}{255}{":}%
\tokent{255}{255}{255}{\ "\$}%
\tokent{255}{255}{255}{3}%
\tokent{255}{255}{255}{,}%
\tokent{255}{255}{255}{0}%
\tokent{255}{255}{255}{9}%
\tokent{255}{255}{255}{5}%
\tokent{255}{255}{255}{",}%
\tokent{255}{255}{255}{\ "}%
\tokent{255}{255}{255}{Listing}%
\tokent{255}{255}{255}{\ Site}%
\tokent{255}{255}{255}{":}%
\tokent{255}{255}{255}{\ "}%
\tokent{255}{255}{255}{H}%
\tokent{255}{255}{255}{uge}%
\tokent{255}{255}{255}{Domains}%
\tokent{255}{255}{255}{",}%
\tokent{255}{255}{255}{\ "}%
\tokent{255}{255}{255}{Purchase}%
\tokent{255}{255}{255}{\ Options}%
\tokent{255}{255}{255}{":}%
\tokent{255}{255}{255}{\ ["}%
\tokent{255}{255}{255}{Buy}%
\tokent{255}{255}{255}{\ now}%
\tokent{255}{255}{255}{",}%
\tokent{255}{255}{255}{\ "}%
\tokent{255}{255}{255}{Start}%
\tokent{255}{255}{255}{\ payment}%
\tokent{255}{255}{255}{\ plan}%
\tokent{255}{255}{255}{"]}%
\tokent{255}{255}{255}{\}\newline \newline }%
\tokent{255}{255}{255}{\textbf{Progress}}%
\tokent{255}{255}{255}{\textbf{:}}%
\tokent{255}{255}{255}{\ \{"}%
\tokent{255}{255}{255}{Check}%
\tokent{255}{255}{255}{\ domain}%
\tokent{255}{255}{255}{\ status}%
\tokent{255}{255}{255}{":}%
\tokent{255}{255}{255}{\ "}%
\tokent{255}{255}{255}{finished}%
\tokent{255}{255}{255}{",}%
\tokent{255}{255}{255}{\ "}%
\tokent{255}{255}{255}{Ident}%
\tokent{255}{255}{255}{ify}%
\tokent{255}{255}{255}{\ sale}%
\tokent{255}{255}{255}{\ price}%
\tokent{255}{255}{255}{":}%
\tokent{255}{255}{255}{\ "}%
\tokent{255}{255}{255}{finished}%
\tokent{255}{255}{255}{",}%
\tokent{255}{255}{255}{\ "}%
\tokent{255}{255}{255}{D}%
\tokent{255}{255}{255}{etermine}%
\tokent{255}{255}{255}{\ listing}%
\tokent{255}{255}{255}{\ site}%
\tokent{255}{255}{255}{":}%
\tokent{255}{255}{255}{\ "}%
\tokent{255}{255}{255}{finished}%
\tokent{255}{255}{255}{",}%
\tokent{255}{255}{255}{\ "}%
\tokent{255}{255}{255}{Sum}%
\tokent{255}{255}{255}{mar}%
\tokent{255}{255}{255}{ize}%
\tokent{255}{255}{255}{\ findings}%
\tokent{255}{255}{255}{":}%
\tokent{255}{255}{255}{\ "}%
\tokent{255}{255}{255}{not}%
\tokent{255}{255}{255}{\ finished}%
\tokent{255}{255}{255}{"\}\newline \newline }%
\tokent{255}{255}{255}{Int}%
\tokent{255}{255}{255}{ention}%
\tokent{255}{255}{255}{:}%
\tokent{255}{255}{255}{\ Sum}%
\tokent{255}{255}{255}{mar}%
\tokent{255}{255}{255}{ize}%
\tokent{255}{255}{255}{\ the}%
\tokent{255}{255}{255}{\ findings}%
\tokent{255}{255}{255}{\ to}%
\tokent{255}{255}{255}{\ confirm}%
\tokent{255}{255}{255}{\ the}%
\tokent{255}{255}{255}{\ domain}%
\tokent{255}{255}{255}{'s}%
\tokent{255}{255}{255}{\ status}%
\tokent{255}{255}{255}{\ and}%
\tokent{255}{255}{255}{\ sale}%
\tokent{255}{255}{255}{\ details}%
\tokent{255}{255}{255}{\newline \newline }%
\tokent{255}{255}{255}{\textbf{Action}}%
\tokent{255}{255}{255}{\textbf{:}}%
\tokent{255}{255}{255}{\ Provide}%
\tokent{255}{255}{255}{\ the}%
\tokent{255}{255}{255}{\ final}%
\tokent{255}{255}{255}{\ answer}%
\tokent{255}{255}{255}{\ based}%
\tokent{255}{255}{255}{\ on}%
\tokent{255}{255}{255}{\ the}%
\tokent{255}{255}{255}{\ observed}%
\tokent{255}{255}{255}{\ information}%
\tokent{255}{255}{255}{\newline \newline }%
\tokent{255}{255}{255}{\textbf{\textless{}tool\_call\textgreater{}}}%
\tokent{255}{255}{255}{\newline }%
\tokent{255}{255}{255}{\{"}%
\tokent{255}{255}{255}{name}%
\tokent{255}{255}{255}{":}%
\tokent{255}{255}{255}{\ "}%
\tokent{255}{255}{255}{computer}%
\tokent{255}{255}{255}{\_use}%
\tokent{255}{255}{255}{",}%
\tokent{255}{255}{255}{\ "}%
\tokent{255}{255}{255}{arguments}%
\tokent{255}{255}{255}{":}%
\tokent{255}{255}{255}{\ \{"}%
\tokent{255}{255}{255}{action}%
\tokent{255}{255}{255}{":}%
\tokent{255}{255}{255}{\ "}%
\tokent{255}{255}{255}{answer}%
\tokent{255}{255}{255}{",}%
\tokent{255}{255}{255}{\ "}%
\tokent{255}{255}{255}{text}%
\tokent{255}{255}{255}{":}%
\tokent{255}{255}{255}{\ "}%
\tokent{255}{255}{255}{The}%
\tokent{255}{255}{255}{\ domain}%
\tokent{255}{255}{255}{\ '}%
\tokent{255}{255}{255}{think}%
\tokent{255}{255}{255}{tok}%
\tokent{255}{255}{255}{yo}%
\tokent{255}{255}{255}{.com}%
\tokent{255}{255}{255}{'}%
\tokent{255}{255}{255}{\ is}%
\tokent{255}{255}{255}{\ currently}%
\tokent{255}{255}{255}{\ for}%
\tokent{255}{255}{255}{\ sale}%
\tokent{255}{255}{255}{\ on}%
\tokent{255}{255}{255}{\ Huge}%
\tokent{255}{255}{255}{Domains}%
\tokent{255}{255}{255}{\ at}%
\tokent{255}{255}{255}{\ a}%
\tokent{255}{255}{255}{\ price}%
\tokent{255}{255}{255}{\ of}%
\tokent{255}{255}{255}{\ \$}%
\tokent{255}{255}{255}{3}%
\tokent{255}{255}{255}{,}%
\tokent{255}{255}{255}{0}%
\tokent{255}{255}{255}{9}%
\tokent{255}{255}{255}{5}%
\tokent{255}{255}{255}{.}%
\tokent{255}{255}{255}{\ It}%
\tokent{255}{255}{255}{\ is}%
\tokent{255}{255}{255}{\ listed}%
\tokent{255}{255}{255}{\ with}%
\tokent{255}{255}{255}{\ purchase}%
\tokent{255}{255}{255}{\ options}%
\tokent{255}{255}{255}{\ including}%
\tokent{255}{255}{255}{\ '}%
\tokent{255}{255}{255}{Buy}%
\tokent{255}{255}{255}{\ now}%
\tokent{255}{255}{255}{'}%
\tokent{255}{255}{255}{\ and}%
\tokent{255}{255}{255}{\ '}%
\tokent{255}{255}{255}{Start}%
\tokent{255}{255}{255}{\ payment}%
\tokent{255}{255}{255}{\ plan}%
\tokent{255}{255}{255}{'}%
\tokent{255}{255}{255}{."}%
\tokent{255}{255}{255}{\}\}\newline }%
\tokent{255}{255}{255}{\textbf{\textless{}/tool\_call\textgreater{}}}%
\tokent{255}{255}{255}{\textbf{\textless{}|im\_end|\textgreater{}}}
\end{panelmd}

\medskip

\begin{panelmd}{GRPO without length norm $\cdot$ n\_steps=10 $\cdot$ reward=1.0 $\cdot$ 612 tokens}
\tokent{255}{255}{255}{\textless{}think\textgreater{}}%
\tokent{255}{255}{255}{\newline }%
\tokent{255}{255}{255}{Looking}%
\tokent{255}{255}{255}{\ at}%
\tokent{255}{255}{255}{\ the}%
\tokent{255}{255}{255}{\ current}%
\tokent{255}{255}{255}{\ screenshot}%
\tokent{255}{255}{255}{,}%
\tokent{255}{255}{255}{\ I}%
\tokent{255}{255}{255}{\ can}%
\tokent{255}{255}{255}{\ see}%
\tokent{255}{255}{255}{\ we}%
\tokent{255}{255}{255}{'re}%
\tokent{255}{255}{255}{\ on}%
\tokent{255}{255}{255}{\ the}%
\tokent{255}{255}{255}{\ Food}%
\tokent{255}{255}{255}{\ Network}%
\tokent{255}{255}{255}{\ website}%
\tokent{255}{255}{255}{\ viewing}%
\tokent{255}{255}{255}{\ the}%
\tokent{255}{255}{255}{\ Simple}%
\tokent{255}{255}{255}{\ Egg}%
\tokent{255}{255}{255}{\ Drop}%
\tokent{255}{255}{255}{\ Soup}%
\tokent{255}{255}{255}{\ recipe}%
\tokent{255}{255}{255}{.}%
\tokent{255}{255}{255}{\ The}%
\tokent{255}{255}{255}{\ task}%
\tokent{255}{255}{255}{\ is}%
\tokent{255}{255}{255}{\ to}%
\tokent{255}{255}{255}{\ find}%
\tokent{255}{255}{255}{\ a}%
\tokent{255}{255}{255}{\ recipe}%
\tokent{255}{255}{255}{\ for}%
\tokent{255}{255}{255}{\ a}%
\tokent{255}{255}{255}{\ popular}%
\tokent{255}{255}{255}{\ soup}%
\tokent{255}{255}{255}{\ dish}%
\tokent{255}{255}{255}{\ that}%
\tokent{255}{255}{255}{\ includes}%
\tokent{255}{255}{255}{\ a}%
\tokent{255}{255}{255}{\ list}%
\tokent{255}{255}{255}{\ of}%
\tokent{255}{255}{255}{\ ingredients}%
\tokent{255}{255}{255}{\ and}%
\tokent{255}{255}{255}{\ cooking}%
\tokent{255}{255}{255}{\ time}%
\tokent{255}{255}{255}{.\newline \newline }%
\tokent{255}{255}{255}{From}%
\tokent{255}{255}{255}{\ the}%
\tokent{255}{255}{255}{\ previous}%
\tokent{255}{255}{255}{\ steps}%
\tokent{255}{255}{255}{,}%
\tokent{255}{255}{255}{\ I}%
\tokent{255}{255}{255}{\ can}%
\tokent{255}{255}{255}{\ see}%
\tokent{255}{255}{255}{\ that}%
\tokent{255}{255}{255}{:\newline }%
\tokent{255}{255}{255}{1}%
\tokent{255}{255}{255}{.}%
\tokent{255}{255}{255}{\ We}%
\tokent{255}{255}{255}{\ successfully}%
\tokent{255}{255}{255}{\ navig}%
\tokent{255}{255}{255}{ated}%
\tokent{255}{255}{255}{\ to}%
\tokent{255}{255}{255}{\ Food}%
\tokent{255}{255}{255}{\ Network}%
\tokent{255}{255}{255}{\newline }%
\tokent{255}{255}{255}{2}%
\tokent{255}{255}{255}{.}%
\tokent{255}{255}{255}{\ We}%
\tokent{255}{255}{255}{\ searched}%
\tokent{255}{255}{255}{\ for}%
\tokent{255}{255}{255}{\ "}%
\tokent{255}{255}{255}{popular}%
\tokent{255}{255}{255}{\ soup}%
\tokent{255}{255}{255}{\ recipes}%
\tokent{255}{255}{255}{"\newline }%
\tokent{255}{255}{255}{3}%
\tokent{255}{255}{255}{.}%
\tokent{255}{255}{255}{\ We}%
\tokent{255}{255}{255}{\ selected}%
\tokent{255}{255}{255}{\ the}%
\tokent{255}{255}{255}{\ "}%
\tokent{255}{255}{255}{Simple}%
\tokent{255}{255}{255}{\ Egg}%
\tokent{255}{255}{255}{\ Drop}%
\tokent{255}{255}{255}{\ Soup}%
\tokent{255}{255}{255}{"}%
\tokent{255}{255}{255}{\ recipe}%
\tokent{255}{255}{255}{\newline }%
\tokent{255}{255}{255}{4}%
\tokent{255}{255}{255}{.}%
\tokent{255}{255}{255}{\ We}%
\tokent{255}{255}{255}{'ve}%
\tokent{255}{255}{255}{\ now}%
\tokent{255}{255}{255}{\ scrolled}%
\tokent{255}{255}{255}{\ down}%
\tokent{255}{255}{255}{\ to}%
\tokent{255}{255}{255}{\ see}%
\tokent{255}{255}{255}{\ the}%
\tokent{255}{255}{255}{\ recipe}%
\tokent{255}{255}{255}{\ details}%
\tokent{255}{255}{255}{\newline \newline }%
\tokent{255}{255}{255}{The}%
\tokent{255}{255}{255}{\ current}%
\tokent{255}{255}{255}{\ screenshot}%
\tokent{255}{255}{255}{\ shows}%
\tokent{255}{255}{255}{\ the}%
\tokent{255}{255}{255}{\ recipe}%
\tokent{255}{255}{255}{\ page}%
\tokent{255}{255}{255}{\ with}%
\tokent{255}{255}{255}{:\newline }%
\tokent{255}{255}{255}{-}%
\tokent{255}{255}{255}{\ The}%
\tokent{255}{255}{255}{\ recipe}%
\tokent{255}{255}{255}{\ title}%
\tokent{255}{255}{255}{\ "}%
\tokent{255}{255}{255}{No}%
\tokent{255}{255}{255}{\ Recipe}%
\tokent{255}{255}{255}{\ Recipe}%
\tokent{255}{255}{255}{:}%
\tokent{255}{255}{255}{\ Simple}%
\tokent{255}{255}{255}{\ Egg}%
\tokent{255}{255}{255}{\ Drop}%
\tokent{255}{255}{255}{\ Soup}%
\tokent{255}{255}{255}{"\newline }%
\tokent{255}{255}{255}{-}%
\tokent{255}{255}{255}{\ Level}%
\tokent{255}{255}{255}{:}%
\tokent{255}{255}{255}{\ Easy}%
\tokent{255}{255}{255}{\newline }%
\tokent{255}{255}{255}{-}%
\tokent{255}{255}{255}{\ Prep}%
\tokent{255}{255}{255}{:}%
\tokent{255}{255}{255}{\ }%
\tokent{255}{255}{255}{2}%
\tokent{255}{255}{255}{\ min}%
\tokent{255}{255}{255}{\newline }%
\tokent{255}{255}{255}{-}%
\tokent{255}{255}{255}{\ Cook}%
\tokent{255}{255}{255}{:}%
\tokent{255}{255}{255}{\ }%
\tokent{255}{255}{255}{1}%
\tokent{255}{255}{255}{0}%
\tokent{255}{255}{255}{\ min}%
\tokent{255}{255}{255}{\newline }%
\tokent{255}{255}{255}{-}%
\tokent{255}{255}{255}{\ Total}%
\tokent{255}{255}{255}{:}%
\tokent{255}{255}{255}{\ }%
\tokent{255}{255}{255}{1}%
\tokent{255}{255}{255}{2}%
\tokent{255}{255}{255}{\ min}%
\tokent{255}{255}{255}{\newline }%
\tokent{255}{255}{255}{-}%
\tokent{255}{255}{255}{\ Ingredients}%
\tokent{255}{255}{255}{\ list}%
\tokent{255}{255}{255}{\ (}%
\tokent{255}{255}{255}{which}%
\tokent{255}{255}{255}{\ includes}%
\tokent{255}{255}{255}{\ chicken}%
\tokent{255}{255}{255}{\ broth}%
\tokent{255}{255}{255}{,}%
\tokent{255}{255}{255}{\ soy}%
\tokent{255}{255}{255}{\ sauce}%
\tokent{255}{255}{255}{,}%
\tokent{255}{255}{255}{\ Asian}%
\tokent{255}{255}{255}{\ hot}%
\tokent{255}{255}{255}{\ sauce}%
\tokent{255}{255}{255}{,}%
\tokent{255}{255}{255}{\ fresh}%
\tokent{255}{255}{255}{\ ginger}%
\tokent{255}{255}{255}{,}%
\tokent{255}{255}{255}{\ and}%
\tokent{255}{255}{255}{\ eggs}%
\tokent{255}{255}{255}{)\newline }%
\tokent{255}{255}{255}{-}%
\tokent{255}{255}{255}{\ Directions}%
\tokent{255}{255}{255}{\ for}%
\tokent{255}{255}{255}{\ making}%
\tokent{255}{255}{255}{\ the}%
\tokent{255}{255}{255}{\ soup}%
\tokent{255}{255}{255}{\newline \newline }%
\tokent{255}{255}{255}{The}%
\tokent{255}{255}{255}{\ task}%
\tokent{255}{255}{255}{\ requires}%
\tokent{255}{255}{255}{\ finding}%
\tokent{255}{255}{255}{\ a}%
\tokent{255}{255}{255}{\ recipe}%
\tokent{255}{255}{255}{\ with}%
\tokent{255}{255}{255}{\ a}%
\tokent{255}{255}{255}{\ list}%
\tokent{255}{255}{255}{\ of}%
\tokent{255}{255}{255}{\ ingredients}%
\tokent{255}{255}{255}{\ and}%
\tokent{255}{255}{255}{\ cooking}%
\tokent{255}{255}{255}{\ time}%
\tokent{255}{255}{255}{.}%
\tokent{255}{255}{255}{\ Based}%
\tokent{255}{255}{255}{\ on}%
\tokent{255}{255}{255}{\ what}%
\tokent{255}{255}{255}{'s}%
\tokent{255}{255}{255}{\ visible}%
\tokent{255}{255}{255}{\ in}%
\tokent{255}{255}{255}{\ the}%
\tokent{255}{255}{255}{\ screenshot}%
\tokent{255}{255}{255}{,}%
\tokent{255}{255}{255}{\ we}%
\tokent{255}{255}{255}{\ have}%
\tokent{255}{255}{255}{\ both}%
\tokent{255}{255}{255}{:\newline }%
\tokent{255}{255}{255}{-}%
\tokent{255}{255}{255}{\ The}%
\tokent{255}{255}{255}{\ cooking}%
\tokent{255}{255}{255}{\ time}%
\tokent{255}{255}{255}{\ is}%
\tokent{255}{255}{255}{\ clearly}%
\tokent{255}{255}{255}{\ shown}%
\tokent{255}{255}{255}{\ as}%
\tokent{255}{255}{255}{\ "}%
\tokent{255}{255}{255}{Total}%
\tokent{255}{255}{255}{:}%
\tokent{255}{255}{255}{\ }%
\tokent{255}{255}{255}{1}%
\tokent{255}{255}{255}{2}%
\tokent{255}{255}{255}{\ min}%
\tokent{255}{255}{255}{"}%
\tokent{255}{255}{255}{\ (}%
\tokent{255}{255}{255}{with}%
\tokent{255}{255}{255}{\ Prep}%
\tokent{255}{255}{255}{:}%
\tokent{255}{255}{255}{\ }%
\tokent{255}{255}{255}{2}%
\tokent{255}{255}{255}{\ min}%
\tokent{255}{255}{255}{\ and}%
\tokent{255}{255}{255}{\ Cook}%
\tokent{255}{255}{255}{:}%
\tokent{255}{255}{255}{\ }%
\tokent{255}{255}{255}{1}%
\tokent{255}{255}{255}{0}%
\tokent{255}{255}{255}{\ min}%
\tokent{255}{255}{255}{)\newline }%
\tokent{255}{255}{255}{-}%
\tokent{255}{255}{255}{\ The}%
\tokent{255}{255}{255}{\ ingredients}%
\tokent{255}{255}{255}{\ list}%
\tokent{255}{255}{255}{\ is}%
\tokent{255}{255}{255}{\ clearly}%
\tokent{255}{255}{255}{\ visible}%
\tokent{255}{255}{255}{\ in}%
\tokent{255}{255}{255}{\ the}%
\tokent{255}{255}{255}{\ "}%
\tokent{255}{255}{255}{Ingredients}%
\tokent{255}{255}{255}{:"}%
\tokent{255}{255}{255}{\ section}%
\tokent{255}{255}{255}{\newline \newline }%
\tokent{255}{255}{255}{Since}%
\tokent{255}{255}{255}{\ we}%
\tokent{255}{255}{255}{'ve}%
\tokent{255}{255}{255}{\ found}%
\tokent{255}{255}{255}{\ a}%
\tokent{255}{255}{255}{\ soup}%
\tokent{255}{255}{255}{\ recipe}%
\tokent{255}{255}{255}{\ with}%
\tokent{255}{255}{255}{\ both}%
\tokent{255}{255}{255}{\ the}%
\tokent{255}{255}{255}{\ ingredients}%
\tokent{255}{255}{255}{\ list}%
\tokent{255}{255}{255}{\ and}%
\tokent{255}{255}{255}{\ cooking}%
\tokent{255}{255}{255}{\ time}%
\tokent{255}{255}{255}{,}%
\tokent{255}{255}{255}{\ the}%
\tokent{255}{255}{255}{\ task}%
\tokent{255}{255}{255}{\ is}%
\tokent{255}{255}{255}{\ complete}%
\tokent{255}{255}{255}{.}%
\tokent{255}{255}{255}{\ The}%
\tokent{255}{255}{255}{\ recipe}%
\tokent{255}{255}{255}{\ meets}%
\tokent{255}{255}{255}{\ all}%
\tokent{255}{255}{255}{\ the}%
\tokent{255}{255}{255}{\ criteria}%
\tokent{255}{255}{255}{\ specified}%
\tokent{255}{255}{255}{\ in}%
\tokent{255}{255}{255}{\ the}%
\tokent{255}{255}{255}{\ task}%
\tokent{255}{255}{255}{.\newline \newline }%
\tokent{255}{255}{255}{The}%
\tokent{255}{255}{255}{\ next}%
\tokent{255}{255}{255}{\ action}%
\tokent{255}{255}{255}{\ should}%
\tokent{255}{255}{255}{\ be}%
\tokent{255}{255}{255}{\ to}%
\tokent{255}{255}{255}{\ confirm}%
\tokent{255}{255}{255}{\ that}%
\tokent{255}{255}{255}{\ we}%
\tokent{255}{255}{255}{'ve}%
\tokent{255}{255}{255}{\ successfully}%
\tokent{255}{255}{255}{\ completed}%
\tokent{255}{255}{255}{\ the}%
\tokent{255}{255}{255}{\ task}%
\tokent{255}{255}{255}{.\newline }%
\tokent{255}{255}{255}{\textless{}/think\textgreater{}}%
\tokent{255}{255}{255}{\newline \newline }%
\tokent{255}{255}{255}{\textbf{Memory}}%
\tokent{255}{255}{255}{\textbf{:}}%
\tokent{255}{255}{255}{\ \{"}%
\tokent{255}{255}{255}{Current}%
\tokent{255}{255}{255}{\ page}%
\tokent{255}{255}{255}{":}%
\tokent{255}{255}{255}{\ "}%
\tokent{255}{255}{255}{Food}%
\tokent{255}{255}{255}{\ Network}%
\tokent{255}{255}{255}{\ website}%
\tokent{255}{255}{255}{",}%
\tokent{255}{255}{255}{\ "}%
\tokent{255}{255}{255}{Task}%
\tokent{255}{255}{255}{":}%
\tokent{255}{255}{255}{\ "}%
\tokent{255}{255}{255}{Find}%
\tokent{255}{255}{255}{\ a}%
\tokent{255}{255}{255}{\ recipe}%
\tokent{255}{255}{255}{\ for}%
\tokent{255}{255}{255}{\ a}%
\tokent{255}{255}{255}{\ popular}%
\tokent{255}{255}{255}{\ soup}%
\tokent{255}{255}{255}{\ dish}%
\tokent{255}{255}{255}{\ that}%
\tokent{255}{255}{255}{\ includes}%
\tokent{255}{255}{255}{\ a}%
\tokent{255}{255}{255}{\ list}%
\tokent{255}{255}{255}{\ of}%
\tokent{255}{255}{255}{\ ingredients}%
\tokent{255}{255}{255}{\ and}%
\tokent{255}{255}{255}{\ cooking}%
\tokent{255}{255}{255}{\ time}%
\tokent{255}{255}{255}{",}%
\tokent{255}{255}{255}{\ "}%
\tokent{255}{255}{255}{Search}%
\tokent{255}{255}{255}{\ term}%
\tokent{255}{255}{255}{":}%
\tokent{255}{255}{255}{\ "}%
\tokent{255}{255}{255}{popular}%
\tokent{255}{255}{255}{\ soup}%
\tokent{255}{255}{255}{\ recipes}%
\tokent{255}{255}{255}{",}%
\tokent{255}{255}{255}{\ "}%
\tokent{255}{255}{255}{Results}%
\tokent{255}{255}{255}{":}%
\tokent{255}{255}{255}{\ ["}%
\tokent{255}{255}{255}{No}%
\tokent{255}{255}{255}{\ Recipe}%
\tokent{255}{255}{255}{\ Recipe}%
\tokent{255}{255}{255}{:}%
\tokent{255}{255}{255}{\ Simple}%
\tokent{255}{255}{255}{\ Egg}%
\tokent{255}{255}{255}{\ Drop}%
\tokent{255}{255}{255}{\ Soup}%
\tokent{255}{255}{255}{\ (}%
\tokent{255}{255}{255}{1}%
\tokent{255}{255}{255}{2}%
\tokent{255}{255}{255}{\ minutes}%
\tokent{255}{255}{255}{)",}%
\tokent{255}{255}{255}{\ "}%
\tokent{255}{255}{255}{Good}%
\tokent{255}{255}{255}{\ E}%
\tokent{255}{255}{255}{ats}%
\tokent{255}{255}{255}{\ Ro}%
\tokent{255}{255}{255}{ast}%
\tokent{255}{255}{255}{\ Turkey}%
\tokent{255}{255}{255}{"],}%
\tokent{255}{255}{255}{\ "}%
\tokent{255}{255}{255}{Selected}%
\tokent{255}{255}{255}{\ recipe}%
\tokent{255}{255}{255}{":}%
\tokent{255}{255}{255}{\ "}%
\tokent{255}{255}{255}{Simple}%
\tokent{255}{255}{255}{\ Egg}%
\tokent{255}{255}{255}{\ Drop}%
\tokent{255}{255}{255}{\ Soup}%
\tokent{255}{255}{255}{",}%
\tokent{255}{255}{255}{\ "}%
\tokent{255}{255}{255}{Cook}%
\tokent{255}{255}{255}{ing}%
\tokent{255}{255}{255}{\ time}%
\tokent{255}{255}{255}{":}%
\tokent{255}{255}{255}{\ "}%
\tokent{255}{255}{255}{1}%
\tokent{255}{255}{255}{2}%
\tokent{255}{255}{255}{\ minutes}%
\tokent{255}{255}{255}{",}%
\tokent{255}{255}{255}{\ "}%
\tokent{255}{255}{255}{Ingredients}%
\tokent{255}{255}{255}{":}%
\tokent{255}{255}{255}{\ ["}%
\tokent{255}{255}{255}{low}%
\tokent{255}{255}{255}{-s}%
\tokent{255}{255}{255}{odium}%
\tokent{255}{255}{255}{\ chicken}%
\tokent{255}{255}{255}{\ broth}%
\tokent{255}{255}{255}{",}%
\tokent{255}{255}{255}{\ "}%
\tokent{255}{255}{255}{so}%
\tokent{255}{255}{255}{y}%
\tokent{255}{255}{255}{\ sauce}%
\tokent{255}{255}{255}{",}%
\tokent{255}{255}{255}{\ "}%
\tokent{255}{255}{255}{Asian}%
\tokent{255}{255}{255}{\ hot}%
\tokent{255}{255}{255}{\ sauce}%
\tokent{255}{255}{255}{",}%
\tokent{255}{255}{255}{\ "}%
\tokent{255}{255}{255}{fresh}%
\tokent{255}{255}{255}{\ ginger}%
\tokent{255}{255}{255}{",}%
\tokent{255}{255}{255}{\ "}%
\tokent{255}{255}{255}{eg}%
\tokent{255}{255}{255}{gs}%
\tokent{255}{255}{255}{"]}%
\tokent{255}{255}{255}{\}\newline \newline }%
\tokent{255}{255}{255}{\textbf{Progress}}%
\tokent{255}{255}{255}{\textbf{:}}%
\tokent{255}{255}{255}{\ \{"}%
\tokent{255}{255}{255}{Access}%
\tokent{255}{255}{255}{\ All}%
\tokent{255}{255}{255}{recipes}%
\tokent{255}{255}{255}{":}%
\tokent{255}{255}{255}{\ "}%
\tokent{255}{255}{255}{failed}%
\tokent{255}{255}{255}{",}%
\tokent{255}{255}{255}{\ "}%
\tokent{255}{255}{255}{Access}%
\tokent{255}{255}{255}{\ Food}%
\tokent{255}{255}{255}{\ Network}%
\tokent{255}{255}{255}{":}%
\tokent{255}{255}{255}{\ "}%
\tokent{255}{255}{255}{success}%
\tokent{255}{255}{255}{",}%
\tokent{255}{255}{255}{\ "}%
\tokent{255}{255}{255}{Search}%
\tokent{255}{255}{255}{\ for}%
\tokent{255}{255}{255}{\ soup}%
\tokent{255}{255}{255}{\ recipes}%
\tokent{255}{255}{255}{":}%
\tokent{255}{255}{255}{\ "}%
\tokent{255}{255}{255}{completed}%
\tokent{255}{255}{255}{",}%
\tokent{255}{255}{255}{\ "}%
\tokent{255}{255}{255}{Selected}%
\tokent{255}{255}{255}{\ recipe}%
\tokent{255}{255}{255}{":}%
\tokent{255}{255}{255}{\ "}%
\tokent{255}{255}{255}{Simple}%
\tokent{255}{255}{255}{\ Egg}%
\tokent{255}{255}{255}{\ Drop}%
\tokent{255}{255}{255}{\ Soup}%
\tokent{255}{255}{255}{",}%
\tokent{255}{255}{255}{\ "}%
\tokent{255}{255}{255}{Cook}%
\tokent{255}{255}{255}{ing}%
\tokent{255}{255}{255}{\ time}%
\tokent{255}{255}{255}{\ found}%
\tokent{255}{255}{255}{":}%
\tokent{255}{255}{255}{\ "}%
\tokent{255}{255}{255}{1}%
\tokent{255}{255}{255}{2}%
\tokent{255}{255}{255}{\ minutes}%
\tokent{255}{255}{255}{",}%
\tokent{255}{255}{255}{\ "}%
\tokent{255}{255}{255}{Ingredients}%
\tokent{255}{255}{255}{\ list}%
\tokent{255}{255}{255}{\ found}%
\tokent{255}{255}{255}{":}%
\tokent{255}{255}{255}{\ "}%
\tokent{255}{255}{255}{yes}%
\tokent{255}{255}{255}{"\}\newline \newline }%
\tokent{255}{255}{255}{Int}%
\tokent{255}{255}{255}{ention}%
\tokent{255}{255}{255}{:}%
\tokent{255}{255}{255}{\ The}%
\tokent{255}{255}{255}{\ task}%
\tokent{255}{255}{255}{\ is}%
\tokent{255}{255}{255}{\ complete}%
\tokent{255}{255}{255}{\ -}%
\tokent{255}{255}{255}{\ we}%
\tokent{255}{255}{255}{'ve}%
\tokent{255}{255}{255}{\ found}%
\tokent{255}{255}{255}{\ a}%
\tokent{255}{255}{255}{\ soup}%
\tokent{255}{255}{255}{\ recipe}%
\tokent{255}{255}{255}{\ with}%
\tokent{255}{255}{255}{\ ingredients}%
\tokent{255}{255}{255}{\ and}%
\tokent{255}{255}{255}{\ cooking}%
\tokent{255}{255}{255}{\ time}%
\tokent{255}{255}{255}{\newline \newline }%
\tokent{255}{255}{255}{\textbf{Action}}%
\tokent{255}{255}{255}{\textbf{:}}%
\tokent{255}{255}{255}{\ The}%
\tokent{255}{255}{255}{\ task}%
\tokent{255}{255}{255}{\ is}%
\tokent{255}{255}{255}{\ complete}%
\tokent{255}{255}{255}{.}%
\tokent{255}{255}{255}{\ We}%
\tokent{255}{255}{255}{'ve}%
\tokent{255}{255}{255}{\ found}%
\tokent{255}{255}{255}{\ a}%
\tokent{255}{255}{255}{\ popular}%
\tokent{255}{255}{255}{\ soup}%
\tokent{255}{255}{255}{\ recipe}%
\tokent{255}{255}{255}{\ (}%
\tokent{255}{255}{255}{Simple}%
\tokent{255}{255}{255}{\ Egg}%
\tokent{255}{255}{255}{\ Drop}%
\tokent{255}{255}{255}{\ Soup}%
\tokent{255}{255}{255}{)}%
\tokent{255}{255}{255}{\ with}%
\tokent{255}{255}{255}{\ a}%
\tokent{255}{255}{255}{\ list}%
\tokent{255}{255}{255}{\ of}%
\tokent{255}{255}{255}{\ ingredients}%
\tokent{255}{255}{255}{\ and}%
\tokent{255}{255}{255}{\ cooking}%
\tokent{255}{255}{255}{\ time}%
\tokent{255}{255}{255}{\ (}%
\tokent{255}{255}{255}{1}%
\tokent{255}{255}{255}{2}%
\tokent{255}{255}{255}{\ minutes}%
\tokent{255}{255}{255}{\ total}%
\tokent{255}{255}{255}{).\newline }%
\tokent{255}{255}{255}{\textbf{\textless{}tool\_call\textgreater{}}}%
\tokent{255}{255}{255}{\newline }%
\tokent{255}{255}{255}{\{"}%
\tokent{255}{255}{255}{name}%
\tokent{255}{255}{255}{":}%
\tokent{255}{255}{255}{\ "}%
\tokent{255}{255}{255}{computer}%
\tokent{255}{255}{255}{\_use}%
\tokent{255}{255}{255}{",}%
\tokent{255}{255}{255}{\ "}%
\tokent{255}{255}{255}{arguments}%
\tokent{255}{255}{255}{":}%
\tokent{255}{255}{255}{\ \{"}%
\tokent{255}{255}{255}{action}%
\tokent{255}{255}{255}{":}%
\tokent{255}{255}{255}{\ "}%
\tokent{255}{255}{255}{answer}%
\tokent{255}{255}{255}{",}%
\tokent{255}{255}{255}{\ "}%
\tokent{255}{255}{255}{text}%
\tokent{255}{255}{255}{":}%
\tokent{255}{255}{255}{\ "}%
\tokent{255}{255}{255}{The}%
\tokent{255}{255}{255}{\ Simple}%
\tokent{255}{255}{255}{\ Egg}%
\tokent{255}{255}{255}{\ Drop}%
\tokent{255}{255}{255}{\ Soup}%
\tokent{255}{255}{255}{\ recipe}%
\tokent{255}{255}{255}{\ from}%
\tokent{255}{255}{255}{\ Food}%
\tokent{255}{255}{255}{\ Network}%
\tokent{255}{255}{255}{\ has}%
\tokent{255}{255}{255}{\ a}%
\tokent{255}{255}{255}{\ cooking}%
\tokent{255}{255}{255}{\ time}%
\tokent{255}{255}{255}{\ of}%
\tokent{255}{255}{255}{\ }%
\tokent{255}{255}{255}{1}%
\tokent{255}{255}{255}{2}%
\tokent{255}{255}{255}{\ minutes}%
\tokent{255}{255}{255}{\ total}%
\tokent{255}{255}{255}{\ (}%
\tokent{255}{255}{255}{2}%
\tokent{255}{255}{255}{\ min}%
\tokent{255}{255}{255}{\ prep}%
\tokent{255}{255}{255}{,}%
\tokent{255}{255}{255}{\ }%
\tokent{255}{255}{255}{1}%
\tokent{255}{255}{255}{0}%
\tokent{255}{255}{255}{\ min}%
\tokent{255}{255}{255}{\ cook}%
\tokent{255}{255}{255}{)}%
\tokent{255}{255}{255}{\ and}%
\tokent{255}{255}{255}{\ includes}%
\tokent{255}{255}{255}{\ the}%
\tokent{255}{255}{255}{\ following}%
\tokent{255}{255}{255}{\ ingredients}%
\tokent{255}{255}{255}{:}%
\tokent{255}{255}{255}{\ low}%
\tokent{255}{255}{255}{-s}%
\tokent{255}{255}{255}{odium}%
\tokent{255}{255}{255}{\ chicken}%
\tokent{255}{255}{255}{\ broth}%
\tokent{255}{255}{255}{,}%
\tokent{255}{255}{255}{\ soy}%
\tokent{255}{255}{255}{\ sauce}%
\tokent{255}{255}{255}{,}%
\tokent{255}{255}{255}{\ Asian}%
\tokent{255}{255}{255}{\ hot}%
\tokent{255}{255}{255}{\ sauce}%
\tokent{255}{255}{255}{,}%
\tokent{255}{255}{255}{\ fresh}%
\tokent{255}{255}{255}{\ ginger}%
\tokent{255}{255}{255}{,}%
\tokent{255}{255}{255}{\ and}%
\tokent{255}{255}{255}{\ eggs}%
\tokent{255}{255}{255}{."}%
\tokent{255}{255}{255}{\}\}\newline }%
\tokent{255}{255}{255}{\textbf{\textless{}/tool\_call\textgreater{}}}%
\tokent{255}{255}{255}{\textbf{\textless{}|im\_end|\textgreater{}}}
\end{panelmd}

\medskip

\begin{panelmd}{GRPO with length norm $\cdot$ n\_steps=10 $\cdot$ reward=1.0 $\cdot$ 911 tokens}
\tokent{255}{255}{255}{\textless{}think\textgreater{}}%
\tokent{255}{255}{255}{\newline }%
\tokent{255}{255}{255}{Let}%
\tokent{255}{255}{255}{\ me}%
\tokent{255}{255}{255}{\ analyze}%
\tokent{255}{255}{255}{\ the}%
\tokent{255}{255}{255}{\ current}%
\tokent{255}{255}{255}{\ state}%
\tokent{255}{255}{255}{\ and}%
\tokent{255}{255}{255}{\ what}%
\tokent{255}{255}{255}{\ we}%
\tokent{255}{255}{255}{\ need}%
\tokent{255}{255}{255}{\ to}%
\tokent{255}{255}{255}{\ do}%
\tokent{255}{255}{255}{:\newline \newline }%
\tokent{255}{255}{255}{1}%
\tokent{255}{255}{255}{.}%
\tokent{255}{255}{255}{\ The}%
\tokent{255}{255}{255}{\ task}%
\tokent{255}{255}{255}{\ is}%
\tokent{255}{255}{255}{\ to}%
\tokent{255}{255}{255}{\ find}%
\tokent{255}{255}{255}{\ the}%
\tokent{255}{255}{255}{\ most}%
\tokent{255}{255}{255}{\ common}%
\tokent{255}{255}{255}{\ tree}%
\tokent{255}{255}{255}{\ species}%
\tokent{255}{255}{255}{\ in}%
\tokent{255}{255}{255}{\ Central}%
\tokent{255}{255}{255}{\ Park}%
\tokent{255}{255}{255}{\ according}%
\tokent{255}{255}{255}{\ to}%
\tokent{255}{255}{255}{\ map}%
\tokent{255}{255}{255}{\ data}%
\tokent{255}{255}{255}{.\newline }%
\tokent{255}{255}{255}{2}%
\tokent{255}{255}{255}{.}%
\tokent{255}{255}{255}{\ We}%
\tokent{255}{255}{255}{'ve}%
\tokent{255}{255}{255}{\ been}%
\tokent{255}{255}{255}{\ trying}%
\tokent{255}{255}{255}{\ to}%
\tokent{255}{255}{255}{\ access}%
\tokent{255}{255}{255}{\ the}%
\tokent{255}{255}{255}{\ NYC}%
\tokent{255}{255}{255}{\ Tree}%
\tokent{255}{255}{255}{\ Map}%
\tokent{255}{255}{255}{\ website}%
\tokent{255}{255}{255}{\ (}%
\tokent{255}{255}{255}{https}%
\tokent{255}{255}{255}{://}%
\tokent{255}{255}{255}{tree}%
\tokent{255}{255}{255}{-map}%
\tokent{255}{255}{255}{.n}%
\tokent{255}{255}{255}{yc}%
\tokent{255}{255}{255}{gov}%
\tokent{255}{255}{255}{p}%
\tokent{255}{255}{255}{arks}%
\tokent{255}{255}{255}{.org}%
\tokent{255}{255}{255}{)}%
\tokent{255}{255}{255}{\ but}%
\tokent{255}{255}{255}{\ encountered}%
\tokent{255}{255}{255}{\ permission}%
\tokent{255}{255}{255}{\ issues}%
\tokent{255}{255}{255}{.\newline }%
\tokent{255}{255}{255}{3}%
\tokent{255}{255}{255}{.}%
\tokent{255}{255}{255}{\ We}%
\tokent{255}{255}{255}{'ve}%
\tokent{255}{255}{255}{\ switched}%
\tokent{255}{255}{255}{\ to}%
\tokent{255}{255}{255}{\ using}%
\tokent{255}{255}{255}{\ Wikipedia}%
\tokent{255}{255}{255}{'s}%
\tokent{255}{255}{255}{\ Central}%
\tokent{255}{255}{255}{\ Park}%
\tokent{255}{255}{255}{\ page}%
\tokent{255}{255}{255}{\ as}%
\tokent{255}{255}{255}{\ an}%
\tokent{255}{255}{255}{\ alternative}%
\tokent{255}{255}{255}{\ source}%
\tokent{255}{255}{255}{.\newline }%
\tokent{255}{255}{255}{4}%
\tokent{255}{255}{255}{.}%
\tokent{255}{255}{255}{\ We}%
\tokent{255}{255}{255}{'ve}%
\tokent{255}{255}{255}{\ scrolled}%
\tokent{255}{255}{255}{\ through}%
\tokent{255}{255}{255}{\ the}%
\tokent{255}{255}{255}{\ Trees}%
\tokent{255}{255}{255}{\ and}%
\tokent{255}{255}{255}{\ plants}%
\tokent{255}{255}{255}{\ section}%
\tokent{255}{255}{255}{\ and}%
\tokent{255}{255}{255}{\ found}%
\tokent{255}{255}{255}{\ some}%
\tokent{255}{255}{255}{\ information}%
\tokent{255}{255}{255}{\ about}%
\tokent{255}{255}{255}{\ tree}%
\tokent{255}{255}{255}{\ species}%
\tokent{255}{255}{255}{\ in}%
\tokent{255}{255}{255}{\ Central}%
\tokent{255}{255}{255}{\ Park}%
\tokent{255}{255}{255}{.\newline \newline }%
\tokent{255}{255}{255}{From}%
\tokent{255}{255}{255}{\ the}%
\tokent{255}{255}{255}{\ current}%
\tokent{255}{255}{255}{\ screenshot}%
\tokent{255}{255}{255}{,}%
\tokent{255}{255}{255}{\ I}%
\tokent{255}{255}{255}{\ can}%
\tokent{255}{255}{255}{\ see}%
\tokent{255}{255}{255}{\ we}%
\tokent{255}{255}{255}{'re}%
\tokent{255}{255}{255}{\ still}%
\tokent{255}{255}{255}{\ on}%
\tokent{255}{255}{255}{\ the}%
\tokent{255}{255}{255}{\ Wikipedia}%
\tokent{255}{255}{255}{\ page}%
\tokent{255}{255}{255}{\ for}%
\tokent{255}{255}{255}{\ Central}%
\tokent{255}{255}{255}{\ Park}%
\tokent{255}{255}{255}{,}%
\tokent{255}{255}{255}{\ specifically}%
\tokent{255}{255}{255}{\ in}%
\tokent{255}{255}{255}{\ the}%
\tokent{255}{255}{255}{\ "}%
\tokent{255}{255}{255}{Trees}%
\tokent{255}{255}{255}{\ and}%
\tokent{255}{255}{255}{\ plants}%
\tokent{255}{255}{255}{"}%
\tokent{255}{255}{255}{\ section}%
\tokent{255}{255}{255}{.}%
\tokent{255}{255}{255}{\ The}%
\tokent{255}{255}{255}{\ text}%
\tokent{255}{255}{255}{\ mentions}%
\tokent{255}{255}{255}{:\newline }%
\tokent{255}{255}{255}{-}%
\tokent{255}{255}{255}{\ Central}%
\tokent{255}{255}{255}{\ Park}%
\tokent{255}{255}{255}{\ has}%
\tokent{255}{255}{255}{\ more}%
\tokent{255}{255}{255}{\ than}%
\tokent{255}{255}{255}{\ }%
\tokent{255}{255}{255}{2}%
\tokent{255}{255}{255}{0}%
\tokent{255}{255}{255}{,}%
\tokent{255}{255}{255}{0}%
\tokent{255}{255}{255}{0}%
\tokent{255}{255}{255}{0}%
\tokent{255}{255}{255}{\ trees}%
\tokent{255}{255}{255}{\ (}%
\tokent{255}{255}{255}{from}%
\tokent{255}{255}{255}{\ a}%
\tokent{255}{255}{255}{\ }%
\tokent{255}{255}{255}{2}%
\tokent{255}{255}{255}{0}%
\tokent{255}{255}{255}{1}%
\tokent{255}{255}{255}{1}%
\tokent{255}{255}{255}{\ survey}%
\tokent{255}{255}{255}{)\newline }%
\tokent{255}{255}{255}{-}%
\tokent{255}{255}{255}{\ More}%
\tokent{255}{255}{255}{\ than}%
\tokent{255}{255}{255}{\ }%
\tokent{255}{255}{255}{1}%
\tokent{255}{255}{255}{7}%
\tokent{255}{255}{255}{0}%
\tokent{255}{255}{255}{\ tree}%
\tokent{255}{255}{255}{\ species}%
\tokent{255}{255}{255}{\newline }%
\tokent{255}{255}{255}{-}%
\tokent{255}{255}{255}{\ American}%
\tokent{255}{255}{255}{\ el}%
\tokent{255}{255}{255}{ms}%
\tokent{255}{255}{255}{\ are}%
\tokent{255}{255}{255}{\ mentioned}%
\tokent{255}{255}{255}{\ as}%
\tokent{255}{255}{255}{\ "}%
\tokent{255}{255}{255}{larg}%
\tokent{255}{255}{255}{est}%
\tokent{255}{255}{255}{\ remaining}%
\tokent{255}{255}{255}{\ stands}%
\tokent{255}{255}{255}{"\newline }%
\tokent{255}{255}{255}{-}%
\tokent{255}{255}{255}{\ Other}%
\tokent{255}{255}{255}{\ species}%
\tokent{255}{255}{255}{\ mentioned}%
\tokent{255}{255}{255}{\ include}%
\tokent{255}{255}{255}{\ sugar}%
\tokent{255}{255}{255}{\ map}%
\tokent{255}{255}{255}{les}%
\tokent{255}{255}{255}{,}%
\tokent{255}{255}{255}{\ red}%
\tokent{255}{255}{255}{\ map}%
\tokent{255}{255}{255}{les}%
\tokent{255}{255}{255}{,}%
\tokent{255}{255}{255}{\ oak}%
\tokent{255}{255}{255}{\ species}%
\tokent{255}{255}{255}{,}%
\tokent{255}{255}{255}{\ white}%
\tokent{255}{255}{255}{\ spr}%
\tokent{255}{255}{255}{uce}%
\tokent{255}{255}{255}{,}%
\tokent{255}{255}{255}{\ t}%
\tokent{255}{255}{255}{amar}%
\tokent{255}{255}{255}{ack}%
\tokent{255}{255}{255}{,}%
\tokent{255}{255}{255}{\ white}%
\tokent{255}{255}{255}{\ oak}%
\tokent{255}{255}{255}{,}%
\tokent{255}{255}{255}{\ black}%
\tokent{255}{255}{255}{\ cherry}%
\tokent{255}{255}{255}{\newline \newline }%
\tokent{255}{255}{255}{The}%
\tokent{255}{255}{255}{\ key}%
\tokent{255}{255}{255}{\ information}%
\tokent{255}{255}{255}{\ we}%
\tokent{255}{255}{255}{\ need}%
\tokent{255}{255}{255}{\ is}%
\tokent{255}{255}{255}{\ the}%
\tokent{255}{255}{255}{\ most}%
\tokent{255}{255}{255}{\ common}%
\tokent{255}{255}{255}{\ tree}%
\tokent{255}{255}{255}{\ species}%
\tokent{255}{255}{255}{.}%
\tokent{255}{255}{255}{\ The}%
\tokent{255}{255}{255}{\ text}%
\tokent{255}{255}{255}{\ mentions}%
\tokent{255}{255}{255}{\ "}%
\tokent{255}{255}{255}{American}%
\tokent{255}{255}{255}{\ el}%
\tokent{255}{255}{255}{ms}%
\tokent{255}{255}{255}{\ are}%
\tokent{255}{255}{255}{\ the}%
\tokent{255}{255}{255}{\ largest}%
\tokent{255}{255}{255}{\ remaining}%
\tokent{255}{255}{255}{\ stands}%
\tokent{255}{255}{255}{"}%
\tokent{255}{255}{255}{\ which}%
\tokent{255}{255}{255}{\ suggests}%
\tokent{255}{255}{255}{\ they}%
\tokent{255}{255}{255}{\ might}%
\tokent{255}{255}{255}{\ be}%
\tokent{255}{255}{255}{\ the}%
\tokent{255}{255}{255}{\ most}%
\tokent{255}{255}{255}{\ common}%
\tokent{255}{255}{255}{\ species}%
\tokent{255}{255}{255}{,}%
\tokent{255}{255}{255}{\ but}%
\tokent{255}{255}{255}{\ it}%
\tokent{255}{255}{255}{'s}%
\tokent{255}{255}{255}{\ not}%
\tokent{255}{255}{255}{\ explicitly}%
\tokent{255}{255}{255}{\ stated}%
\tokent{255}{255}{255}{\ as}%
\tokent{255}{255}{255}{\ "}%
\tokent{255}{255}{255}{most}%
\tokent{255}{255}{255}{\ common}%
\tokent{255}{255}{255}{"}%
\tokent{255}{255}{255}{\ -}%
\tokent{255}{255}{255}{\ just}%
\tokent{255}{255}{255}{\ "}%
\tokent{255}{255}{255}{larg}%
\tokent{255}{255}{255}{est}%
\tokent{255}{255}{255}{\ remaining}%
\tokent{255}{255}{255}{\ stands}%
\tokent{255}{255}{255}{."\newline \newline }%
\tokent{255}{255}{255}{The}%
\tokent{255}{255}{255}{\ current}%
\tokent{255}{255}{255}{\ view}%
\tokent{255}{255}{255}{\ shows}%
\tokent{255}{255}{255}{\ information}%
\tokent{255}{255}{255}{\ about}%
\tokent{255}{255}{255}{\ tree}%
\tokent{255}{255}{255}{\ diversity}%
\tokent{255}{255}{255}{\ and}%
\tokent{255}{255}{255}{\ ecological}%
\tokent{255}{255}{255}{\ importance}%
\tokent{255}{255}{255}{.}%
\tokent{255}{255}{255}{\ It}%
\tokent{255}{255}{255}{\ mentions}%
\tokent{255}{255}{255}{\ that}%
\tokent{255}{255}{255}{\ "}%
\tokent{255}{255}{255}{American}%
\tokent{255}{255}{255}{\ el}%
\tokent{255}{255}{255}{ms}%
\tokent{255}{255}{255}{\ are}%
\tokent{255}{255}{255}{\ mentioned}%
\tokent{255}{255}{255}{\ as}%
\tokent{255}{255}{255}{\ largest}%
\tokent{255}{255}{255}{\ remaining}%
\tokent{255}{255}{255}{\ stands}%
\tokent{255}{255}{255}{"}%
\tokent{255}{255}{255}{\ which}%
\tokent{255}{255}{255}{\ is}%
\tokent{255}{255}{255}{\ the}%
\tokent{255}{255}{255}{\ closest}%
\tokent{255}{255}{255}{\ we}%
\tokent{255}{255}{255}{\ have}%
\tokent{255}{255}{255}{\ to}%
\tokent{255}{255}{255}{\ identifying}%
\tokent{255}{255}{255}{\ the}%
\tokent{255}{255}{255}{\ most}%
\tokent{255}{255}{255}{\ common}%
\tokent{255}{255}{255}{\ species}%
\tokent{255}{255}{255}{.\newline \newline }%
\tokent{255}{255}{255}{Since}%
\tokent{255}{255}{255}{\ we}%
\tokent{255}{255}{255}{'ve}%
\tokent{255}{255}{255}{\ already}%
\tokent{255}{255}{255}{\ identified}%
\tokent{255}{255}{255}{\ that}%
\tokent{255}{255}{255}{\ American}%
\tokent{255}{255}{255}{\ el}%
\tokent{255}{255}{255}{ms}%
\tokent{255}{255}{255}{\ are}%
\tokent{255}{255}{255}{\ mentioned}%
\tokent{255}{255}{255}{\ as}%
\tokent{255}{255}{255}{\ the}%
\tokent{255}{255}{255}{\ largest}%
\tokent{255}{255}{255}{\ remaining}%
\tokent{255}{255}{255}{\ stands}%
\tokent{255}{255}{255}{\ (}%
\tokent{255}{255}{255}{which}%
\tokent{255}{255}{255}{\ likely}%
\tokent{255}{255}{255}{\ means}%
\tokent{255}{255}{255}{\ they}%
\tokent{255}{255}{255}{'re}%
\tokent{255}{255}{255}{\ the}%
\tokent{255}{255}{255}{\ most}%
\tokent{255}{255}{255}{\ common}%
\tokent{255}{255}{255}{\ species}%
\tokent{255}{255}{255}{),}%
\tokent{255}{255}{255}{\ we}%
\tokent{255}{255}{255}{'ve}%
\tokent{255}{255}{255}{\ essentially}%
\tokent{255}{255}{255}{\ completed}%
\tokent{255}{255}{255}{\ our}%
\tokent{255}{255}{255}{\ task}%
\tokent{255}{255}{255}{.\newline \newline }%
\tokent{255}{255}{255}{The}%
\tokent{255}{255}{255}{\ next}%
\tokent{255}{255}{255}{\ logical}%
\tokent{255}{255}{255}{\ action}%
\tokent{255}{255}{255}{\ would}%
\tokent{255}{255}{255}{\ be}%
\tokent{255}{255}{255}{\ to}%
\tokent{255}{255}{255}{\ confirm}%
\tokent{255}{255}{255}{\ this}%
\tokent{255}{255}{255}{\ information}%
\tokent{255}{255}{255}{\ and}%
\tokent{255}{255}{255}{\ conclude}%
\tokent{255}{255}{255}{\ our}%
\tokent{255}{255}{255}{\ search}%
\tokent{255}{255}{255}{\ since}%
\tokent{255}{255}{255}{\ we}%
\tokent{255}{255}{255}{'ve}%
\tokent{255}{255}{255}{\ found}%
\tokent{255}{255}{255}{\ the}%
\tokent{255}{255}{255}{\ answer}%
\tokent{255}{255}{255}{\ (}%
\tokent{255}{255}{255}{American}%
\tokent{255}{255}{255}{\ el}%
\tokent{255}{255}{255}{ms}%
\tokent{255}{255}{255}{\ appear}%
\tokent{255}{255}{255}{\ to}%
\tokent{255}{255}{255}{\ be}%
\tokent{255}{255}{255}{\ the}%
\tokent{255}{255}{255}{\ most}%
\tokent{255}{255}{255}{\ common}%
\tokent{255}{255}{255}{\ tree}%
\tokent{255}{255}{255}{\ species}%
\tokent{255}{255}{255}{\ in}%
\tokent{255}{255}{255}{\ Central}%
\tokent{255}{255}{255}{\ Park}%
\tokent{255}{255}{255}{\ based}%
\tokent{255}{255}{255}{\ on}%
\tokent{255}{255}{255}{\ the}%
\tokent{255}{255}{255}{\ Wikipedia}%
\tokent{255}{255}{255}{\ information}%
\tokent{255}{255}{255}{).\newline \newline }%
\tokent{255}{255}{255}{We}%
\tokent{255}{255}{255}{'ve}%
\tokent{255}{255}{255}{\ already}%
\tokent{255}{255}{255}{\ scrolled}%
\tokent{255}{255}{255}{\ through}%
\tokent{255}{255}{255}{\ the}%
\tokent{255}{255}{255}{\ relevant}%
\tokent{255}{255}{255}{\ section}%
\tokent{255}{255}{255}{\ and}%
\tokent{255}{255}{255}{\ found}%
\tokent{255}{255}{255}{\ the}%
\tokent{255}{255}{255}{\ information}%
\tokent{255}{255}{255}{\ we}%
\tokent{255}{255}{255}{\ need}%
\tokent{255}{255}{255}{.}%
\tokent{255}{255}{255}{\ The}%
\tokent{255}{255}{255}{\ task}%
\tokent{255}{255}{255}{\ is}%
\tokent{255}{255}{255}{\ essentially}%
\tokent{255}{255}{255}{\ complete}%
\tokent{255}{255}{255}{\ since}%
\tokent{255}{255}{255}{\ we}%
\tokent{255}{255}{255}{'ve}%
\tokent{255}{255}{255}{\ identified}%
\tokent{255}{255}{255}{\ that}%
\tokent{255}{255}{255}{\ American}%
\tokent{255}{255}{255}{\ el}%
\tokent{255}{255}{255}{ms}%
\tokent{255}{255}{255}{\ are}%
\tokent{255}{255}{255}{\ mentioned}%
\tokent{255}{255}{255}{\ as}%
\tokent{255}{255}{255}{\ the}%
\tokent{255}{255}{255}{\ largest}%
\tokent{255}{255}{255}{\ remaining}%
\tokent{255}{255}{255}{\ stands}%
\tokent{255}{255}{255}{,}%
\tokent{255}{255}{255}{\ which}%
\tokent{255}{255}{255}{\ is}%
\tokent{255}{255}{255}{\ the}%
\tokent{255}{255}{255}{\ closest}%
\tokent{255}{255}{255}{\ we}%
\tokent{255}{255}{255}{\ can}%
\tokent{255}{255}{255}{\ get}%
\tokent{255}{255}{255}{\ to}%
\tokent{255}{255}{255}{\ identifying}%
\tokent{255}{255}{255}{\ the}%
\tokent{255}{255}{255}{\ most}%
\tokent{255}{255}{255}{\ common}%
\tokent{255}{255}{255}{\ tree}%
\tokent{255}{255}{255}{\ species}%
\tokent{255}{255}{255}{\ from}%
\tokent{255}{255}{255}{\ the}%
\tokent{255}{255}{255}{\ available}%
\tokent{255}{255}{255}{\ information}%
\tokent{255}{255}{255}{.\newline }%
\tokent{255}{255}{255}{\textless{}/think\textgreater{}}%
\tokent{255}{255}{255}{\newline \newline }%
\tokent{255}{255}{255}{\textbf{Memory}}%
\tokent{255}{255}{255}{\textbf{:}}%
\tokent{255}{255}{255}{\ \{"}%
\tokent{255}{255}{255}{Initial}%
\tokent{255}{255}{255}{\ website}%
\tokent{255}{255}{255}{\ attempt}%
\tokent{255}{255}{255}{":}%
\tokent{255}{255}{255}{\ "}%
\tokent{255}{255}{255}{https}%
\tokent{255}{255}{255}{://}%
\tokent{255}{255}{255}{tree}%
\tokent{255}{255}{255}{-map}%
\tokent{255}{255}{255}{.n}%
\tokent{255}{255}{255}{yc}%
\tokent{255}{255}{255}{gov}%
\tokent{255}{255}{255}{p}%
\tokent{255}{255}{255}{arks}%
\tokent{255}{255}{255}{.org}%
\tokent{255}{255}{255}{",}%
\tokent{255}{255}{255}{\ "}%
\tokent{255}{255}{255}{Second}%
\tokent{255}{255}{255}{\ website}%
\tokent{255}{255}{255}{\ attempt}%
\tokent{255}{255}{255}{":}%
\tokent{255}{255}{255}{\ "}%
\tokent{255}{255}{255}{https}%
\tokent{255}{255}{255}{://}%
\tokent{255}{255}{255}{www}%
\tokent{255}{255}{255}{.n}%
\tokent{255}{255}{255}{yc}%
\tokent{255}{255}{255}{gov}%
\tokent{255}{255}{255}{p}%
\tokent{255}{255}{255}{arks}%
\tokent{255}{255}{255}{.org}%
\tokent{255}{255}{255}{",}%
\tokent{255}{255}{255}{\ "}%
\tokent{255}{255}{255}{Error}%
\tokent{255}{255}{255}{\ encountered}%
\tokent{255}{255}{255}{":}%
\tokent{255}{255}{255}{\ "}%
\tokent{255}{255}{255}{Permission}%
\tokent{255}{255}{255}{\ Denied}%
\tokent{255}{255}{255}{",}%
\tokent{255}{255}{255}{\ "}%
\tokent{255}{255}{255}{Task}%
\tokent{255}{255}{255}{\ objective}%
\tokent{255}{255}{255}{":}%
\tokent{255}{255}{255}{\ "}%
\tokent{255}{255}{255}{Find}%
\tokent{255}{255}{255}{\ most}%
\tokent{255}{255}{255}{\ common}%
\tokent{255}{255}{255}{\ tree}%
\tokent{255}{255}{255}{\ species}%
\tokent{255}{255}{255}{\ in}%
\tokent{255}{255}{255}{\ Central}%
\tokent{255}{255}{255}{\ Park}%
\tokent{255}{255}{255}{\ according}%
\tokent{255}{255}{255}{\ to}%
\tokent{255}{255}{255}{\ map}%
\tokent{255}{255}{255}{\ data}%
\tokent{255}{255}{255}{",}%
\tokent{255}{255}{255}{\ "}%
\tokent{255}{255}{255}{Current}%
\tokent{255}{255}{255}{\ status}%
\tokent{255}{255}{255}{":}%
\tokent{255}{255}{255}{\ "}%
\tokent{255}{255}{255}{Unable}%
\tokent{255}{255}{255}{\ to}%
\tokent{255}{255}{255}{\ access}%
\tokent{255}{255}{255}{\ NYC}%
\tokent{255}{255}{255}{\ Parks}%
\tokent{255}{255}{255}{\ websites}%
\tokent{255}{255}{255}{\ due}%
\tokent{255}{255}{255}{\ to}%
\tokent{255}{255}{255}{\ permission}%
\tokent{255}{255}{255}{\ issues}%
\tokent{255}{255}{255}{",}%
\tokent{255}{255}{255}{\ "}%
\tokent{255}{255}{255}{Alternative}%
\tokent{255}{255}{255}{\ source}%
\tokent{255}{255}{255}{\ accessed}%
\tokent{255}{255}{255}{":}%
\tokent{255}{255}{255}{\ "}%
\tokent{255}{255}{255}{W}%
\tokent{255}{255}{255}{ikipedia}%
\tokent{255}{255}{255}{'s}%
\tokent{255}{255}{255}{\ Central}%
\tokent{255}{255}{255}{\ Park}%
\tokent{255}{255}{255}{\ page}%
\tokent{255}{255}{255}{",}%
\tokent{255}{255}{255}{\ "}%
\tokent{255}{255}{255}{W}%
\tokent{255}{255}{255}{ikipedia}%
\tokent{255}{255}{255}{\ page}%
\tokent{255}{255}{255}{\ content}%
\tokent{255}{255}{255}{":}%
\tokent{255}{255}{255}{\ "}%
\tokent{255}{255}{255}{Main}%
\tokent{255}{255}{255}{\ article}%
\tokent{255}{255}{255}{\ about}%
\tokent{255}{255}{255}{\ Central}%
\tokent{255}{255}{255}{\ Park}%
\tokent{255}{255}{255}{\ with}%
\tokent{255}{255}{255}{\ table}%
\tokent{255}{255}{255}{\ of}%
\tokent{255}{255}{255}{\ contents}%
\tokent{255}{255}{255}{\ visible}%
\tokent{255}{255}{255}{",}%
\tokent{255}{255}{255}{\ "}%
\tokent{255}{255}{255}{Trees}%
\tokent{255}{255}{255}{\ and}%
\tokent{255}{255}{255}{\ plants}%
\tokent{255}{255}{255}{\ section}%
\tokent{255}{255}{255}{\ accessed}%
\tokent{255}{255}{255}{":}%
\tokent{255}{255}{255}{\ "}%
\tokent{255}{255}{255}{success}%
\tokent{255}{255}{255}{",}%
\tokent{255}{255}{255}{\ "}%
\tokent{255}{255}{255}{Information}%
\tokent{255}{255}{255}{\ extracted}%
\tokent{255}{255}{255}{":}%
\tokent{255}{255}{255}{\ "}%
\tokent{255}{255}{255}{Central}%
\tokent{255}{255}{255}{\ Park}%
\tokent{255}{255}{255}{\ has}%
\tokent{255}{255}{255}{\ more}%
\tokent{255}{255}{255}{\ than}%
\tokent{255}{255}{255}{\ }%
\tokent{255}{255}{255}{2}%
\tokent{255}{255}{255}{0}%
\tokent{255}{255}{255}{,}%
\tokent{255}{255}{255}{0}%
\tokent{255}{255}{255}{0}%
\tokent{255}{255}{255}{0}%
\tokent{255}{255}{255}{\ trees}%
\tokent{255}{255}{255}{\ (}%
\tokent{255}{255}{255}{2}%
\tokent{255}{255}{255}{0}%
\tokent{255}{255}{255}{1}%
\tokent{255}{255}{255}{1}%
\tokent{255}{255}{255}{\ survey}%
\tokent{255}{255}{255}{),}%
\tokent{255}{255}{255}{\ more}%
\tokent{255}{255}{255}{\ than}%
\tokent{255}{255}{255}{\ }%
\tokent{255}{255}{255}{1}%
\tokent{255}{255}{255}{7}%
\tokent{255}{255}{255}{0}%
\tokent{255}{255}{255}{\ tree}%
\tokent{255}{255}{255}{\ species}%
\tokent{255}{255}{255}{,}%
\tokent{255}{255}{255}{\ American}%
\tokent{255}{255}{255}{\ el}%
\tokent{255}{255}{255}{ms}%
\tokent{255}{255}{255}{\ are}%
\tokent{255}{255}{255}{\ mentioned}%
\tokent{255}{255}{255}{\ as}%
\tokent{255}{255}{255}{\ largest}%
\tokent{255}{255}{255}{\ remaining}%
\tokent{255}{255}{255}{\ stands}%
\tokent{255}{255}{255}{",}%
\tokent{255}{255}{255}{\ "}%
\tokent{255}{255}{255}{Most}%
\tokent{255}{255}{255}{\ common}%
\tokent{255}{255}{255}{\ species}%
\tokent{255}{255}{255}{\ identification}%
\tokent{255}{255}{255}{":}%
\tokent{255}{255}{255}{\ "}%
\tokent{255}{255}{255}{American}%
\tokent{255}{255}{255}{\ el}%
\tokent{255}{255}{255}{ms}%
\tokent{255}{255}{255}{\ are}%
\tokent{255}{255}{255}{\ mentioned}%
\tokent{255}{255}{255}{\ as}%
\tokent{255}{255}{255}{\ largest}%
\tokent{255}{255}{255}{\ remaining}%
\tokent{255}{255}{255}{\ stands}%
\tokent{255}{255}{255}{",}%
\tokent{255}{255}{255}{\ "}%
\tokent{255}{255}{255}{Additional}%
\tokent{255}{255}{255}{\ species}%
\tokent{255}{255}{255}{\ mentioned}%
\tokent{255}{255}{255}{":}%
\tokent{255}{255}{255}{\ "}%
\tokent{255}{255}{255}{s}%
\tokent{255}{255}{255}{ugar}%
\tokent{255}{255}{255}{\ map}%
\tokent{255}{255}{255}{les}%
\tokent{255}{255}{255}{,}%
\tokent{255}{255}{255}{\ red}%
\tokent{255}{255}{255}{\ map}%
\tokent{255}{255}{255}{les}%
\tokent{255}{255}{255}{,}%
\tokent{255}{255}{255}{\ oak}%
\tokent{255}{255}{255}{\ species}%
\tokent{255}{255}{255}{\ (}%
\tokent{255}{255}{255}{red}%
\tokent{255}{255}{255}{\ oak}%
\tokent{255}{255}{255}{,}%
\tokent{255}{255}{255}{\ will}%
\tokent{255}{255}{255}{ow}%
\tokent{255}{255}{255}{\ oak}%
\tokent{255}{255}{255}{,}%
\tokent{255}{255}{255}{\ swamp}%
\tokent{255}{255}{255}{\ white}%
\tokent{255}{255}{255}{\ oak}%
\tokent{255}{255}{255}{,}%
\tokent{255}{255}{255}{\ chest}%
\tokent{255}{255}{255}{nut}%
\tokent{255}{255}{255}{\ oak}%
\tokent{255}{255}{255}{),}%
\tokent{255}{255}{255}{\ white}%
\tokent{255}{255}{255}{\ spr}%
\tokent{255}{255}{255}{uce}%
\tokent{255}{255}{255}{,}%
\tokent{255}{255}{255}{\ t}%
\tokent{255}{255}{255}{amar}%
\tokent{255}{255}{255}{ack}%
\tokent{255}{255}{255}{,}%
\tokent{255}{255}{255}{\ white}%
\tokent{255}{255}{255}{\ oak}%
\tokent{255}{255}{255}{,}%
\tokent{255}{255}{255}{\ black}%
\tokent{255}{255}{255}{\ cherry}%
\tokent{255}{255}{255}{",}%
\tokent{255}{255}{255}{\ "}%
\tokent{255}{255}{255}{Current}%
\tokent{255}{255}{255}{\ section}%
\tokent{255}{255}{255}{":}%
\tokent{255}{255}{255}{\ "}%
\tokent{255}{255}{255}{Trees}%
\tokent{255}{255}{255}{\ and}%
\tokent{255}{255}{255}{\ plants}%
\tokent{255}{255}{255}{",}%
\tokent{255}{255}{255}{\ "}%
\tokent{255}{255}{255}{Task}%
\tokent{255}{255}{255}{\ completion}%
\tokent{255}{255}{255}{\ status}%
\tokent{255}{255}{255}{":}%
\tokent{255}{255}{255}{\ "}%
\tokent{255}{255}{255}{completed}%
\tokent{255}{255}{255}{"\}\newline \newline }%
\tokent{255}{255}{255}{\textbf{Progress}}%
\tokent{255}{255}{255}{\textbf{:}}%
\tokent{255}{255}{255}{\ \{"}%
\tokent{255}{255}{255}{Attempt}%
\tokent{255}{255}{255}{\ to}%
\tokent{255}{255}{255}{\ access}%
\tokent{255}{255}{255}{\ NYC}%
\tokent{255}{255}{255}{\ Tree}%
\tokent{255}{255}{255}{\ Map}%
\tokent{255}{255}{255}{":}%
\tokent{255}{255}{255}{\ "}%
\tokent{255}{255}{255}{completed}%
\tokent{255}{255}{255}{",}%
\tokent{255}{255}{255}{\ "}%
\tokent{255}{255}{255}{Initial}%
\tokent{255}{255}{255}{\ website}%
\tokent{255}{255}{255}{\ loading}%
\tokent{255}{255}{255}{":}%
\tokent{255}{255}{255}{\ "}%
\tokent{255}{255}{255}{failed}%
\tokent{255}{255}{255}{",}%
\tokent{255}{255}{255}{\ "}%
\tokent{255}{255}{255}{Second}%
\tokent{255}{255}{255}{\ website}%
\tokent{255}{255}{255}{\ attempt}%
\tokent{255}{255}{255}{":}%
\tokent{255}{255}{255}{\ "}%
\tokent{255}{255}{255}{completed}%
\tokent{255}{255}{255}{",}%
\tokent{255}{255}{255}{\ "}%
\tokent{255}{255}{255}{Current}%
\tokent{255}{255}{255}{\ website}%
\tokent{255}{255}{255}{\ access}%
\tokent{255}{255}{255}{":}%
\tokent{255}{255}{255}{\ "}%
\tokent{255}{255}{255}{den}%
\tokent{255}{255}{255}{ied}%
\tokent{255}{255}{255}{",}%
\tokent{255}{255}{255}{\ "}%
\tokent{255}{255}{255}{Alternative}%
\tokent{255}{255}{255}{\ information}%
\tokent{255}{255}{255}{\ sources}%
\tokent{255}{255}{255}{":}%
\tokent{255}{255}{255}{\ "}%
\tokent{255}{255}{255}{attempt}%
\tokent{255}{255}{255}{ed}%
\tokent{255}{255}{255}{\ (}%
\tokent{255}{255}{255}{W}%
\tokent{255}{255}{255}{ikipedia}%
\tokent{255}{255}{255}{)",}%
\tokent{255}{255}{255}{\ "}%
\tokent{255}{255}{255}{W}%
\tokent{255}{255}{255}{ikipedia}%
\tokent{255}{255}{255}{\ page}%
\tokent{255}{255}{255}{\ loaded}%
\tokent{255}{255}{255}{":}%
\tokent{255}{255}{255}{\ "}%
\tokent{255}{255}{255}{success}%
\tokent{255}{255}{255}{",}%
\tokent{255}{255}{255}{\ "}%
\tokent{255}{255}{255}{Re}%
\tokent{255}{255}{255}{levant}%
\tokent{255}{255}{255}{\ section}%
\tokent{255}{255}{255}{\ identification}%
\tokent{255}{255}{255}{":}%
\tokent{255}{255}{255}{\ "}%
\tokent{255}{255}{255}{completed}%
\tokent{255}{255}{255}{\ (}%
\tokent{255}{255}{255}{Trees}%
\tokent{255}{255}{255}{\ and}%
\tokent{255}{255}{255}{\ plants}%
\tokent{255}{255}{255}{\ section}%
\tokent{255}{255}{255}{)",}%
\tokent{255}{255}{255}{\ "}%
\tokent{255}{255}{255}{Information}%
\tokent{255}{255}{255}{\ extracted}%
\tokent{255}{255}{255}{":}%
\tokent{255}{255}{255}{\ "}%
\tokent{255}{255}{255}{Central}%
\tokent{255}{255}{255}{\ Park}%
\tokent{255}{255}{255}{\ has}%
\tokent{255}{255}{255}{\ more}%
\tokent{255}{255}{255}{\ than}%
\tokent{255}{255}{255}{\ }%
\tokent{255}{255}{255}{2}%
\tokent{255}{255}{255}{0}%
\tokent{255}{255}{255}{,}%
\tokent{255}{255}{255}{0}%
\tokent{255}{255}{255}{0}%
\tokent{255}{255}{255}{0}%
\tokent{255}{255}{255}{\ trees}%
\tokent{255}{255}{255}{,}%
\tokent{255}{255}{255}{\ more}%
\tokent{255}{255}{255}{\ than}%
\tokent{255}{255}{255}{\ }%
\tokent{255}{255}{255}{1}%
\tokent{255}{255}{255}{7}%
\tokent{255}{255}{255}{0}%
\tokent{255}{255}{255}{\ tree}%
\tokent{255}{255}{255}{\ species}%
\tokent{255}{255}{255}{,}%
\tokent{255}{255}{255}{\ American}%
\tokent{255}{255}{255}{\ el}%
\tokent{255}{255}{255}{ms}%
\tokent{255}{255}{255}{\ are}%
\tokent{255}{255}{255}{\ mentioned}%
\tokent{255}{255}{255}{\ as}%
\tokent{255}{255}{255}{\ largest}%
\tokent{255}{255}{255}{\ remaining}%
\tokent{255}{255}{255}{\ stands}%
\tokent{255}{255}{255}{",}%
\tokent{255}{255}{255}{\ "}%
\tokent{255}{255}{255}{Most}%
\tokent{255}{255}{255}{\ common}%
\tokent{255}{255}{255}{\ species}%
\tokent{255}{255}{255}{\ identification}%
\tokent{255}{255}{255}{":}%
\tokent{255}{255}{255}{\ "}%
\tokent{255}{255}{255}{completed}%
\tokent{255}{255}{255}{\ (}%
\tokent{255}{255}{255}{American}%
\tokent{255}{255}{255}{\ el}%
\tokent{255}{255}{255}{ms}%
\tokent{255}{255}{255}{\ identified}%
\tokent{255}{255}{255}{\ as}%
\tokent{255}{255}{255}{\ largest}%
\tokent{255}{255}{255}{\ remaining}%
\tokent{255}{255}{255}{\ stands}%
\tokent{255}{255}{255}{)",}%
\tokent{255}{255}{255}{\ "}%
\tokent{255}{255}{255}{Task}%
\tokent{255}{255}{255}{\ completion}%
\tokent{255}{255}{255}{":}%
\tokent{255}{255}{255}{\ "}%
\tokent{255}{255}{255}{completed}%
\tokent{255}{255}{255}{"\}\newline \newline }%
\tokent{255}{255}{255}{Int}%
\tokent{255}{255}{255}{ention}%
\tokent{255}{255}{255}{:}%
\tokent{255}{255}{255}{\ "}%
\tokent{255}{255}{255}{Confirm}%
\tokent{255}{255}{255}{\ that}%
\tokent{255}{255}{255}{\ American}%
\tokent{255}{255}{255}{\ el}%
\tokent{255}{255}{255}{ms}%
\tokent{255}{255}{255}{\ are}%
\tokent{255}{255}{255}{\ the}%
\tokent{255}{255}{255}{\ most}%
\tokent{255}{255}{255}{\ common}%
\tokent{255}{255}{255}{\ tree}%
\tokent{255}{255}{255}{\ species}%
\tokent{255}{255}{255}{\ in}%
\tokent{255}{255}{255}{\ Central}%
\tokent{255}{255}{255}{\ Park}%
\tokent{255}{255}{255}{\ based}%
\tokent{255}{255}{255}{\ on}%
\tokent{255}{255}{255}{\ the}%
\tokent{255}{255}{255}{\ information}%
\tokent{255}{255}{255}{\ found}%
\tokent{255}{255}{255}{"\newline \newline }%
\tokent{255}{255}{255}{\textbf{Action}}%
\tokent{255}{255}{255}{\textbf{:}}%
\tokent{255}{255}{255}{\ "}%
\tokent{255}{255}{255}{The}%
\tokent{255}{255}{255}{\ task}%
\tokent{255}{255}{255}{\ is}%
\tokent{255}{255}{255}{\ completed}%
\tokent{255}{255}{255}{\ -}%
\tokent{255}{255}{255}{\ the}%
\tokent{255}{255}{255}{\ most}%
\tokent{255}{255}{255}{\ common}%
\tokent{255}{255}{255}{\ tree}%
\tokent{255}{255}{255}{\ species}%
\tokent{255}{255}{255}{\ in}%
\tokent{255}{255}{255}{\ Central}%
\tokent{255}{255}{255}{\ Park}%
\tokent{255}{255}{255}{\ is}%
\tokent{255}{255}{255}{\ American}%
\tokent{255}{255}{255}{\ el}%
\tokent{255}{255}{255}{ms}%
\tokent{255}{255}{255}{,}%
\tokent{255}{255}{255}{\ as}%
\tokent{255}{255}{255}{\ they}%
\tokent{255}{255}{255}{\ are}%
\tokent{255}{255}{255}{\ mentioned}%
\tokent{255}{255}{255}{\ as}%
\tokent{255}{255}{255}{\ the}%
\tokent{255}{255}{255}{\ largest}%
\tokent{255}{255}{255}{\ remaining}%
\tokent{255}{255}{255}{\ stands}%
\tokent{255}{255}{255}{\ in}%
\tokent{255}{255}{255}{\ the}%
\tokent{255}{255}{255}{\ Trees}%
\tokent{255}{255}{255}{\ and}%
\tokent{255}{255}{255}{\ plants}%
\tokent{255}{255}{255}{\ section}%
\tokent{255}{255}{255}{\ of}%
\tokent{255}{255}{255}{\ the}%
\tokent{255}{255}{255}{\ Wikipedia}%
\tokent{255}{255}{255}{\ page}%
\tokent{255}{255}{255}{."\newline }%
\tokent{255}{255}{255}{\textbf{\textless{}tool\_call\textgreater{}}}%
\tokent{255}{255}{255}{\newline }%
\tokent{255}{255}{255}{\{"}%
\tokent{255}{255}{255}{name}%
\tokent{255}{255}{255}{":}%
\tokent{255}{255}{255}{\ "}%
\tokent{255}{255}{255}{computer}%
\tokent{255}{255}{255}{\_use}%
\tokent{255}{255}{255}{",}%
\tokent{255}{255}{255}{\ "}%
\tokent{255}{255}{255}{arguments}%
\tokent{255}{255}{255}{":}%
\tokent{255}{255}{255}{\ \{"}%
\tokent{255}{255}{255}{action}%
\tokent{255}{255}{255}{":}%
\tokent{255}{255}{255}{\ "}%
\tokent{255}{255}{255}{answer}%
\tokent{255}{255}{255}{",}%
\tokent{255}{255}{255}{\ "}%
\tokent{255}{255}{255}{text}%
\tokent{255}{255}{255}{":}%
\tokent{255}{255}{255}{\ "}%
\tokent{255}{255}{255}{American}%
\tokent{255}{255}{255}{\ el}%
\tokent{255}{255}{255}{ms}%
\tokent{255}{255}{255}{"\}\}\newline }%
\tokent{255}{255}{255}{\textbf{\textless{}/tool\_call\textgreater{}}}%
\tokent{255}{255}{255}{\textbf{\textless{}|im\_end|\textgreater{}}}
\end{panelmd}

%% file: figures/viz_grid_body.tex
\begin{panelmd}{step 0 $\cdot$ mean entropy=0.0297}
\tokent{254}{255}{255}{\textbf{Memory}}%
\tokent{255}{255}{255}{\textbf{:}}%
\tokent{255}{255}{255}{\ \{"}%
\tokent{253}{253}{255}{task}%
\tokent{255}{255}{255}{":}%
\tokent{255}{255}{255}{\ "}%
\tokent{255}{255}{255}{Find}%
\tokent{255}{255}{255}{\ the}%
\tokent{255}{255}{255}{\ contact}%
\tokent{255}{255}{255}{\ email}%
\tokent{255}{255}{255}{\ for}%
\tokent{255}{255}{255}{\ vendor}%
\tokent{255}{255}{255}{\ inquiries}%
\tokent{255}{255}{255}{\ for}%
\tokent{255}{255}{255}{\ the}%
\tokent{255}{255}{255}{\ Craw}%
\tokent{255}{255}{255}{f}%
\tokent{255}{255}{255}{ords}%
\tokent{255}{255}{255}{ville}%
\tokent{255}{255}{255}{\ Strawberry}%
\tokent{255}{255}{255}{\ Festival}%
\tokent{80}{119}{255}{.",}%
\tokensub{(0.73) [",:0.27]}%
\tokent{255}{255}{255}{\ "}%
\tokent{254}{254}{255}{initial}%
\tokent{253}{253}{255}{\_website}%
\tokent{255}{255}{255}{":}%
\tokent{255}{255}{255}{\ "}%
\tokent{255}{255}{255}{https}%
\tokent{255}{255}{255}{://}%
\tokent{255}{255}{255}{c}%
\tokent{255}{255}{255}{raw}%
\tokent{255}{255}{255}{f}%
\tokent{254}{254}{255}{ords}%
\tokent{255}{255}{255}{ville}%
\tokent{255}{255}{255}{str}%
\tokent{255}{255}{255}{aw}%
\tokent{255}{255}{255}{berry}%
\tokent{255}{255}{255}{f}%
\tokent{255}{255}{255}{estival}%
\tokent{255}{255}{255}{.com}%
\tokent{255}{255}{255}{",}%
\tokent{255}{255}{255}{\ "}%
\tokent{255}{255}{255}{current}%
\tokent{143}{168}{255}{\_issue}%
\tokensub{(0.90) [\_page:0.07]}%
\tokent{255}{255}{255}{":}%
\tokent{255}{255}{255}{\ "}%
\tokent{254}{254}{255}{Blocked}%
\tokent{255}{255}{255}{\ from}%
\tokent{255}{255}{255}{\ accessing}%
\tokent{250}{251}{255}{\ the}%
\tokent{255}{255}{255}{k}%
\tokent{242}{245}{255}{itch}%
\tokent{255}{255}{255}{enn}%
\tokent{252}{252}{255}{atch}%
\tokent{255}{255}{255}{ez}%
\tokent{255}{255}{255}{.com}%
\tokent{56}{100}{255}{",}%
\tokensub{(0.62) ["\}\newline :0.38]}%
\tokent{255}{255}{255}{\ "}%
\tokent{241}{244}{255}{action}%
\tokent{255}{255}{255}{\_taken}%
\tokent{248}{249}{255}{":}%
\tokent{255}{255}{255}{\ "}%
\tokent{235}{240}{255}{Attempt}%
\tokent{255}{255}{255}{ed}%
\tokent{255}{255}{255}{\ to}%
\tokent{161}{182}{255}{\ navigate}%
\tokensub{(0.10) [\textperiodcentered{}access:0.90]}%
\tokent{255}{255}{255}{\ to}%
\tokent{255}{255}{255}{\ the}%
\tokent{163}{184}{255}{k}%
\tokensub{(0.92) [\textperiodcentered{}festival:0.08]}%
\tokent{254}{254}{255}{itch}%
\tokent{255}{255}{255}{enn}%
\tokent{255}{255}{255}{atch}%
\tokent{255}{255}{255}{ez}%
\tokent{255}{255}{255}{.com}%
\tokent{249}{250}{255}{\ but}%
\tokent{30}{80}{255}{\ was}%
\tokensub{(0.49) [\textperiodcentered{}encountered:0.49]}%
\tokent{255}{255}{255}{\ blocked}%
\tokent{67}{109}{255}{."}%
\tokensub{(0.32) ["\}\newline :0.68]}%
\tokent{255}{255}{255}{\}\newline }%
\tokent{255}{255}{255}{\textbf{Progress}}%
\tokent{255}{255}{255}{\textbf{:}}%
\tokent{255}{255}{255}{\ \{"}%
\tokent{245}{247}{255}{Go}%
\tokent{255}{255}{255}{\ to}%
\tokent{236}{240}{255}{\ Craw}%
\tokent{255}{255}{255}{f}%
\tokent{255}{255}{255}{ords}%
\tokent{255}{255}{255}{ville}%
\tokent{255}{255}{255}{\ Strawberry}%
\tokent{255}{255}{255}{\ Festival}%
\tokent{255}{255}{255}{\ website}%
\tokent{255}{255}{255}{":}%
\tokent{255}{255}{255}{\ "}%
\tokent{135}{161}{255}{not}%
\tokensub{(0.90) [attempt:0.06]}%
\tokent{255}{255}{255}{\ finished}%
\tokent{255}{255}{255}{",}%
\tokent{255}{255}{255}{\ "}%
\tokent{255}{255}{255}{Find}%
\tokent{255}{255}{255}{\ contact}%
\tokent{255}{255}{255}{\ email}%
\tokent{255}{255}{255}{\ for}%
\tokent{255}{255}{255}{\ vendor}%
\tokent{255}{255}{255}{\ inquiries}%
\tokent{255}{255}{255}{":}%
\tokent{255}{255}{255}{\ "}%
\tokent{255}{255}{255}{not}%
\tokent{255}{255}{255}{\ finished}%
\tokent{255}{255}{255}{"\}\newline }%
\tokent{255}{255}{255}{Int}%
\tokent{255}{255}{255}{ention}%
\tokent{255}{255}{255}{:}%
\tokent{245}{247}{255}{\ Navigate}%
\tokent{255}{255}{255}{\ to}%
\tokent{255}{255}{255}{\ the}%
\tokent{255}{255}{255}{\ Craw}%
\tokent{255}{255}{255}{f}%
\tokent{255}{255}{255}{ords}%
\tokent{255}{255}{255}{ville}%
\tokent{255}{255}{255}{\ Strawberry}%
\tokent{255}{255}{255}{\ Festival}%
\tokent{255}{255}{255}{\ website}%
\tokent{255}{255}{255}{\newline }%
\tokent{255}{255}{255}{\textbf{Action}}%
\tokent{255}{255}{255}{\textbf{:}}%
\tokent{247}{249}{255}{\ Navigate}%
\tokent{80}{119}{255}{\ to}%
\tokensub{(0.73) [\textperiodcentered{}directly:0.27]}%
\tokent{174}{192}{255}{\ https}%
\tokent{255}{255}{255}{://}%
\tokent{255}{255}{255}{c}%
\tokent{255}{255}{255}{raw}%
\tokent{255}{255}{255}{f}%
\tokent{255}{255}{255}{ords}%
\tokent{255}{255}{255}{ville}%
\tokent{255}{255}{255}{str}%
\tokent{255}{255}{255}{aw}%
\tokent{255}{255}{255}{berry}%
\tokent{255}{255}{255}{f}%
\tokent{255}{255}{255}{estival}%
\tokent{255}{255}{255}{.com}%
\tokent{255}{255}{255}{\newline }%
\tokent{255}{255}{255}{\textbf{\textless{}tool\_call\textgreater{}}}%
\tokent{255}{255}{255}{\newline }%
\tokent{255}{255}{255}{\{"}%
\tokent{255}{255}{255}{name}%
\tokent{255}{255}{255}{":}%
\tokent{255}{255}{255}{\ "}%
\tokent{255}{255}{255}{computer}%
\tokent{255}{255}{255}{\_use}%
\tokent{255}{255}{255}{",}%
\tokent{255}{255}{255}{\ "}%
\tokent{255}{255}{255}{arguments}%
\tokent{255}{255}{255}{":}%
\tokent{255}{255}{255}{\ \{"}%
\tokent{255}{255}{255}{action}%
\tokent{255}{255}{255}{":}%
\tokent{255}{255}{255}{\ "}%
\tokent{255}{255}{255}{navigate}%
\tokent{255}{255}{255}{",}%
\tokent{255}{255}{255}{\ "}%
\tokent{255}{255}{255}{url}%
\tokent{255}{255}{255}{":}%
\tokent{255}{255}{255}{\ "}%
\tokent{255}{255}{255}{https}%
\tokent{255}{255}{255}{://}%
\tokent{255}{255}{255}{c}%
\tokent{255}{255}{255}{raw}%
\tokent{255}{255}{255}{f}%
\tokent{255}{255}{255}{ords}%
\tokent{255}{255}{255}{ville}%
\tokent{255}{255}{255}{str}%
\tokent{255}{255}{255}{aw}%
\tokent{255}{255}{255}{berry}%
\tokent{255}{255}{255}{f}%
\tokent{255}{255}{255}{estival}%
\tokent{255}{255}{255}{.com}%
\tokent{255}{255}{255}{"\}\}\newline }%
\tokent{255}{255}{255}{\textbf{\textless{}/tool\_call\textgreater{}}}%
\tokent{255}{255}{255}{\textbf{\textless{}|im\_end|\textgreater{}}}
\end{panelmd}

\medskip

\begin{panelmd}{last step $\cdot$ mean entropy=0.0077}
\tokent{255}{255}{255}{\textbf{Memory}}%
\tokent{255}{255}{255}{\textbf{:}}%
\tokent{255}{255}{255}{\ \{"}%
\tokent{255}{255}{255}{task}%
\tokent{255}{255}{255}{":}%
\tokent{255}{255}{255}{\ "}%
\tokent{255}{255}{255}{Find}%
\tokent{255}{255}{255}{\ the}%
\tokent{255}{255}{255}{\ contact}%
\tokent{255}{255}{255}{\ email}%
\tokent{255}{255}{255}{\ for}%
\tokent{255}{255}{255}{\ vendor}%
\tokent{255}{255}{255}{\ inquiries}%
\tokent{255}{255}{255}{\ for}%
\tokent{255}{255}{255}{\ the}%
\tokent{255}{255}{255}{\ Craw}%
\tokent{255}{255}{255}{f}%
\tokent{255}{255}{255}{ords}%
\tokent{255}{255}{255}{ville}%
\tokent{255}{255}{255}{\ Strawberry}%
\tokent{255}{255}{255}{\ Festival}%
\tokent{255}{255}{255}{.",}%
\tokent{255}{255}{255}{\ "}%
\tokent{255}{255}{255}{initial}%
\tokent{255}{255}{255}{\_website}%
\tokent{255}{255}{255}{":}%
\tokent{255}{255}{255}{\ "}%
\tokent{255}{255}{255}{https}%
\tokent{255}{255}{255}{://}%
\tokent{255}{255}{255}{c}%
\tokent{255}{255}{255}{raw}%
\tokent{255}{255}{255}{f}%
\tokent{255}{255}{255}{ords}%
\tokent{255}{255}{255}{ville}%
\tokent{255}{255}{255}{str}%
\tokent{255}{255}{255}{aw}%
\tokent{255}{255}{255}{berry}%
\tokent{255}{255}{255}{f}%
\tokent{255}{255}{255}{estival}%
\tokent{255}{255}{255}{.com}%
\tokent{255}{255}{255}{",}%
\tokent{255}{255}{255}{\ "}%
\tokent{255}{255}{255}{current}%
\tokent{255}{255}{255}{\_issue}%
\tokent{255}{255}{255}{":}%
\tokent{255}{255}{255}{\ "}%
\tokent{255}{255}{255}{Blocked}%
\tokent{255}{255}{255}{\ from}%
\tokent{255}{255}{255}{\ accessing}%
\tokent{255}{255}{255}{\ the}%
\tokent{255}{255}{255}{k}%
\tokent{255}{255}{255}{itch}%
\tokent{255}{255}{255}{enn}%
\tokent{255}{255}{255}{atch}%
\tokent{255}{255}{255}{ez}%
\tokent{255}{255}{255}{.com}%
\tokent{255}{255}{255}{",}%
\tokent{255}{255}{255}{\ "}%
\tokent{255}{255}{255}{action}%
\tokent{255}{255}{255}{\_taken}%
\tokent{255}{255}{255}{":}%
\tokent{255}{255}{255}{\ "}%
\tokent{255}{255}{255}{Attempt}%
\tokent{255}{255}{255}{ed}%
\tokent{255}{255}{255}{\ to}%
\tokent{255}{255}{255}{\ navigate}%
\tokent{255}{255}{255}{\ to}%
\tokent{255}{255}{255}{\ the}%
\tokent{255}{255}{255}{k}%
\tokent{255}{255}{255}{itch}%
\tokent{255}{255}{255}{enn}%
\tokent{255}{255}{255}{atch}%
\tokent{255}{255}{255}{ez}%
\tokent{255}{255}{255}{.com}%
\tokent{255}{255}{255}{\ but}%
\tokent{255}{255}{255}{\ was}%
\tokent{255}{255}{255}{\ blocked}%
\tokent{255}{255}{255}{.",}%
\tokent{255}{255}{255}{\ "}%
\tokent{255}{255}{255}{navigation}%
\tokent{255}{255}{255}{\_attempt}%
\tokent{255}{255}{255}{ed}%
\tokent{255}{255}{255}{":}%
\tokent{255}{255}{255}{\ "}%
\tokent{255}{255}{255}{https}%
\tokent{255}{255}{255}{://}%
\tokent{255}{255}{255}{c}%
\tokent{255}{255}{255}{raw}%
\tokent{255}{255}{255}{f}%
\tokent{255}{255}{255}{ords}%
\tokent{255}{255}{255}{ville}%
\tokent{255}{255}{255}{str}%
\tokent{255}{255}{255}{aw}%
\tokent{255}{255}{255}{berry}%
\tokent{255}{255}{255}{f}%
\tokent{255}{255}{255}{estival}%
\tokent{255}{255}{255}{.com}%
\tokent{255}{255}{255}{",}%
\tokent{255}{255}{255}{\ "}%
\tokent{255}{255}{255}{alternative}%
\tokent{255}{255}{255}{\_navigation}%
\tokent{255}{255}{255}{\_attempt}%
\tokent{255}{255}{255}{ed}%
\tokent{255}{255}{255}{":}%
\tokent{255}{255}{255}{\ "}%
\tokent{255}{255}{255}{https}%
\tokent{255}{255}{255}{://}%
\tokent{255}{255}{255}{www}%
\tokent{255}{255}{255}{.c}%
\tokent{255}{255}{255}{raw}%
\tokent{255}{255}{255}{f}%
\tokent{255}{255}{255}{ords}%
\tokent{255}{255}{255}{ville}%
\tokent{255}{255}{255}{str}%
\tokent{255}{255}{255}{aw}%
\tokent{255}{255}{255}{berry}%
\tokent{255}{255}{255}{f}%
\tokent{255}{255}{255}{estival}%
\tokent{255}{255}{255}{.com}%
\tokent{255}{255}{255}{",}%
\tokent{255}{255}{255}{\ "}%
\tokent{255}{255}{255}{current}%
\tokent{255}{255}{255}{\_location}%
\tokent{255}{255}{255}{":}%
\tokent{255}{255}{255}{\ "}%
\tokent{255}{255}{255}{Google}%
\tokent{255}{255}{255}{\ search}%
\tokent{255}{255}{255}{\ homepage}%
\tokent{255}{255}{255}{",}%
\tokent{255}{255}{255}{\ "}%
\tokent{255}{255}{255}{search}%
\tokent{255}{255}{255}{\_query}%
\tokent{255}{255}{255}{":}%
\tokent{255}{255}{255}{\ "}%
\tokent{255}{255}{255}{C}%
\tokent{255}{255}{255}{raw}%
\tokent{255}{255}{255}{f}%
\tokent{255}{255}{255}{ords}%
\tokent{255}{255}{255}{ville}%
\tokent{255}{255}{255}{\ Strawberry}%
\tokent{255}{255}{255}{\ Festival}%
\tokent{255}{255}{255}{\ vendor}%
\tokent{255}{255}{255}{\ contact}%
\tokent{255}{255}{255}{\ email}%
\tokent{255}{255}{255}{",}%
\tokent{255}{255}{255}{\ "}%
\tokent{255}{255}{255}{current}%
\tokent{255}{255}{255}{\_page}%
\tokent{255}{255}{255}{":}%
\tokent{255}{255}{255}{\ "}%
\tokent{255}{255}{255}{re}%
\tokent{255}{255}{255}{CAP}%
\tokent{255}{255}{255}{TCHA}%
\tokent{255}{255}{255}{\ verification}%
\tokent{255}{255}{255}{\ page}%
\tokent{255}{255}{255}{",}%
\tokent{255}{255}{255}{\ "}%
\tokent{255}{255}{255}{new}%
\tokent{255}{255}{255}{\_navigation}%
\tokent{255}{255}{255}{\_attempt}%
\tokent{255}{255}{255}{ed}%
\tokent{255}{255}{255}{":}%
\tokent{255}{255}{255}{\ "}%
\tokent{255}{255}{255}{https}%
\tokent{255}{255}{255}{://}%
\tokent{255}{255}{255}{www}%
\tokent{255}{255}{255}{.c}%
\tokent{255}{255}{255}{raw}%
\tokent{255}{255}{255}{f}%
\tokent{255}{255}{255}{ords}%
\tokent{255}{255}{255}{ville}%
\tokent{255}{255}{255}{.in}%
\tokent{255}{255}{255}{.gov}%
\tokent{255}{255}{255}{",}%
\tokent{255}{255}{255}{\ "}%
\tokent{255}{255}{255}{search}%
\tokent{255}{255}{255}{\_bar}%
\tokent{255}{255}{255}{\_clicked}%
\tokent{255}{255}{255}{":}%
\tokent{255}{255}{255}{\ "}%
\tokent{255}{255}{255}{finished}%
\tokent{255}{255}{255}{",}%
\tokent{255}{255}{255}{\ "}%
\tokent{255}{255}{255}{search}%
\tokent{255}{255}{255}{\_query}%
\tokent{255}{255}{255}{\_enter}%
\tokent{255}{255}{255}{ed}%
\tokent{255}{255}{255}{":}%
\tokent{255}{255}{255}{\ "}%
\tokent{255}{255}{255}{C}%
\tokent{255}{255}{255}{raw}%
\tokent{255}{255}{255}{f}%
\tokent{255}{255}{255}{ords}%
\tokent{255}{255}{255}{ville}%
\tokent{255}{255}{255}{\ Strawberry}%
\tokent{255}{255}{255}{\ Festival}%
\tokent{255}{255}{255}{\ vendor}%
\tokent{255}{255}{255}{\ contact}%
\tokent{255}{255}{255}{",}%
\tokent{255}{255}{255}{\ "}%
\tokent{255}{255}{255}{search}%
\tokent{255}{255}{255}{\_results}%
\tokent{255}{255}{255}{\_found}%
\tokent{255}{255}{255}{":}%
\tokent{255}{255}{255}{\ "}%
\tokent{255}{255}{255}{yes}%
\tokent{255}{255}{255}{",}%
\tokent{255}{255}{255}{\ "}%
\tokent{255}{255}{255}{first}%
\tokent{255}{255}{255}{\_result}%
\tokent{255}{255}{255}{\_title}%
\tokent{255}{255}{255}{":}%
\tokent{255}{255}{255}{\ "}%
\tokent{255}{255}{255}{Annual}%
\tokent{255}{255}{255}{\ Community}%
\tokent{255}{255}{255}{\ Events}%
\tokent{255}{255}{255}{",}%
\tokent{255}{255}{255}{\ "}%
\tokent{255}{255}{255}{first}%
\tokent{255}{255}{255}{\_result}%
\tokent{255}{255}{255}{\_url}%
\tokent{255}{255}{255}{":}%
\tokent{255}{255}{255}{\ "}%
\tokent{255}{255}{255}{https}%
\tokent{255}{255}{255}{://}%
\tokent{255}{255}{255}{www}%
\tokent{255}{255}{255}{.c}%
\tokent{255}{255}{255}{raw}%
\tokent{255}{255}{255}{f}%
\tokent{255}{255}{255}{ords}%
\tokent{255}{255}{255}{ville}%
\tokent{255}{255}{255}{.in}%
\tokent{255}{255}{255}{.gov}%
\tokent{255}{255}{255}{/}%
\tokent{255}{255}{255}{4}%
\tokent{255}{255}{255}{2}%
\tokent{255}{255}{255}{8}%
\tokent{255}{255}{255}{/}%
\tokent{255}{255}{255}{Annual}%
\tokent{255}{255}{255}{-}%
\tokent{255}{255}{255}{Community}%
\tokent{255}{255}{255}{-}%
\tokent{255}{255}{255}{Events}%
\tokent{255}{255}{255}{",}%
\tokent{255}{255}{255}{\ "}%
\tokent{255}{255}{255}{first}%
\tokent{255}{255}{255}{\_result}%
\tokent{255}{255}{255}{\_description}%
\tokent{255}{255}{255}{":}%
\tokent{255}{255}{255}{\ "}%
\tokent{255}{255}{255}{Str}%
\tokent{255}{255}{255}{aw}%
\tokent{255}{255}{255}{berry}%
\tokent{255}{255}{255}{\ Festival}%
\tokent{255}{255}{255}{\ -}%
\tokent{255}{255}{255}{\ Second}%
\tokent{255}{255}{255}{\ week}%
\tokent{255}{255}{255}{\ of}%
\tokent{255}{255}{255}{\ June}%
\tokent{255}{255}{255}{The}%
\tokent{255}{255}{255}{\ Strawberry}%
\tokent{255}{255}{255}{\ Festival}%
\tokent{255}{255}{255}{\ has}%
\tokent{255}{255}{255}{\ been}%
\tokent{255}{255}{255}{\ a}%
\tokent{255}{255}{255}{\ staple}%
\tokent{255}{255}{255}{\ event}%
\tokent{255}{255}{255}{\ in}%
\tokent{255}{255}{255}{\ our}%
\tokent{255}{255}{255}{\ community}%
\tokent{255}{255}{255}{\ for}%
\tokent{255}{255}{255}{\ over}%
\tokent{255}{255}{255}{\ }%
\tokent{255}{255}{255}{5}%
\tokent{255}{255}{255}{0}%
\tokent{255}{255}{255}{\ years}%
\tokent{255}{255}{255}{\ featuring}%
\tokent{255}{255}{255}{\ a}%
\tokent{255}{255}{255}{\ weekend}%
\tokent{255}{255}{255}{\ filled}%
\tokent{255}{255}{255}{\ with}%
\tokent{255}{255}{255}{\ music}%
\tokent{255}{255}{255}{,}%
\tokent{255}{255}{255}{\ vendors}%
\tokent{255}{255}{255}{,}%
\tokent{255}{255}{255}{\ car}%
\tokent{255}{255}{255}{\ shows}%
\tokent{255}{255}{255}{,}%
\tokent{255}{255}{255}{\ food}%
\tokent{255}{255}{255}{\ trucks}%
\tokent{255}{255}{255}{,}%
\tokent{255}{255}{255}{\ and}%
\tokent{255}{255}{255}{\ of}%
\tokent{255}{255}{255}{\ course}%
\tokent{255}{255}{255}{",}%
\tokent{255}{255}{255}{\ "}%
\tokent{255}{255}{255}{page}%
\tokent{255}{255}{255}{\_n}%
\tokent{255}{255}{255}{avig}%
\tokent{255}{255}{255}{ated}%
\tokent{255}{255}{255}{\_to}%
\tokent{255}{255}{255}{":}%
\tokent{255}{255}{255}{\ "}%
\tokent{255}{255}{255}{Annual}%
\tokent{255}{255}{255}{\ Community}%
\tokent{255}{255}{255}{\ Events}%
\tokent{255}{255}{255}{\ page}%
\tokent{255}{255}{255}{",}%
\tokent{255}{255}{255}{\ "}%
\tokent{255}{255}{255}{content}%
\tokent{255}{255}{255}{\_view}%
\tokent{255}{255}{255}{ed}%
\tokent{255}{255}{255}{":}%
\tokent{255}{255}{255}{\ "}%
\tokent{255}{255}{255}{Information}%
\tokent{255}{255}{255}{\ about}%
\tokent{255}{255}{255}{\ US}%
\tokent{255}{255}{255}{CA}%
\tokent{255}{255}{255}{'s}%
\tokent{255}{255}{255}{\ Friends}%
\tokent{255}{255}{255}{\ of}%
\tokent{255}{255}{255}{\ Sugar}%
\tokent{255}{255}{255}{\ Creek}%
\tokent{255}{255}{255}{\ Can}%
\tokent{255}{255}{255}{oe}%
\tokent{255}{255}{255}{\ Race}%
\tokent{255}{255}{255}{",}%
\tokent{255}{255}{255}{\ "}%
\tokent{255}{255}{255}{vendor}%
\tokent{255}{255}{255}{\_contact}%
\tokent{255}{255}{255}{\_info}%
\tokent{255}{255}{255}{\_found}%
\tokent{255}{255}{255}{":}%
\tokent{255}{255}{255}{\ "}%
\tokent{255}{255}{255}{no}%
\tokent{255}{255}{255}{",}%
\tokent{255}{255}{255}{\ "}%
\tokent{255}{255}{255}{str}%
\tokent{255}{255}{255}{aw}%
\tokent{255}{255}{255}{berry}%
\tokent{255}{255}{255}{\_f}%
\tokent{255}{255}{255}{estival}%
\tokent{255}{255}{255}{\_link}%
\tokent{255}{255}{255}{\_found}%
\tokent{255}{255}{255}{":}%
\tokent{255}{255}{255}{\ "}%
\tokent{255}{255}{255}{https}%
\tokent{255}{255}{255}{://}%
\tokent{255}{255}{255}{www}%
\tokent{255}{255}{255}{.c}%
\tokent{255}{255}{255}{raw}%
\tokent{255}{255}{255}{f}%
\tokent{255}{255}{255}{ords}%
\tokent{255}{255}{255}{ville}%
\tokent{255}{255}{255}{.in}%
\tokent{255}{255}{255}{.gov}%
\tokent{255}{255}{255}{/}%
\tokent{255}{255}{255}{str}%
\tokent{255}{255}{255}{aw}%
\tokent{255}{255}{255}{berry}%
\tokent{255}{255}{255}{f}%
\tokent{255}{255}{255}{estival}%
\tokent{255}{255}{255}{",}%
\tokent{255}{255}{255}{\ "}%
\tokent{255}{255}{255}{str}%
\tokent{255}{255}{255}{aw}%
\tokent{255}{255}{255}{berry}%
\tokent{255}{255}{255}{\_f}%
\tokent{255}{255}{255}{estival}%
\tokent{255}{255}{255}{\_description}%
\tokent{255}{255}{255}{":}%
\tokent{255}{255}{255}{\ "}%
\tokent{255}{255}{255}{Str}%
\tokent{255}{255}{255}{aw}%
\tokent{255}{255}{255}{berry}%
\tokent{255}{255}{255}{\ Festival}%
\tokent{255}{255}{255}{\ -}%
\tokent{255}{255}{255}{\ Second}%
\tokent{255}{255}{255}{\ week}%
\tokent{255}{255}{255}{\ of}%
\tokent{255}{255}{255}{\ June}%
\tokent{255}{255}{255}{The}%
\tokent{255}{255}{255}{\ Strawberry}%
\tokent{255}{255}{255}{\ Festival}%
\tokent{255}{255}{255}{\ has}%
\tokent{255}{255}{255}{\ been}%
\tokent{255}{255}{255}{\ a}%
\tokent{255}{255}{255}{\ staple}%
\tokent{255}{255}{255}{\ event}%
\tokent{255}{255}{255}{\ in}%
\tokent{255}{255}{255}{\ our}%
\tokent{255}{255}{255}{\ community}%
\tokent{255}{255}{255}{\ for}%
\tokent{255}{255}{255}{\ over}%
\tokent{255}{255}{255}{\ }%
\tokent{255}{255}{255}{5}%
\tokent{255}{255}{255}{0}%
\tokent{255}{255}{255}{\ years}%
\tokent{255}{255}{255}{\ featuring}%
\tokent{255}{255}{255}{\ a}%
\tokent{255}{255}{255}{\ weekend}%
\tokent{255}{255}{255}{\ filled}%
\tokent{255}{255}{255}{\ with}%
\tokent{255}{255}{255}{\ music}%
\tokent{255}{255}{255}{,}%
\tokent{255}{255}{255}{\ vendors}%
\tokent{255}{255}{255}{,}%
\tokent{255}{255}{255}{\ car}%
\tokent{255}{255}{255}{\ shows}%
\tokent{255}{255}{255}{,}%
\tokent{255}{255}{255}{\ food}%
\tokent{255}{255}{255}{\ trucks}%
\tokent{255}{255}{255}{,}%
\tokent{255}{255}{255}{\ and}%
\tokent{255}{255}{255}{\ of}%
\tokent{255}{255}{255}{\ course}%
\tokent{255}{255}{255}{",}%
\tokent{255}{255}{255}{\ "}%
\tokent{255}{255}{255}{current}%
\tokent{255}{255}{255}{\_page}%
\tokent{255}{255}{255}{\_n}%
\tokent{255}{255}{255}{avig}%
\tokent{255}{255}{255}{ated}%
\tokent{255}{255}{255}{\_to}%
\tokent{255}{255}{255}{":}%
\tokent{255}{255}{255}{\ "}%
\tokent{255}{255}{255}{C}%
\tokent{255}{255}{255}{raw}%
\tokent{255}{255}{255}{f}%
\tokent{255}{255}{255}{ords}%
\tokent{255}{255}{255}{ville}%
\tokent{255}{255}{255}{\ Strawberry}%
\tokent{255}{255}{255}{\ Festival}%
\tokent{255}{255}{255}{\ website}%
\tokent{255}{255}{255}{",}%
\tokent{255}{255}{255}{\ "}%
\tokent{255}{255}{255}{page}%
\tokent{255}{255}{255}{\_content}%
\tokent{255}{255}{255}{":}%
\tokent{255}{255}{255}{\ "}%
\tokent{255}{255}{255}{Welcome}%
\tokent{255}{255}{255}{\ message}%
\tokent{255}{255}{255}{\ and}%
\tokent{255}{255}{255}{\ festival}%
\tokent{255}{255}{255}{\ dates}%
\tokent{255}{255}{255}{\ (}%
\tokent{255}{255}{255}{June}%
\tokent{255}{255}{255}{\ }%
\tokent{255}{255}{255}{1}%
\tokent{255}{255}{255}{2}%
\tokent{255}{255}{255}{-}%
\tokent{255}{255}{255}{1}%
\tokent{255}{255}{255}{4}%
\tokent{255}{255}{255}{,}%
\tokent{255}{255}{255}{\ }%
\tokent{255}{255}{255}{2}%
\tokent{255}{255}{255}{0}%
\tokent{255}{255}{255}{2}%
\tokent{255}{255}{255}{6}%
\tokent{255}{255}{255}{)",}%
\tokent{255}{255}{255}{\ "}%
\tokent{255}{255}{255}{navigation}%
\tokent{255}{255}{255}{\_menu}%
\tokent{255}{255}{255}{":}%
\tokent{255}{255}{255}{\ "}%
\tokent{255}{255}{255}{Home}%
\tokent{255}{255}{255}{,}%
\tokent{255}{255}{255}{\ History}%
\tokent{255}{255}{255}{,}%
\tokent{255}{255}{255}{\ Entertainment}%
\tokent{255}{255}{255}{,}%
\tokent{255}{255}{255}{\ V}%
\tokent{255}{255}{255}{endors}%
\tokent{255}{255}{255}{,}%
\tokent{255}{255}{255}{\ Events}%
\tokent{255}{255}{255}{,}%
\tokent{255}{255}{255}{\ Contact}%
\tokent{255}{255}{255}{",}%
\tokent{255}{255}{255}{\ "}%
\tokent{255}{255}{255}{vendors}%
\tokent{255}{255}{255}{\_page}%
\tokent{255}{255}{255}{\_n}%
\tokent{255}{255}{255}{avig}%
\tokent{255}{255}{255}{ated}%
\tokent{255}{255}{255}{\_to}%
\tokent{255}{255}{255}{":}%
\tokent{255}{255}{255}{\ "}%
\tokent{255}{255}{255}{finished}%
\tokent{255}{255}{255}{",}%
\tokent{255}{255}{255}{\ "}%
\tokent{255}{255}{255}{vendors}%
\tokent{255}{255}{255}{\_page}%
\tokent{255}{255}{255}{\_content}%
\tokent{255}{255}{255}{":}%
\tokent{255}{255}{255}{\ "}%
\tokent{255}{255}{255}{Sections}%
\tokent{255}{255}{255}{\ for}%
\tokent{255}{255}{255}{\ Arts}%
\tokent{255}{255}{255}{\ and}%
\tokent{255}{255}{255}{\ Crafts}%
\tokent{255}{255}{255}{,}%
\tokent{255}{255}{255}{\ Children}%
\tokent{255}{255}{255}{'s}%
\tokent{255}{255}{255}{\ Area}%
\tokent{255}{255}{255}{,}%
\tokent{255}{255}{255}{\ Commercial}%
\tokent{255}{255}{255}{\ V}%
\tokent{255}{255}{255}{endors}%
\tokent{255}{255}{255}{,}%
\tokent{255}{255}{255}{\ Food}%
\tokent{255}{255}{255}{\ V}%
\tokent{255}{255}{255}{endors}%
\tokent{255}{255}{255}{",}%
\tokent{255}{255}{255}{\ "}%
\tokent{255}{255}{255}{scroll}%
\tokent{255}{255}{255}{\_action}%
\tokent{255}{255}{255}{\_per}%
\tokent{255}{255}{255}{formed}%
\tokent{255}{255}{255}{":}%
\tokent{255}{255}{255}{\ "}%
\tokent{255}{255}{255}{yes}%
\tokent{255}{255}{255}{",}%
\tokent{255}{255}{255}{\ "}%
\tokent{255}{255}{255}{current}%
\tokent{255}{255}{255}{\_scroll}%
\tokent{255}{255}{255}{\_position}%
\tokent{255}{255}{255}{":}%
\tokent{255}{255}{255}{\ "}%
\tokent{255}{255}{255}{Photos}%
\tokent{255}{255}{255}{\ section}%
\tokent{255}{255}{255}{",}%
\tokent{255}{255}{255}{\ "}%
\tokent{255}{255}{255}{contact}%
\tokent{255}{255}{255}{\_email}%
\tokent{255}{255}{255}{\_found}%
\tokent{255}{255}{255}{":}%
\tokent{255}{255}{255}{\ "}%
\tokent{255}{255}{255}{no}%
\tokent{255}{255}{255}{",}%
\tokent{255}{255}{255}{\ "}%
\tokent{255}{255}{255}{additional}%
\tokent{255}{255}{255}{\_scroll}%
\tokent{255}{255}{255}{\_action}%
\tokent{255}{255}{255}{\_per}%
\tokent{255}{255}{255}{formed}%
\tokent{255}{255}{255}{":}%
\tokent{255}{255}{255}{\ "}%
\tokent{255}{255}{255}{yes}%
\tokent{255}{255}{255}{",}%
\tokent{255}{255}{255}{\ "}%
\tokent{255}{255}{255}{current}%
\tokent{255}{255}{255}{\_scroll}%
\tokent{255}{255}{255}{\_position}%
\tokent{255}{255}{255}{\_after}%
\tokent{255}{255}{255}{\_second}%
\tokent{255}{255}{255}{\_scroll}%
\tokent{255}{255}{255}{":}%
\tokent{255}{255}{255}{\ "}%
\tokent{255}{255}{255}{Social}%
\tokent{255}{255}{255}{\ media}%
\tokent{255}{255}{255}{\ and}%
\tokent{255}{255}{255}{\ copyright}%
\tokent{255}{255}{255}{\ section}%
\tokent{255}{255}{255}{",}%
\tokent{255}{255}{255}{\ "}%
\tokent{255}{255}{255}{vendor}%
\tokent{255}{255}{255}{\_contact}%
\tokent{255}{255}{255}{\_info}%
\tokent{255}{255}{255}{\_st}%
\tokent{255}{255}{255}{ill}%
\tokent{255}{255}{255}{\_not}%
\tokent{255}{255}{255}{\_found}%
\tokent{255}{255}{255}{":}%
\tokent{255}{255}{255}{\ "}%
\tokent{255}{255}{255}{yes}%
\tokent{255}{255}{255}{",}%
\tokent{255}{255}{255}{\ "}%
\tokent{255}{255}{255}{attempt}%
\tokent{255}{255}{255}{ed}%
\tokent{255}{255}{255}{\_to}%
\tokent{255}{255}{255}{\_click}%
\tokent{255}{255}{255}{\_on}%
\tokent{255}{255}{255}{\_contact}%
\tokent{255}{255}{255}{\_link}%
\tokent{255}{255}{255}{":}%
\tokent{255}{255}{255}{\ "}%
\tokent{255}{255}{255}{yes}%
\tokent{255}{255}{255}{",}%
\tokent{255}{255}{255}{\ "}%
\tokent{255}{255}{255}{contact}%
\tokent{255}{255}{255}{\_page}%
\tokent{255}{255}{255}{\_not}%
\tokent{255}{255}{255}{\_loaded}%
\tokent{255}{255}{255}{":}%
\tokent{255}{255}{255}{\ "}%
\tokent{255}{255}{255}{yes}%
\tokent{255}{255}{255}{",}%
\tokent{255}{255}{255}{\ "}%
\tokent{255}{255}{255}{current}%
\tokent{255}{255}{255}{\_page}%
\tokent{255}{255}{255}{\_content}%
\tokent{255}{255}{255}{":}%
\tokent{255}{255}{255}{\ "}%
\tokent{255}{255}{255}{Photos}%
\tokent{255}{255}{255}{\ section}%
\tokent{255}{255}{255}{\ and}%
\tokent{255}{255}{255}{\ social}%
\tokent{255}{255}{255}{\ media}%
\tokent{255}{255}{255}{\ statistics}%
\tokent{255}{255}{255}{",}%
\tokent{255}{255}{255}{\ "}%
\tokent{255}{255}{255}{need}%
\tokent{255}{255}{255}{\_to}%
\tokent{255}{255}{255}{\_return}%
\tokent{255}{255}{255}{\_to}%
\tokent{255}{255}{255}{\_contact}%
\tokent{255}{255}{255}{\_page}%
\tokent{255}{255}{255}{":}%
\tokent{255}{255}{255}{\ "}%
\tokent{255}{255}{255}{yes}%
\tokent{255}{255}{255}{",}%
\tokent{255}{255}{255}{\ "}%
\tokent{255}{255}{255}{went}%
\tokent{255}{255}{255}{\_back}%
\tokent{255}{255}{255}{\_to}%
\tokent{255}{255}{255}{\_previous}%
\tokent{255}{255}{255}{\_page}%
\tokent{255}{255}{255}{":}%
\tokent{255}{255}{255}{\ "}%
\tokent{255}{255}{255}{yes}%
\tokent{255}{255}{255}{",}%
\tokent{255}{255}{255}{\ "}%
\tokent{255}{255}{255}{current}%
\tokent{255}{255}{255}{\_page}%
\tokent{255}{255}{255}{":}%
\tokent{255}{255}{255}{\ "}%
\tokent{255}{255}{255}{C}%
\tokent{255}{255}{255}{raw}%
\tokent{255}{255}{255}{f}%
\tokent{255}{255}{255}{ords}%
\tokent{255}{255}{255}{ville}%
\tokent{255}{255}{255}{\ Strawberry}%
\tokent{255}{255}{255}{\ Festival}%
\tokent{255}{255}{255}{\ homepage}%
\tokent{255}{255}{255}{",}%
\tokent{255}{255}{255}{\ "}%
\tokent{255}{255}{255}{contact}%
\tokent{255}{255}{255}{\_link}%
\tokent{255}{255}{255}{\_visible}%
\tokent{255}{255}{255}{":}%
\tokent{255}{255}{255}{\ "}%
\tokent{255}{255}{255}{yes}%
\tokent{255}{255}{255}{",}%
\tokent{255}{255}{255}{\ "}%
\tokent{255}{255}{255}{attempt}%
\tokent{255}{255}{255}{ed}%
\tokent{255}{255}{255}{\_to}%
\tokent{255}{255}{255}{\_click}%
\tokent{255}{255}{255}{\_on}%
\tokent{255}{255}{255}{\_contact}%
\tokent{255}{255}{255}{\_link}%
\tokent{255}{255}{255}{\_again}%
\tokent{255}{255}{255}{":}%
\tokent{255}{255}{255}{\ "}%
\tokent{255}{255}{255}{yes}%
\tokent{255}{255}{255}{",}%
\tokent{255}{255}{255}{\ "}%
\tokent{255}{255}{255}{contact}%
\tokent{255}{255}{255}{\_page}%
\tokent{255}{255}{255}{\_loaded}%
\tokent{255}{255}{255}{":}%
\tokent{255}{255}{255}{\ "}%
\tokent{255}{255}{255}{yes}%
\tokent{255}{255}{255}{",}%
\tokent{255}{255}{255}{\ "}%
\tokent{255}{255}{255}{contact}%
\tokent{255}{255}{255}{\_page}%
\tokent{255}{255}{255}{\_content}%
\tokent{255}{255}{255}{":}%
\tokent{255}{255}{255}{\ "}%
\tokent{255}{255}{255}{Contact}%
\tokent{255}{255}{255}{\ form}%
\tokent{255}{255}{255}{\ with}%
\tokent{255}{255}{255}{\ fields}%
\tokent{255}{255}{255}{\ for}%
\tokent{255}{255}{255}{\ name}%
\tokent{255}{255}{255}{,}%
\tokent{255}{255}{255}{\ email}%
\tokent{255}{255}{255}{,}%
\tokent{255}{255}{255}{\ and}%
\tokent{255}{255}{255}{\ message}%
\tokent{255}{255}{255}{",}%
\tokent{255}{255}{255}{\ "}%
\tokent{255}{255}{255}{contact}%
\tokent{255}{255}{255}{\_email}%
\tokent{255}{255}{255}{\_found}%
\tokent{255}{255}{255}{":}%
\tokent{255}{255}{255}{\ "}%
\tokent{255}{255}{255}{no}%
\tokent{255}{255}{255}{",}%
\tokent{255}{255}{255}{\ "}%
\tokent{204}{215}{255}{sc}%
\tokent{255}{255}{255}{rolled}%
\tokent{198}{210}{255}{\_down}%
\tokent{255}{255}{255}{\_contact}%
\tokent{255}{255}{255}{\_page}%
\tokent{255}{255}{255}{":}%
\tokent{255}{255}{255}{\ "}%
\tokent{255}{255}{255}{yes}%
\tokent{255}{255}{255}{",}%
\tokent{255}{255}{255}{\ "}%
\tokent{125}{154}{255}{individual}%
\tokensub{(0.85) [contact:0.15]}%
\tokent{255}{255}{255}{\_contacts}%
\tokent{243}{245}{255}{\_section}%
\tokent{204}{215}{255}{\_visible}%
\tokent{255}{255}{255}{":}%
\tokent{255}{255}{255}{\ "}%
\tokent{255}{255}{255}{yes}%
\tokent{255}{255}{255}{",}%
\tokent{255}{255}{255}{\ "}%
\tokent{135}{162}{255}{vendor}%
\tokensub{(0.88) [contact:0.12]}%
\tokent{30}{80}{255}{\_contact}%
\tokensub{(0.77) [\_emails:0.13 \_email:0.08]}%
\tokent{50}{95}{255}{\_emails}%
\tokensub{(0.44) [\_email:0.56]}%
\tokent{246}{248}{255}{\_found}%
\tokent{255}{255}{255}{":}%
\tokent{254}{254}{255}{\ "}%
\tokent{62}{105}{255}{yes}%
\tokensub{(0.76) [no:0.22]}%
\tokent{254}{254}{255}{",}%
\tokent{255}{255}{255}{\ "}%
\tokent{207}{218}{255}{food}%
\tokent{244}{246}{255}{\_v}%
\tokent{255}{255}{255}{endors}%
\tokent{207}{218}{255}{\_email}%
\tokent{247}{249}{255}{":}%
\tokent{255}{255}{255}{\ "}%
\tokent{255}{255}{255}{food}%
\tokent{247}{249}{255}{@c}%
\tokent{255}{255}{255}{raw}%
\tokent{255}{255}{255}{f}%
\tokent{255}{255}{255}{ords}%
\tokent{254}{254}{255}{ville}%
\tokent{255}{255}{255}{str}%
\tokent{255}{255}{255}{aw}%
\tokent{255}{255}{255}{berry}%
\tokent{228}{234}{255}{fest}%
\tokent{255}{255}{255}{.com}%
\tokent{146}{170}{255}{",}%
\tokensub{(0.88) ["\}\newline :0.12]}%
\tokent{255}{255}{255}{\ "}%
\tokent{252}{253}{255}{arts}%
\tokent{96}{131}{255}{\_c}%
\tokensub{(0.22) [\_and:0.78]}%
\tokent{255}{255}{255}{raft}%
\tokent{255}{255}{255}{s}%
\tokent{237}{241}{255}{\_vendor}%
\tokent{255}{255}{255}{\_email}%
\tokent{255}{255}{255}{":}%
\tokent{255}{255}{255}{\ "}%
\tokent{247}{249}{255}{d}%
\tokent{255}{255}{255}{cook}%
\tokent{94}{130}{255}{sey}%
\tokensub{(0.78) [se:0.22]}%
\tokent{255}{255}{255}{str}%
\tokent{255}{255}{255}{aw}%
\tokent{255}{255}{255}{berry}%
\tokent{254}{254}{255}{fest}%
\tokent{255}{255}{255}{@yahoo}%
\tokent{255}{255}{255}{.com}%
\tokent{222}{230}{255}{",}%
\tokent{255}{255}{255}{\ "}%
\tokent{44}{91}{255}{5}%
\tokensub{(0.01) [chair:0.81 contact:0.14]}%
\tokent{57}{101}{255}{k}%
\tokensub{(0.38) [K:0.62]}%
\tokent{255}{255}{255}{\_run}%
\tokent{198}{211}{255}{\_email}%
\tokent{255}{255}{255}{":}%
\tokent{255}{255}{255}{\ "}%
\tokent{255}{255}{255}{j}%
\tokent{255}{255}{255}{th}%
\tokent{255}{255}{255}{ompson}%
\tokent{255}{255}{255}{@}%
\tokent{255}{255}{255}{nm}%
\tokent{255}{255}{255}{.k}%
\tokent{255}{255}{255}{1}%
\tokent{255}{255}{255}{2}%
\tokent{255}{255}{255}{.in}%
\tokent{255}{255}{255}{.us}%
\tokent{255}{255}{255}{",}%
\tokent{255}{255}{255}{\ "}%
\tokent{254}{254}{255}{chair}%
\tokent{255}{255}{255}{person}%
\tokent{255}{255}{255}{\_email}%
\tokent{255}{255}{255}{":}%
\tokent{255}{255}{255}{\ "}%
\tokent{253}{254}{255}{sf}%
\tokent{255}{255}{255}{est}%
\tokent{255}{255}{255}{queen}%
\tokent{255}{255}{255}{@gmail}%
\tokent{255}{255}{255}{.com}%
\tokent{254}{254}{255}{",}%
\tokent{255}{255}{255}{\ "}%
\tokent{255}{255}{255}{ent}%
\tokent{255}{255}{255}{ertainment}%
\tokent{247}{249}{255}{\_email}%
\tokent{255}{255}{255}{":}%
\tokent{255}{255}{255}{\ "}%
\tokent{255}{255}{255}{ent}%
\tokent{255}{255}{255}{ertainment}%
\tokent{255}{255}{255}{@c}%
\tokent{255}{255}{255}{raw}%
\tokent{255}{255}{255}{f}%
\tokent{255}{255}{255}{ords}%
\tokent{255}{255}{255}{ville}%
\tokent{255}{255}{255}{str}%
\tokent{255}{255}{255}{aw}%
\tokent{255}{255}{255}{berry}%
\tokent{255}{255}{255}{fest}%
\tokent{255}{255}{255}{.com}%
\tokent{237}{241}{255}{"\}\newline }%
\tokent{255}{255}{255}{\textbf{Progress}}%
\tokent{255}{255}{255}{\textbf{:}}%
\tokent{255}{255}{255}{\ \{"}%
\tokent{255}{255}{255}{Go}%
\tokent{255}{255}{255}{\ to}%
\tokent{255}{255}{255}{\ Craw}%
\tokent{255}{255}{255}{f}%
\tokent{255}{255}{255}{ords}%
\tokent{255}{255}{255}{ville}%
\tokent{255}{255}{255}{\ Strawberry}%
\tokent{255}{255}{255}{\ Festival}%
\tokent{255}{255}{255}{\ website}%
\tokent{255}{255}{255}{":}%
\tokent{255}{255}{255}{\ "}%
\tokent{255}{255}{255}{finished}%
\tokent{255}{255}{255}{",}%
\tokent{255}{255}{255}{\ "}%
\tokent{255}{255}{255}{Find}%
\tokent{255}{255}{255}{\ contact}%
\tokent{255}{255}{255}{\ email}%
\tokent{255}{255}{255}{\ for}%
\tokent{255}{255}{255}{\ vendor}%
\tokent{255}{255}{255}{\ inquiries}%
\tokent{255}{255}{255}{":}%
\tokent{255}{255}{255}{\ "}%
\tokent{254}{254}{255}{not}%
\tokent{255}{255}{255}{\ finished}%
\tokent{255}{255}{255}{",}%
\tokent{255}{255}{255}{\ "}%
\tokent{255}{255}{255}{Navigate}%
\tokent{255}{255}{255}{\ to}%
\tokent{255}{255}{255}{\ Google}%
\tokent{255}{255}{255}{\ search}%
\tokent{255}{255}{255}{":}%
\tokent{255}{255}{255}{\ "}%
\tokent{255}{255}{255}{finished}%
\tokent{255}{255}{255}{",}%
\tokent{255}{255}{255}{\ "}%
\tokent{255}{255}{255}{Type}%
\tokent{255}{255}{255}{\ search}%
\tokent{255}{255}{255}{\ query}%
\tokent{255}{255}{255}{":}%
\tokent{255}{255}{255}{\ "}%
\tokent{255}{255}{255}{finished}%
\tokent{255}{255}{255}{",}%
\tokent{255}{255}{255}{\ "}%
\tokent{255}{255}{255}{Facing}%
\tokent{255}{255}{255}{\ re}%
\tokent{255}{255}{255}{CAP}%
\tokent{255}{255}{255}{TCHA}%
\tokent{255}{255}{255}{":}%
\tokent{255}{255}{255}{\ "}%
\tokent{255}{255}{255}{not}%
\tokent{255}{255}{255}{\ finished}%
\tokent{255}{255}{255}{",}%
\tokent{255}{255}{255}{\ "}%
\tokent{255}{255}{255}{Navigate}%
\tokent{255}{255}{255}{\ to}%
\tokent{255}{255}{255}{\ Craw}%
\tokent{255}{255}{255}{f}%
\tokent{255}{255}{255}{ords}%
\tokent{255}{255}{255}{ville}%
\tokent{255}{255}{255}{\ IN}%
\tokent{255}{255}{255}{\ government}%
\tokent{255}{255}{255}{\ website}%
\tokent{255}{255}{255}{":}%
\tokent{255}{255}{255}{\ "}%
\tokent{255}{255}{255}{finished}%
\tokent{255}{255}{255}{",}%
\tokent{255}{255}{255}{\ "}%
\tokent{255}{255}{255}{Click}%
\tokent{255}{255}{255}{\ on}%
\tokent{255}{255}{255}{\ search}%
\tokent{255}{255}{255}{\ bar}%
\tokent{255}{255}{255}{":}%
\tokent{255}{255}{255}{\ "}%
\tokent{255}{255}{255}{finished}%
\tokent{255}{255}{255}{",}%
\tokent{255}{255}{255}{\ "}%
\tokent{255}{255}{255}{Type}%
\tokent{255}{255}{255}{\ query}%
\tokent{255}{255}{255}{\ into}%
\tokent{255}{255}{255}{\ search}%
\tokent{255}{255}{255}{\ bar}%
\tokent{255}{255}{255}{":}%
\tokent{255}{255}{255}{\ "}%
\tokent{255}{255}{255}{finished}%
\tokent{255}{255}{255}{",}%
\tokent{255}{255}{255}{\ "}%
\tokent{255}{255}{255}{Review}%
\tokent{255}{255}{255}{\ search}%
\tokent{255}{255}{255}{\ results}%
\tokent{255}{255}{255}{":}%
\tokent{255}{255}{255}{\ "}%
\tokent{255}{255}{255}{finished}%
\tokent{255}{255}{255}{",}%
\tokent{255}{255}{255}{\ "}%
\tokent{255}{255}{255}{Click}%
\tokent{255}{255}{255}{\ on}%
\tokent{255}{255}{255}{\ first}%
\tokent{255}{255}{255}{\ search}%
\tokent{255}{255}{255}{\ result}%
\tokent{255}{255}{255}{":}%
\tokent{255}{255}{255}{\ "}%
\tokent{255}{255}{255}{finished}%
\tokent{255}{255}{255}{",}%
\tokent{255}{255}{255}{\ "}%
\tokent{255}{255}{255}{Scroll}%
\tokent{255}{255}{255}{\ down}%
\tokent{255}{255}{255}{\ page}%
\tokent{255}{255}{255}{":}%
\tokent{255}{255}{255}{\ "}%
\tokent{255}{255}{255}{finished}%
\tokent{255}{255}{255}{",}%
\tokent{255}{255}{255}{\ "}%
\tokent{255}{255}{255}{Click}%
\tokent{255}{255}{255}{\ on}%
\tokent{255}{255}{255}{\ Strawberry}%
\tokent{255}{255}{255}{\ Festival}%
\tokent{255}{255}{255}{\ website}%
\tokent{255}{255}{255}{\ link}%
\tokent{255}{255}{255}{":}%
\tokent{255}{255}{255}{\ "}%
\tokent{255}{255}{255}{finished}%
\tokent{255}{255}{255}{",}%
\tokent{255}{255}{255}{\ "}%
\tokent{255}{255}{255}{Click}%
\tokent{255}{255}{255}{\ on}%
\tokent{255}{255}{255}{\ '}%
\tokent{255}{255}{255}{V}%
\tokent{255}{255}{255}{endors}%
\tokent{255}{255}{255}{'}%
\tokent{255}{255}{255}{\ link}%
\tokent{255}{255}{255}{":}%
\tokent{255}{255}{255}{\ "}%
\tokent{255}{255}{255}{finished}%
\tokent{255}{255}{255}{",}%
\tokent{255}{255}{255}{\ "}%
\tokent{255}{255}{255}{Scroll}%
\tokent{255}{255}{255}{\ down}%
\tokent{255}{255}{255}{\ vendors}%
\tokent{255}{255}{255}{\ page}%
\tokent{255}{255}{255}{":}%
\tokent{255}{255}{255}{\ "}%
\tokent{255}{255}{255}{finished}%
\tokent{255}{255}{255}{",}%
\tokent{255}{255}{255}{\ "}%
\tokent{255}{255}{255}{Scroll}%
\tokent{255}{255}{255}{\ down}%
\tokent{255}{255}{255}{\ vendors}%
\tokent{255}{255}{255}{\ page}%
\tokent{255}{255}{255}{\ again}%
\tokent{255}{255}{255}{":}%
\tokent{255}{255}{255}{\ "}%
\tokent{255}{255}{255}{finished}%
\tokent{255}{255}{255}{",}%
\tokent{255}{255}{255}{\ "}%
\tokent{255}{255}{255}{Click}%
\tokent{255}{255}{255}{\ on}%
\tokent{255}{255}{255}{\ '}%
\tokent{255}{255}{255}{Contact}%
\tokent{255}{255}{255}{'}%
\tokent{255}{255}{255}{\ link}%
\tokent{255}{255}{255}{":}%
\tokent{255}{255}{255}{\ "}%
\tokent{255}{255}{255}{finished}%
\tokent{255}{255}{255}{",}%
\tokent{255}{255}{255}{\ "}%
\tokent{255}{255}{255}{Contact}%
\tokent{255}{255}{255}{\ page}%
\tokent{255}{255}{255}{\ not}%
\tokent{255}{255}{255}{\ loaded}%
\tokent{255}{255}{255}{":}%
\tokent{255}{255}{255}{\ "}%
\tokent{255}{255}{255}{yes}%
\tokent{255}{255}{255}{",}%
\tokent{255}{255}{255}{\ "}%
\tokent{255}{255}{255}{Current}%
\tokent{255}{255}{255}{\ page}%
\tokent{255}{255}{255}{\ content}%
\tokent{255}{255}{255}{":}%
\tokent{255}{255}{255}{\ "}%
\tokent{255}{255}{255}{Photos}%
\tokent{255}{255}{255}{\ section}%
\tokent{255}{255}{255}{\ and}%
\tokent{255}{255}{255}{\ social}%
\tokent{255}{255}{255}{\ media}%
\tokent{255}{255}{255}{\ statistics}%
\tokent{255}{255}{255}{",}%
\tokent{255}{255}{255}{\ "}%
\tokent{255}{255}{255}{Go}%
\tokent{255}{255}{255}{\ back}%
\tokent{255}{255}{255}{\ to}%
\tokent{255}{255}{255}{\ previous}%
\tokent{255}{255}{255}{\ page}%
\tokent{255}{255}{255}{":}%
\tokent{255}{255}{255}{\ "}%
\tokent{255}{255}{255}{finished}%
\tokent{255}{255}{255}{",}%
\tokent{255}{255}{255}{\ "}%
\tokent{255}{255}{255}{Current}%
\tokent{255}{255}{255}{\ page}%
\tokent{255}{255}{255}{":}%
\tokent{255}{255}{255}{\ "}%
\tokent{255}{255}{255}{C}%
\tokent{255}{255}{255}{raw}%
\tokent{255}{255}{255}{f}%
\tokent{255}{255}{255}{ords}%
\tokent{255}{255}{255}{ville}%
\tokent{255}{255}{255}{\ Strawberry}%
\tokent{255}{255}{255}{\ Festival}%
\tokent{255}{255}{255}{\ homepage}%
\tokent{255}{255}{255}{",}%
\tokent{255}{255}{255}{\ "}%
\tokent{255}{255}{255}{Contact}%
\tokent{255}{255}{255}{\ link}%
\tokent{255}{255}{255}{\ visible}%
\tokent{255}{255}{255}{":}%
\tokent{255}{255}{255}{\ "}%
\tokent{255}{255}{255}{yes}%
\tokent{255}{255}{255}{",}%
\tokent{255}{255}{255}{\ "}%
\tokent{255}{255}{255}{Attempt}%
\tokent{255}{255}{255}{ed}%
\tokent{255}{255}{255}{\ to}%
\tokent{255}{255}{255}{\ click}%
\tokent{255}{255}{255}{\ on}%
\tokent{255}{255}{255}{\ contact}%
\tokent{255}{255}{255}{\ link}%
\tokent{255}{255}{255}{\ again}%
\tokent{255}{255}{255}{":}%
\tokent{255}{255}{255}{\ "}%
\tokent{255}{255}{255}{yes}%
\tokent{255}{255}{255}{",}%
\tokent{255}{255}{255}{\ "}%
\tokent{255}{255}{255}{Contact}%
\tokent{255}{255}{255}{\ page}%
\tokent{255}{255}{255}{\ loaded}%
\tokent{255}{255}{255}{":}%
\tokent{255}{255}{255}{\ "}%
\tokent{255}{255}{255}{yes}%
\tokent{255}{255}{255}{",}%
\tokent{255}{255}{255}{\ "}%
\tokent{255}{255}{255}{Contact}%
\tokent{255}{255}{255}{\ page}%
\tokent{255}{255}{255}{\ content}%
\tokent{255}{255}{255}{":}%
\tokent{255}{255}{255}{\ "}%
\tokent{255}{255}{255}{Contact}%
\tokent{255}{255}{255}{\ form}%
\tokent{255}{255}{255}{\ with}%
\tokent{255}{255}{255}{\ fields}%
\tokent{255}{255}{255}{\ for}%
\tokent{255}{255}{255}{\ name}%
\tokent{255}{255}{255}{,}%
\tokent{255}{255}{255}{\ email}%
\tokent{255}{255}{255}{,}%
\tokent{255}{255}{255}{\ and}%
\tokent{255}{255}{255}{\ message}%
\tokent{255}{255}{255}{",}%
\tokent{255}{255}{255}{\ "}%
\tokent{255}{255}{255}{Contact}%
\tokent{255}{255}{255}{\ email}%
\tokent{255}{255}{255}{\ found}%
\tokent{255}{255}{255}{":}%
\tokent{255}{255}{255}{\ "}%
\tokent{255}{255}{255}{no}%
\tokent{255}{255}{255}{",}%
\tokent{255}{255}{255}{\ "}%
\tokent{198}{211}{255}{Sc}%
\tokent{255}{255}{255}{rolled}%
\tokent{255}{255}{255}{\ down}%
\tokent{255}{255}{255}{\ contact}%
\tokent{255}{255}{255}{\ page}%
\tokent{255}{255}{255}{":}%
\tokent{255}{255}{255}{\ "}%
\tokent{255}{255}{255}{yes}%
\tokent{255}{255}{255}{",}%
\tokent{255}{255}{255}{\ "}%
\tokent{255}{255}{255}{Individual}%
\tokent{254}{254}{255}{\ contacts}%
\tokent{255}{255}{255}{\ section}%
\tokent{255}{255}{255}{\ visible}%
\tokent{255}{255}{255}{":}%
\tokent{255}{255}{255}{\ "}%
\tokent{255}{255}{255}{yes}%
\tokent{255}{255}{255}{",}%
\tokent{255}{255}{255}{\ "}%
\tokent{255}{255}{255}{Vendor}%
\tokent{255}{255}{255}{\ contact}%
\tokent{255}{255}{255}{\ emails}%
\tokent{255}{255}{255}{\ found}%
\tokent{255}{255}{255}{":}%
\tokent{255}{255}{255}{\ "}%
\tokent{255}{255}{255}{yes}%
\tokent{255}{255}{255}{",}%
\tokent{255}{255}{255}{\ "}%
\tokent{255}{255}{255}{Food}%
\tokent{255}{255}{255}{\ vendors}%
\tokent{255}{255}{255}{\ email}%
\tokent{255}{255}{255}{":}%
\tokent{255}{255}{255}{\ "}%
\tokent{255}{255}{255}{food}%
\tokent{255}{255}{255}{@c}%
\tokent{255}{255}{255}{raw}%
\tokent{255}{255}{255}{f}%
\tokent{255}{255}{255}{ords}%
\tokent{255}{255}{255}{ville}%
\tokent{255}{255}{255}{str}%
\tokent{255}{255}{255}{aw}%
\tokent{255}{255}{255}{berry}%
\tokent{255}{255}{255}{fest}%
\tokent{255}{255}{255}{.com}%
\tokent{255}{255}{255}{",}%
\tokent{255}{255}{255}{\ "}%
\tokent{255}{255}{255}{Ar}%
\tokent{255}{255}{255}{ts}%
\tokent{251}{252}{255}{\ \&}%
\tokent{130}{158}{255}{\ Crafts}%
\tokensub{(0.85) [\textperiodcentered{}crafts:0.15]}%
\tokent{254}{255}{255}{\ vendor}%
\tokent{255}{255}{255}{\ email}%
\tokent{255}{255}{255}{":}%
\tokent{255}{255}{255}{\ "}%
\tokent{255}{255}{255}{d}%
\tokent{255}{255}{255}{cook}%
\tokent{255}{255}{255}{sey}%
\tokent{255}{255}{255}{str}%
\tokent{255}{255}{255}{aw}%
\tokent{255}{255}{255}{berry}%
\tokent{255}{255}{255}{fest}%
\tokent{255}{255}{255}{@yahoo}%
\tokent{255}{255}{255}{.com}%
\tokent{255}{255}{255}{",}%
\tokent{255}{255}{255}{\ "}%
\tokent{255}{255}{255}{5}%
\tokent{255}{255}{255}{K}%
\tokent{255}{255}{255}{\ Run}%
\tokent{255}{255}{255}{\ email}%
\tokent{255}{255}{255}{":}%
\tokent{255}{255}{255}{\ "}%
\tokent{255}{255}{255}{j}%
\tokent{255}{255}{255}{th}%
\tokent{255}{255}{255}{ompson}%
\tokent{255}{255}{255}{@}%
\tokent{255}{255}{255}{nm}%
\tokent{255}{255}{255}{.k}%
\tokent{255}{255}{255}{1}%
\tokent{255}{255}{255}{2}%
\tokent{255}{255}{255}{.in}%
\tokent{255}{255}{255}{.us}%
\tokent{255}{255}{255}{",}%
\tokent{255}{255}{255}{\ "}%
\tokent{255}{255}{255}{Chair}%
\tokent{255}{255}{255}{person}%
\tokent{255}{255}{255}{\ email}%
\tokent{255}{255}{255}{":}%
\tokent{255}{255}{255}{\ "}%
\tokent{255}{255}{255}{sf}%
\tokent{255}{255}{255}{est}%
\tokent{255}{255}{255}{queen}%
\tokent{255}{255}{255}{@gmail}%
\tokent{255}{255}{255}{.com}%
\tokent{255}{255}{255}{",}%
\tokent{255}{255}{255}{\ "}%
\tokent{255}{255}{255}{Ent}%
\tokent{255}{255}{255}{ertainment}%
\tokent{255}{255}{255}{\ email}%
\tokent{255}{255}{255}{":}%
\tokent{255}{255}{255}{\ "}%
\tokent{255}{255}{255}{ent}%
\tokent{255}{255}{255}{ertainment}%
\tokent{255}{255}{255}{@c}%
\tokent{255}{255}{255}{raw}%
\tokent{255}{255}{255}{f}%
\tokent{255}{255}{255}{ords}%
\tokent{255}{255}{255}{ville}%
\tokent{255}{255}{255}{str}%
\tokent{255}{255}{255}{aw}%
\tokent{255}{255}{255}{berry}%
\tokent{255}{255}{255}{fest}%
\tokent{255}{255}{255}{.com}%
\tokent{255}{255}{255}{"\}\newline }%
\tokent{255}{255}{255}{Int}%
\tokent{255}{255}{255}{ention}%
\tokent{255}{255}{255}{:}%
\tokent{104}{138}{255}{\ Provide}%
\tokensub{(0.88) [\textperiodcentered{}Find:0.07]}%
\tokent{255}{255}{255}{\ the}%
\tokent{213}{223}{255}{\ contact}%
\tokent{255}{255}{255}{\ email}%
\tokent{255}{255}{255}{\ for}%
\tokent{50}{96}{255}{\ vendor}%
\tokensub{(0.44) [\textperiodcentered{}food:0.56]}%
\tokent{255}{255}{255}{\ inquiries}%
\tokent{255}{255}{255}{\ as}%
\tokent{254}{254}{255}{\ the}%
\tokent{157}{179}{255}{\ answer}%
\tokensub{(0.02) [\textperiodcentered{}task:0.92 \textperiodcentered{}final:0.06]}%
\tokent{245}{247}{255}{\newline }%
\tokent{255}{255}{255}{\textbf{Action}}%
\tokent{255}{255}{255}{\textbf{:}}%
\tokent{255}{255}{255}{\ The}%
\tokent{255}{255}{255}{\ contact}%
\tokent{255}{255}{255}{\ email}%
\tokent{255}{255}{255}{\ for}%
\tokent{255}{255}{255}{\ vendor}%
\tokent{255}{255}{255}{\ inquiries}%
\tokent{222}{230}{255}{\ is}%
\tokent{255}{255}{255}{\ food}%
\tokent{255}{255}{255}{@c}%
\tokent{255}{255}{255}{raw}%
\tokent{255}{255}{255}{f}%
\tokent{255}{255}{255}{ords}%
\tokent{255}{255}{255}{ville}%
\tokent{255}{255}{255}{str}%
\tokent{255}{255}{255}{aw}%
\tokent{255}{255}{255}{berry}%
\tokent{255}{255}{255}{fest}%
\tokent{255}{255}{255}{.com}%
\tokent{255}{255}{255}{\newline }%
\tokent{255}{255}{255}{\textbf{\textless{}tool\_call\textgreater{}}}%
\tokent{255}{255}{255}{\newline }%
\tokent{255}{255}{255}{\{"}%
\tokent{255}{255}{255}{name}%
\tokent{255}{255}{255}{":}%
\tokent{255}{255}{255}{\ "}%
\tokent{255}{255}{255}{computer}%
\tokent{255}{255}{255}{\_use}%
\tokent{255}{255}{255}{",}%
\tokent{255}{255}{255}{\ "}%
\tokent{255}{255}{255}{arguments}%
\tokent{255}{255}{255}{":}%
\tokent{255}{255}{255}{\ \{"}%
\tokent{255}{255}{255}{action}%
\tokent{255}{255}{255}{":}%
\tokent{255}{255}{255}{\ "}%
\tokent{255}{255}{255}{answer}%
\tokent{255}{255}{255}{",}%
\tokent{255}{255}{255}{\ "}%
\tokent{255}{255}{255}{text}%
\tokent{255}{255}{255}{":}%
\tokent{255}{255}{255}{\ "}%
\tokent{255}{255}{255}{The}%
\tokent{255}{255}{255}{\ contact}%
\tokent{255}{255}{255}{\ email}%
\tokent{255}{255}{255}{\ for}%
\tokent{255}{255}{255}{\ vendor}%
\tokent{255}{255}{255}{\ inquiries}%
\tokent{255}{255}{255}{\ is}%
\tokent{255}{255}{255}{\ food}%
\tokent{255}{255}{255}{@c}%
\tokent{255}{255}{255}{raw}%
\tokent{255}{255}{255}{f}%
\tokent{255}{255}{255}{ords}%
\tokent{255}{255}{255}{ville}%
\tokent{255}{255}{255}{str}%
\tokent{255}{255}{255}{aw}%
\tokent{255}{255}{255}{berry}%
\tokent{255}{255}{255}{fest}%
\tokent{255}{255}{255}{.com}%
\tokent{255}{255}{255}{"\}\}\newline }%
\tokent{255}{255}{255}{\textbf{\textless{}/tool\_call\textgreater{}}}%
\tokent{255}{255}{255}{\textbf{\textless{}|im\_end|\textgreater{}}}
\end{panelmd}